\documentclass{article} 
\usepackage[preprint, nonatbib]{nips_2018}

\usepackage{hyperref}
\usepackage{url}
\usepackage{graphicx}
\usepackage{booktabs}
\usepackage{amssymb}
\usepackage{amsmath}
\usepackage{subcaption}

\usepackage{adjustbox} 
\usepackage{multirow}
\usepackage{wrapfig}
\usepackage{tikz}
\usepackage{xspace}
\usepackage{soul}
\usepackage[utf8]{inputenc}
\usepackage[T1]{fontenc}
\usepackage{amsfonts} 
\usepackage{nicefrac}
\usepackage{microtype}  
\usepackage{xcolor}
\usepackage{amsbsy}

\newcommand{\eg}{\textit{e}.\textit{g}., }
\newcommand{\etal}{\textit{et al}. }

\definecolor{bg_blue}{RGB}{213,227,251}
\definecolor{bg_yellow}{RGB}{250,243,187}
\definecolor{bg_purple}{RGB}{177,167,207}
\definecolor{bg_red}{RGB}{200,169,188}
\definecolor{bg_green}{RGB}{192,213,175}
\definecolor{bg_skin}{RGB}{245,232,210}

\usepackage[utf8]{inputenc} % allow utf-8 input
\usepackage[T1]{fontenc}    % use 8-bit T1 fonts
\usepackage{tikz}
\usetikzlibrary{arrows,shapes,snakes,automata,backgrounds,fit,petri}
\usepackage{adjustbox}
\usepackage{tcolorbox}

\definecolor{red_color}{RGB}{255,0,0}

\definecolor{yellow_color}{RGB}{255,202,47}

\definecolor{purple_color}{RGB}{64,103,139}

\definecolor{dark_red}{RGB}{153, 31, 41}

\definecolor{green_color}{RGB}{130,139,78}

\definecolor{brown_color}{RGB}{205,90,161}

\definecolor{lg_color}{RGB}{63,147,139}

\definecolor{com_color}{RGB}{0,0,139}

\definecolor{orange_color}{RGB}{255,148,63}

\definecolor{gray_color}{RGB}{169,169,169}

\definecolor{lightgray}{RGB}{220,220,220}

\definecolor{lightgreen}{RGB}{179,207,176}

\definecolor{lightblue}{RGB}{215,227,240}

\definecolor{lightyellow}{RGB}{233,212,136}

\newcommand{\hatememe}{{\fontfamily{lmtt}\selectfont HatefulMemes}}

\newcommand{\mvsasingle}{{\fontfamily{lmtt}\selectfont MVSA-Single}}

\newcommand{\mvsamultiple}{{\fontfamily{lmtt}\selectfont MVSA-Multiple}}

\newcommand{\pan}{{\fontfamily{lmtt}\selectfont PAN18}}

\newcommand{\uppam}{{\fontfamily{lmtt}\selectfont UPPAM}}

\newcommand{\fakenewsnet}{{\fontfamily{lmtt}\selectfont FakeNewsNet}}

\title{GPT-4V(ision) as A Social Media Analysis Engine}

\author{
Hanjia~Lyu$^{1,*}$ \quad Jinfa~Huang$^{1,*}$ \quad Daoan~Zhang$^{1,*}$ \quad Yongsheng~Yu$^{1,*}$ \quad Xinyi~Mou$^{2,*}$\\
\textbf{Jinsheng~Pan$^{1}$} \quad \textbf{Zhengyuan Yang$^{1}$} \quad \textbf{Zhongyu~Wei$^{2}$} \quad \textbf{Jiebo~Luo}$^{1}$\\
$^{1}$University of Rochester \quad $^{2}$Fudan University\\
\\
$^{*}$ Core Contributor
}

\begin{document}

\maketitle
\begin{figure}[htbp]
    \centering
\vspace{-10mm}
    \includegraphics[width=0.18\linewidth]{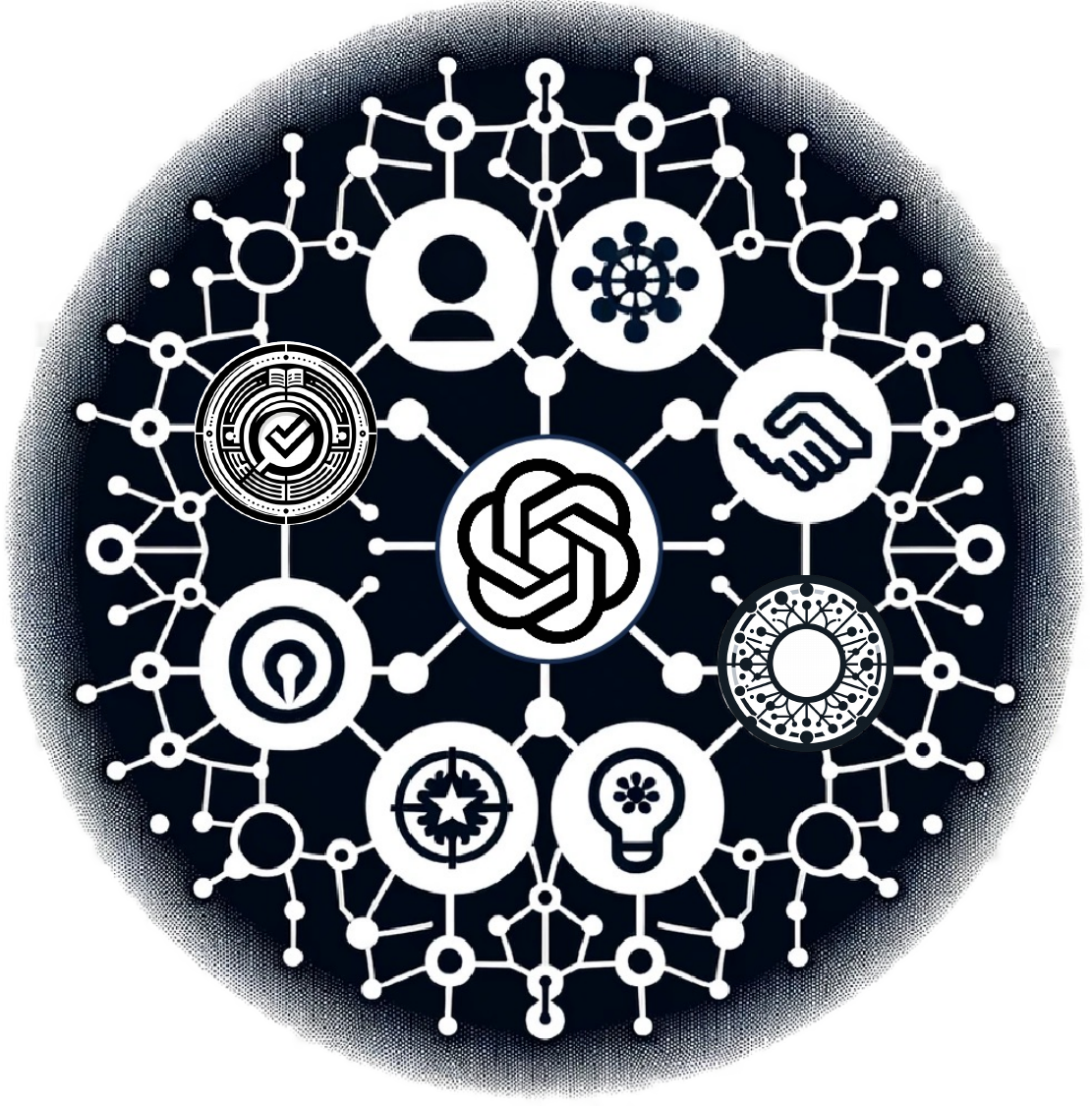}
% \vspace{-5mm}
\end{figure}
\begin{abstract}
    Recent research has offered insights into the extraordinary capabilities of Large Multimodal Models (LMMs) in various general vision and language tasks. There is growing interest in how LMMs perform in more specialized domains. Social media content, inherently multimodal, blends text, images, videos, and sometimes audio. To effectively understand such content, models need to interpret the intricate interactions between these diverse communication modalities and their impact on the conveyed message. Understanding social multimedia content remains a challenging problem for contemporary machine learning frameworks. In this paper, we explore GPT-4V(ision)'s capabilities for social multimedia analysis. We select five representative tasks, including sentiment analysis, hate speech detection, fake news identification, demographic inference, and political ideology detection, to evaluate GPT-4V. Our investigation begins with a preliminary quantitative analysis for each task using existing benchmark datasets, followed by a careful review of the results and a selection of qualitative samples that illustrate GPT-4V’s potential in understanding multimodal social media content. GPT-4V demonstrates remarkable efficacy in these tasks, showcasing strengths such as joint understanding of image-text pairs, contextual and cultural awareness, and extensive commonsense knowledge. Despite the overall impressive capacity of GPT-4V in the social media domain, there remain notable challenges. GPT-4V struggles with tasks involving multilingual social multimedia comprehension and has difficulties in generalizing to the latest trends in social media. Additionally, it exhibits a tendency to generate erroneous information in the context of evolving celebrity and politician knowledge, reflecting the known hallucination problem. Our hope is that this preliminary study will provide insights into further research across disciplines, particularly in computational social science and social media-related studies. The insights gleaned from our findings underscore a promising future for LMMs in enhancing our comprehension of social media content and its users through the analysis of multimodal information. All images and prompts used in this report will be available at \href{https://github.com/VIStA-H/GPT-4V_Social_Media}{https://github.com/VIStA-H/GPT-4V\_Social\_Media}.\footnote{Note that we conducted our experiments and analysis before November 6$^{th}$, 2023 via the official ChatGPT webpage.}

    \textcolor{red}{\textbf{Disclaimer: This paper contains some examples of offensive social media content. Reader discretion is advised.}}
\end{abstract}

\clearpage

\tableofcontents
% \clearpage

\listoffigures
\clearpage

\section{Introduction}\label{sec:intro}
% \subsection{Motivation and Overview}\label{sec:motivation}
\textbf{Motivation and Overview.} Social multimedia analysis encompasses a diverse range of tasks, each characterized by its inherent complexity and nuanced characteristics~\cite{baltruvsaitis2018multimodal,you2015robust,kiela2020hateful,tan2020detecting,rangel2018overview,conover2011predicting,li2019survey,park2014computational,naaman2012social,yazdavar2020multimodal}. This field is distinguished by its multimodal nature, where text, images, videos, and other media forms intertwine to convey rich, multifaceted messages. The tasks within this domain, such as sentiment analysis, hate speech detection, and fake news identification, are not only about parsing textual content but also about deciphering the subtleties conveyed through visual cues. The interplay of various forms of data adds layers of context, requiring an advanced level of comprehension that transcends traditional unimodal analysis. Additionally, the rapid pace of social media evolution introduces constant variability, with new slang, symbols, and trends emerging continuously, making the analysis ever more challenging. 

In this intricate milieu, large multimodal models like GPT-4V(ision)~\cite{openai2023gpt4,yang2023dawn,zhang2023gpt} represent a leap forward in artificial intelligence. Their ability to process and interpret complex, multimodal information makes them particularly suited for the demands of social multimedia analysis. Recent research has undertaken an extensive evaluation of GPT-4V's capabilities across a spectrum of vision and language tasks~\cite{yang2023dawn, wu2023early, zhang2023lost, zhang2023gpt, cui2023holistic, shi2023exploring, lin2023mm, lu2023mathvista, liu2023hallusionbench, yang2023performance, li2023comprehensive, wu2023can, cao2023towards, zhou2023exploring, jin2022expectation}. These investigations have shed valuable light on the potential of Large Multimodal Models (LMMs) like GPT-4V in the domains of both vision and language. However, the broader application of these techniques for societal benefit remains a question. Social media-related studies have the potential to enhance social good by addressing harmful content, fostering positive behavior, and bolstering public safety. They can also enable community building, improved communication, and the empowerment of marginalized groups, leading to a more inclusive and informed society. Assessing the effectiveness of LMMs like GPT-4V in comprehending and responding to social media nuances is pivotal for several reasons. First, it helps assess the model's capability to handle the diversity and complexity of real-world data, providing insights into its practical applicability. Second, such evaluation sheds light on the model's understanding of cultural and contextual nuances, which are vital for accurate analysis in a globally connected, digital world. Lastly, it aids in identifying areas where these models excel and where they need improvement, guiding future advancements in AI and machine learning.

Our study aims to contribute to the understanding of how effectively GPT-4V can navigate the nuanced domain of social multimedia analysis, highlighting its strengths and areas for development. Considering the unique characteristics of multimodal social media content, our exploration of GPT-4V's capabilities for social multimedia analysis is guided by the following questions:
\begin{itemize}
    \item[1.] \textit{How effectively does GPT-4V understand the multimodal content on social media platforms?} Social media data is inherently multimodal, which requires sophisticated models that can process and relate information from these different modalities. GPT-4V exhibits a remarkable capability to understand the visual and textual elements jointly. It demonstrates its aptitude for explaining the connections between the images and text frequently found in social media posts, and how these elements can collaboratively serve a multitude of analytical purposes.
    \item[2.] \textit{Does GPT-4V demonstrate contextual awareness in its interpretation of social media multimedia content?} The meaning of content on social media is often context-dependent. Effective analysis requires understanding the context in which content is produced and consumed. Our observations suggest that GPT-4V possesses the capability to understand context and subtleties within multimodal social media content such as memes, puns, and misspellings, \textit{etc}. This ability paves the way for its application across various domains, including but not limited to, the automatic detection of sarcasm~\cite{joshi2017automatic} and the study of visual persuasion~\cite{joo2014visual}.
    \item[3.] \textit{What is GPT-4V's performance when analyzing novel content?}  Social media language and content trends are constantly changing. Analysis methods should be able to adapt to new knowledge, slang, memes, and communication styles. GPT-4V, while adept in various analytical tasks, still faces challenges in accurately assessing areas like ideology detection and fake news identification, particularly when these require insights into knowledge or contexts beyond its training data.
    \item[4.] \textit{How does GPT-4V navigate the complexities of language and cultural diversity in its analysis of multimodal social media content?} Social media platforms are used globally, necessitating models that can handle multiple languages and cultural contexts. GPT-4V, in its current state, demonstrates limitations when it comes to managing the intricacies involved in language and cultural heterogeneity in its evaluations of multimodal social media content. Its proficiency is notably centered around English, with less effective understanding of other languages. Furthermore, the model exhibits a lack of comprehensive cultural sensitivity, which is essential for navigating the complexities of global social media communications effectively.
\end{itemize}

\begin{figure}[t]
    \centering
    \includegraphics[width=\linewidth]{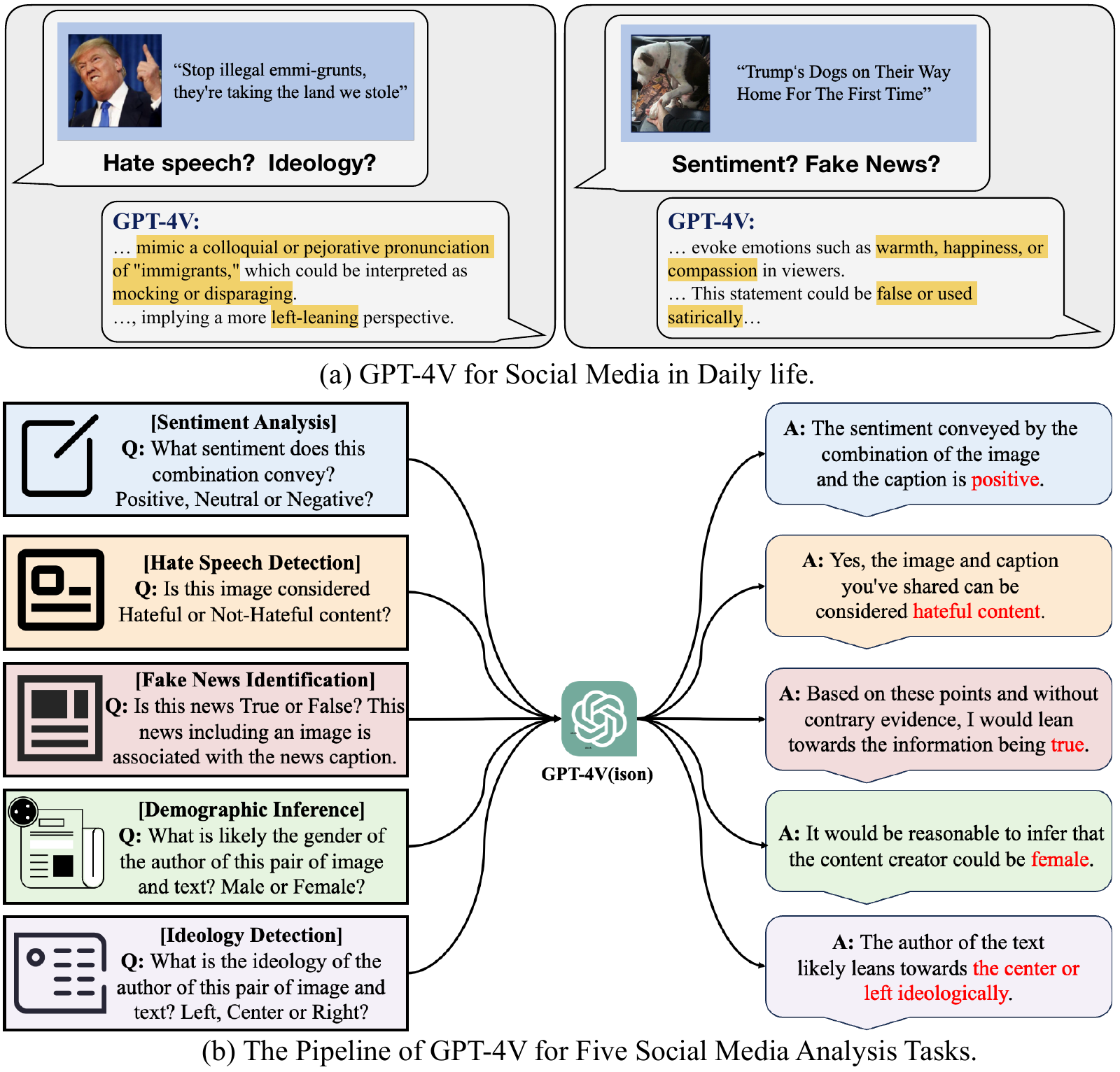}
    \caption[Section~\ref{sec:intro}: social multimedia analysis task description.]{In this study, we carefully select five typical social multimedia analysis tasks including sentiment analysis, hateful speech detection, fake news identification, demographic inference, and ideology detection. We adopt GPT-4V as a unified framework with different prompts (\eg \textit{What sentiment does this combination convey?}) to explore the GPT-4V's ability for social multimedia.}
    \vspace{-3mm}
    \label{fig:task_intro}
\end{figure}

% \subsection{Our Approach in Exploring GPT-4V}\label{sec:appraoch}
\begin{figure}[htbp]
    \centering
    \includegraphics[width=\linewidth]{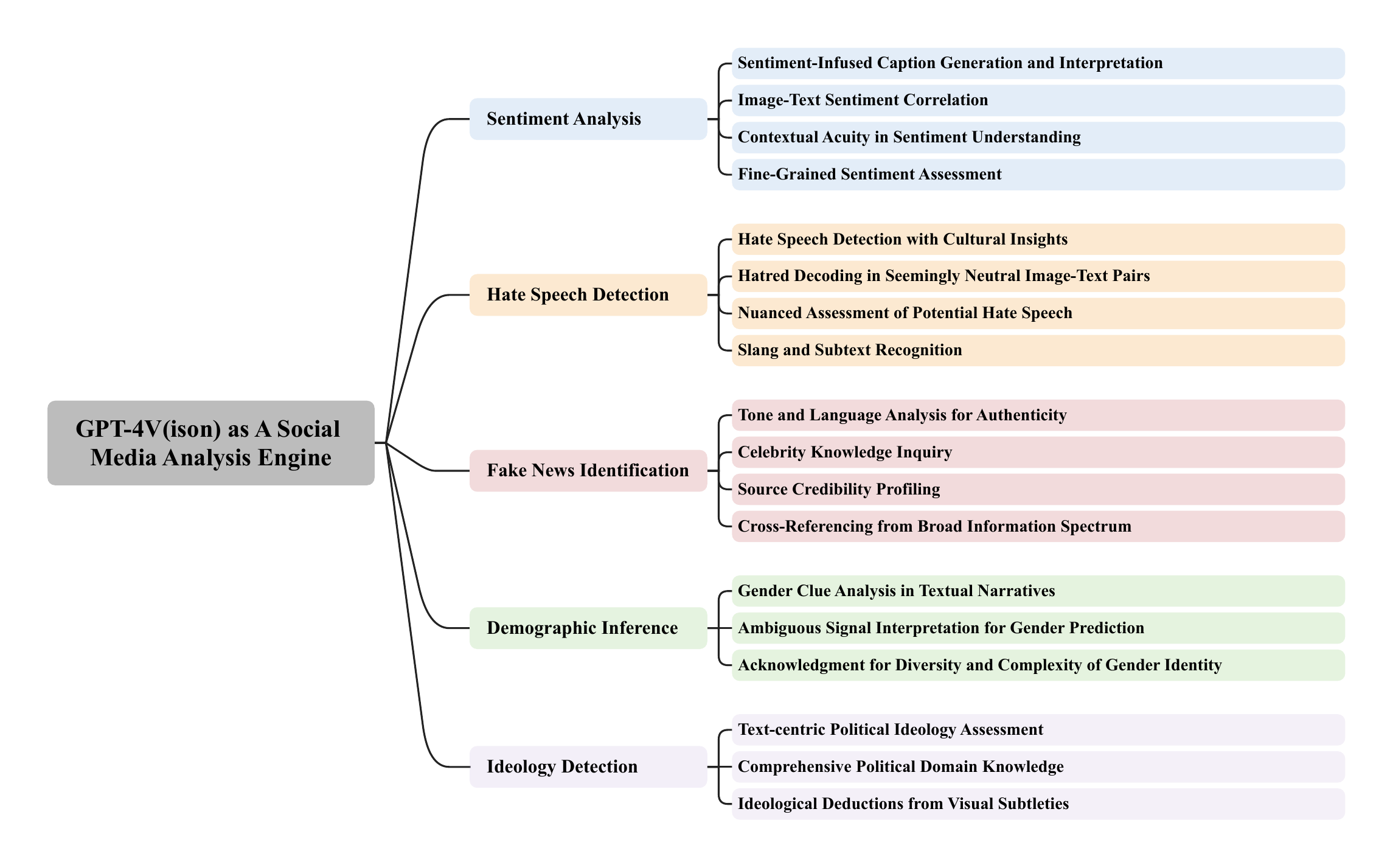}
    \caption[Section~\ref{sec:properties}: overview of emerging properties of social multimedia analysis tasks.]{Overview of the emerging properties of GPT-4V for social multimedia analysis tasks.}
    \vspace{-5mm}
    \label{fig:overview_sec_3}
\end{figure}
\textbf{Our Approach in Exploring GPT-4V(ison).} Social multimedia analysis includes various tasks that involve the extraction, interpretation, and classification of information from content shared on social media platforms. In this report, we focus on tasks that are pivotal in this area, each of which is associated with its own set of quantitative benchmark datasets. These tasks encompass sentiment analysis, hate speech detection, fake news identification, demographic inference, and ideology detection (summarized in Figure~\ref{fig:task_intro}). For each of these tasks, we draw samples from two to three well-established benchmark datasets to present preliminary quantitative results and qualitative insights into the performance of GPT-4V in the context of social multimedia analysis. Figure~\ref{fig:overview_sec_3} provides an overview of the emerging properties identified in GPT-4V within the social multimedia analysis area. As highlighted by Yang~\etal~\cite{yang2023dawn}, it has become apparent that some of the existing benchmarks may no longer effectively evaluate LMMs. In the course of our exploration, we have encountered similar challenges such as GPT-4V's propensity for memorization during training within the context of social media analysis. Therefore, in alignment with Yang~\etal's approach, we primarily rely on qualitative results to provide preliminary insights into the potential capabilities of GPT-4V in the domain of social multimedia analysis. Moreover, we have intentionally included a mix of examples, drawing from both existing datasets and newly designed challenging scenarios aimed at evaluating LMMs' performance.
%For each dataset, we have conducted a random sampling of data samples, subsequently reviewing GPT-4V's performance on them. Next, we have curated a selection of representative examples, which are presented in this report. %It is important to note that our primary objective is to show the potential capabilities of GPT-4V in social multimedia analysis tasks through the presentation of qualitative examples. Our intention here is not to offer a comprehensive quantitative evaluation of the reliability or consistency of these capabilities.

In Section~\ref{sec:properties}, we present our exploration of GPT-4V on the five tasks. In particular, analysis on sentiment analysis, hate speech detection, fake news identification, demographic inference, and ideology detection are discussed in Sections~\ref{sec:sentiment}, \ref{sec:hate}, \ref{sec:fake}, \ref{sec:demographic}, and \ref{sec:ideology}, respectively. In Section~\ref{sec:challenge}, we discuss challenges and potential future opportunities. Finally, we conclude our report in Section~\ref{sec:conclusion}.

\paragraph{Ethical Consideration.} This report evaluates the capabilities of GPT-4V across five social multimedia analysis tasks: sentiment analysis, hate speech detection, fake news identification, demographic inference, and ideology detection. Please be advised that some sections of this report \textit{may include content that readers find offensive or sensitive}. This is an inherent aspect of analyzing real-world social media data, especially when dealing with topics such as hate speech and fake news. The intent is purely academic and informational, with no aim to offend or harm. In our commitment to privacy and ethical standards, we have taken careful measures to anonymize the data. To protect human facial privacy, we have applied masks to obscure the identities of all individuals during figure visualizations, except for public figures and celebrities. This anonymization is done to respect the privacy of individuals and to adhere to common ethical guidelines in data handling and reporting. The findings and conclusions within this report are based on the data available and the capabilities of the GPT-4V model as of the date of this publication, which intends to contribute to the ongoing discourse in the field of AI and social media analysis and does not necessarily reflect the full scope of complexities involved in real-world scenarios.

\section{Experiments for Selected Social Multimedia Analysis Tasks}\label{sec:properties}

\subsection{Sentiment Analysis}\label{sec:sentiment}

\subsubsection{Task Setting and Preliminary Quantitative Results}
\begin{figure}[htbp]
    \centering
    \includegraphics[width=\linewidth]{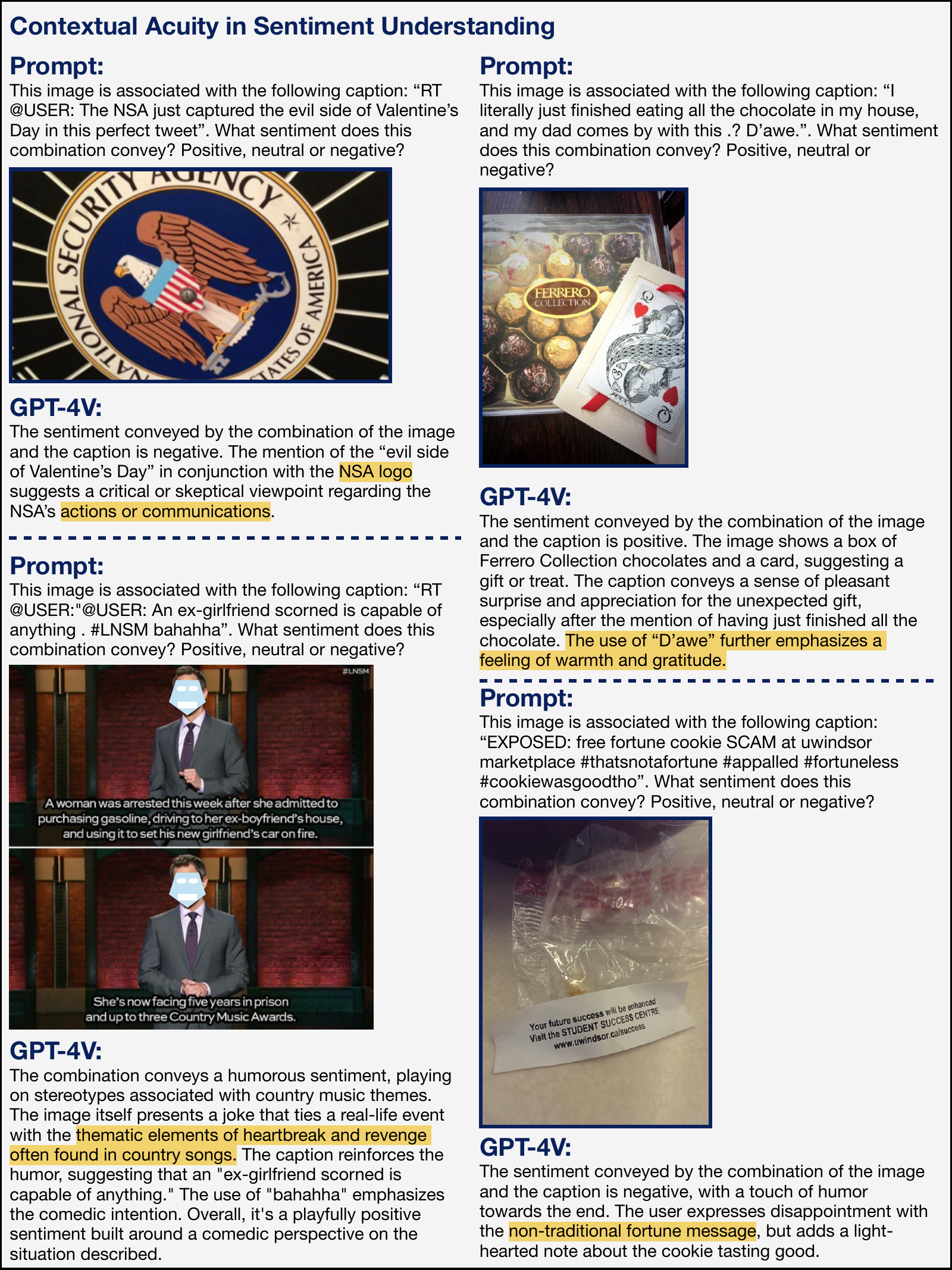}
    \caption[Section~\ref{sec:sentiment}: contextual acuity in sentiment understandng.]{Qualitative results on contextual acuity in sentiment understandng. GPT-4V generates responses that demonstrate cultural awareness and are contextually relevant within the context of social media. The components that are related to \colorbox{lightyellow}{contextual understanding} are highlighted. Refer to Section~\ref{sec:sentiment} for detailed discussions.}
\label{fig:sentiment_contextual_cultural_understanding}
\end{figure}

\begin{figure}[htbp]
    \centering
    \includegraphics[width=\linewidth]{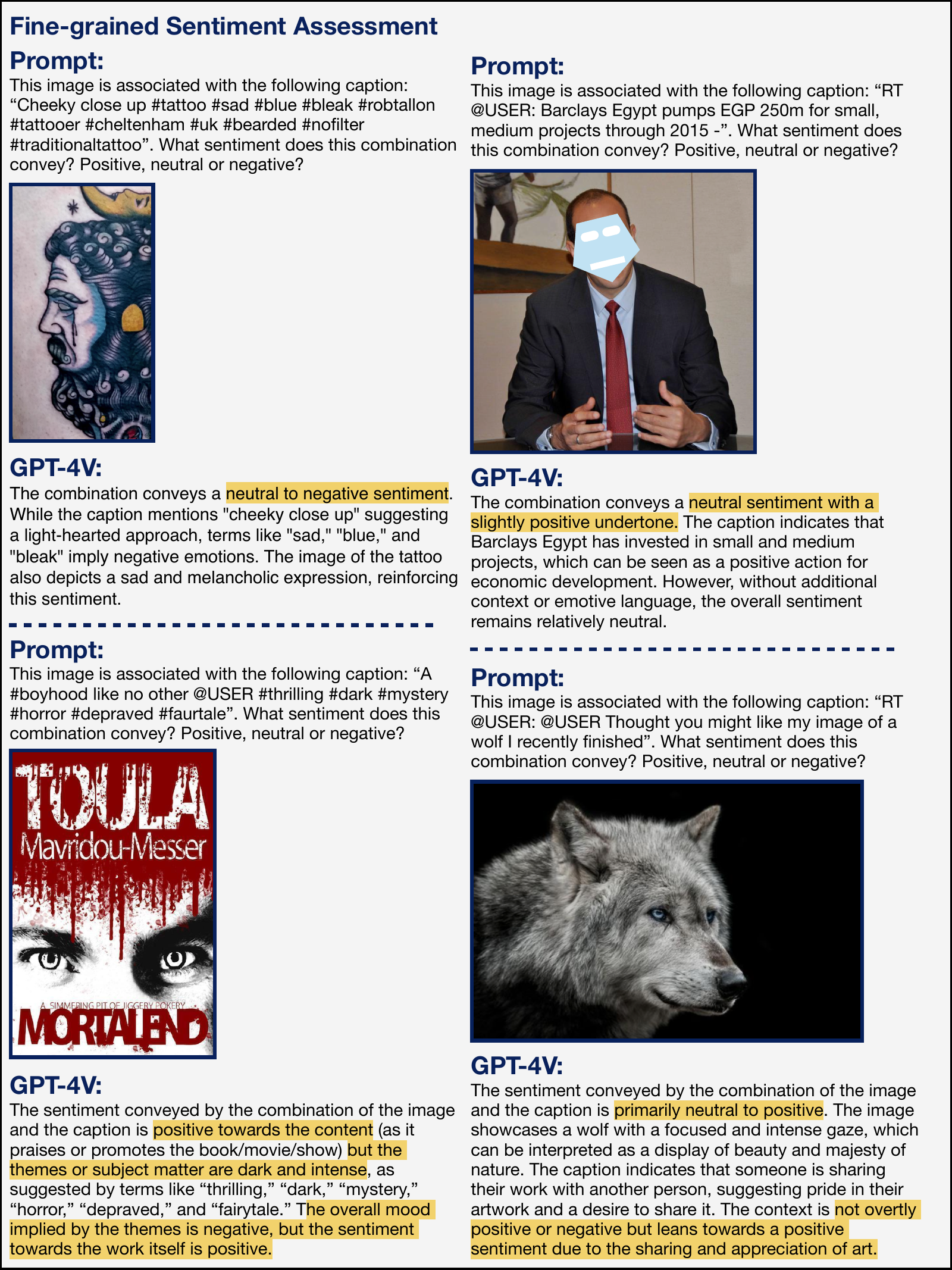}
    \caption[Section~\ref{sec:sentiment}: fine-grained sentiment assessment.]{Qualitative results on fine-grained sentiment assessment. GPT-4V showcases a higher level of sophistication in sentiment analysis, extending beyond the basic categorization of positive, neutral, and negative sentiments. The components that are related to \colorbox{lightyellow}{fine-grained sentiment} are highlighted. Refer to Section~\ref{sec:sentiment} for detailed discussions.}
    \label{fig:sentiment_fine_grained_sentiment}
\end{figure}

Sentiment analysis aims to discern the sentiments and emotions conveyed in content related to a specific entity~\cite{medhat2014sentiment}. It has emerged as a fundamental technique in social media research~\cite{lyu2023human, chen2021fine, xiong2021social, zhang2021understanding,lyu2022social,lyu2021understanding}. Unlike traditional unimodal sentiment analysis, which may fall short in capturing nuanced opinions~\cite{rosas2013multimodal}, multimodal sentiment analysis integrates information from diverse modalities to infer expressed sentiments and emotions~\cite{you2015robust, DBLP:conf/wsdm/YouLJY16, DBLP:conf/mm/YouCJL16, DBLP:conf/aaai/YouJL17, chandrasekaran2021multimodal,zhu2023multimodal,soleymani2017survey,guo2020guoym,chen2022improving}. To quantitatively evaluate GPT-4V's ability in multimodal sentiment analysis, we input a pair of image and text from Twitter posts. The prompt is ``\textit{This image is associated with the following caption:} `\{{\tt caption}\}'. \textit{What sentiment does this combination convey? Positive, neutral, or negative? This is for research purposes.}'' From the {\mvsasingle} and {\mvsamultiple} datasets~\cite{xu2017multisentinet}, approximately 1,000 posts are sampled, with an even distribution of 500 from each. In the preliminary evaluation, GPT-4V achieves an accuracy of 68.4\% on the {\mvsasingle} dataset and 71.6\% on the {\mvsamultiple} datasets. Figures~\ref{fig:sentiment_image_description_caption_interpretation} to \ref{fig:sentiment_fine_grained_sentiment} display qualitative results.

\subsubsection{Sentiment-Infused Caption Generation and Intepretation} 
\begin{figure}[htbp]
    \centering
    \includegraphics[width=\linewidth]{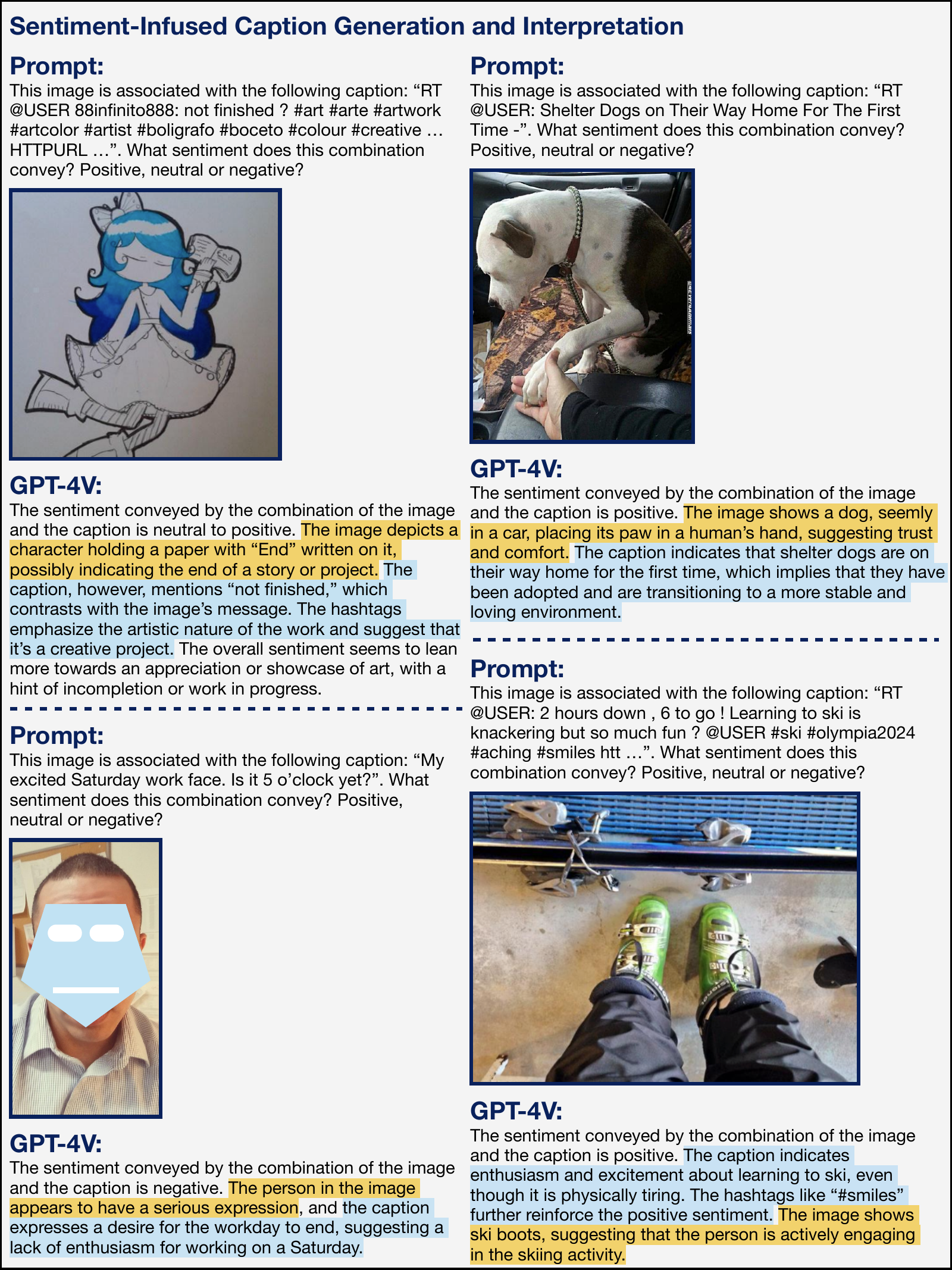}
    \caption[Section~\ref{sec:sentiment}: sentiment-infused caption generation and interpretation.]{Qualitative results on sentiment-infused caption generation and interpretation. GPT-4V is able to describe the image contents and interpret the original caption, with a particular emphasis on capturing the conveyed sentiment. The \colorbox{lightblue}{image descriptions} and \colorbox{lightyellow}{caption interpretations} are highlighted. Refer to Section~\ref{sec:sentiment} for detailed discussions.}
    \label{fig:sentiment_image_description_caption_interpretation}
\end{figure}

Visual sentiment analysis focuses on analyzing emotions and sentiments conveyed through visual content, while textual sentiment analysis deals with extracting sentiment and emotional information from text data. Even within a single modality, assessing sentiment presents significant challenges due to various factors such as subjectivity, ambiguity, cultural variability, and limited semantic comprehension~\cite{zadeh2017tensor, zhu2023multimodal,chandrasekaran2021multimodal,soleymani2017survey,attia2018multilingual, DBLP:conf/eccv/ChenZYFWJL18}.  In Figure~\ref{fig:sentiment_image_description_caption_interpretation}, we observe that GPT-4V adeptly describes the contents of the image (\eg \textit{``... depicts a character holding a paper with `End' written on it ...''}) and interprets the original caption (\eg \textit{``... implies that they have been adopted and are transitioning to a more stable and loving environment ...''}) with regard to sentiment (\eg \textit{``... suggesting trust and comfort ...''}).

\subsubsection{Image-Text Sentiment Correlation} 
\begin{figure}[htbp]
    \centering
    \includegraphics[width=\linewidth]{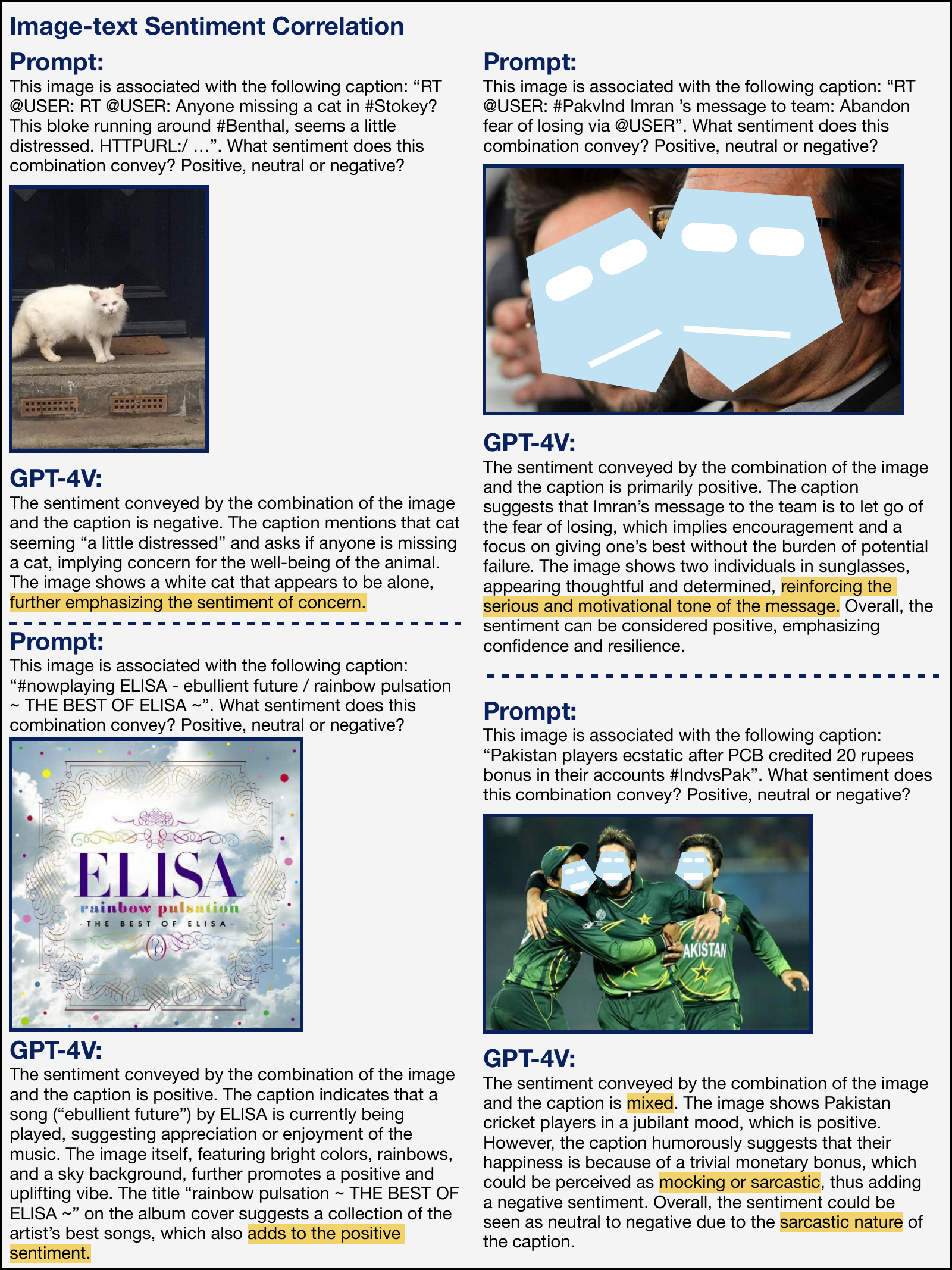}
    \caption[Section~\ref{sec:sentiment}: image-text sentiment correlation.]{Qualitative results on image-text sentiment correlation. GPT-4V explicitly explains the interplay between the image and text. The \colorbox{lightyellow}{image-text relation interpretations} are highlighted. Refer to Section~\ref{sec:sentiment} for detailed discussions.}
    \label{fig:sentiment_explicit_image_text_relation_interpretation}
\end{figure}

Combining information from different modalities is a challenging task as different modalities may have varying degrees of influence on sentiment~\cite{zadeh2017tensor,chandrasekaran2021multimodal}. Beyond the generation of sentiment-focused image descriptions and caption interpretations, GPT-4V goes a step further by explicitly explaining the relationship between the image and text in the context of sentiment analysis. For instance, it articulates how one modality can accentuate the sentiment conveyed by another (see Figure~\ref{fig:sentiment_explicit_image_text_relation_interpretation}). Moreover, there can be divergent sentiment polarities among different modalities; for instance, a piece of text may express negativity while its accompanying visual element conveys positivity~\cite{ji2018cross}. As illustrated in the fourth example in Figure~\ref{fig:sentiment_explicit_image_text_relation_interpretation}, GPT-4V shows its capability to identify such contrasting sentiments in image-text social media posts.

\subsubsection{Contextual Acuity in Sentiment Understanding} Sentiments and emotions are profoundly influenced by culture. What might be regarded as a positive or negative emotion in one culture could be interpreted differently in another. Additionally, the sentiment expressed in social media posts often hinges on the broader contextual backdrop. However, many studies in sentiment analysis tend to focus primarily on the content of social media while overlooking its contextual and cultural dimensions. Neglecting these crucial aspects can complicate the determination of accurate polarity~\cite{chandrasekaran2021multimodal}. Impressively, we have observed that GPT-4V has the capacity to generate prompts that exhibit cultural awareness (\eg \textit{``... the thematic elements of heartbreak and revenge often found in country songs ...''}) and maintain contextual relevance within the context of social media (\eg \textit{``The use of `D'awe' further emphasizes a feeling of warmth and gratitude.''}). This finding is consistent with prior research, which has demonstrated that large language models have the capacity to enhance contextual information~\cite{lyu2023llm}.

\subsubsection{Fine-grained Sentiment Assessment} While basic sentiment analysis typically categorizes image-text pairs into broad sentiments like positive, negative, or neutral, exploring more nuanced sentiment distinctions (such as differentiating between varying degrees of positivity or negativity) is both challenging and often indispensable for both textual and visual content~\cite{you2016building}. More importantly, social media posts also often employ sarcasm and irony for sentiment expression. These modes of expression often rely on intricate context and subtleties that machines find challenging to grasp~\cite{chandrasekaran2021multimodal,zhu2023multimodal,soleymani2017survey, lal2019mixing}. However, our findings indicate that GPT-4V possesses an impressive capability to discern sarcasm and predict nuanced levels of positivity and negativity.

\clearpage

\subsection{Hate Speech Detection}\label{sec:hate}
\subsubsection{Task Setting and Preliminary Quantitative Results}
Hate speech refers to speech or statements, that discriminate against, threaten, or incite violence or prejudicial actions against individuals or groups based on attributes such as their race, religion, ethnicity, sexual orientation, gender, or disease~\cite{kiela2020hateful}. Multimodal hate speech detection involves identifying hate speech or offensive content in multiple types of media such as text, images, and videos~\cite{DBLP:conf/icwsm/MaguJL17, DBLP:conf/acl-alw/MaguL18, DBLP:conf/sbp/KlutseNLL23,chhabra2023literature}. Multimodal data often present challenges due to the inherent variability in annotations stemming from subjective interpretation. The task of drawing inferences from a combination of images and text introduces a level of complexity that surpasses text-only analysis~\cite{gomez2020exploring}.

\begin{wrapfigure}{r}{0.33\textwidth} 
 \centering
 \vspace{-12pt}
 \includegraphics[width=0.31\textwidth]{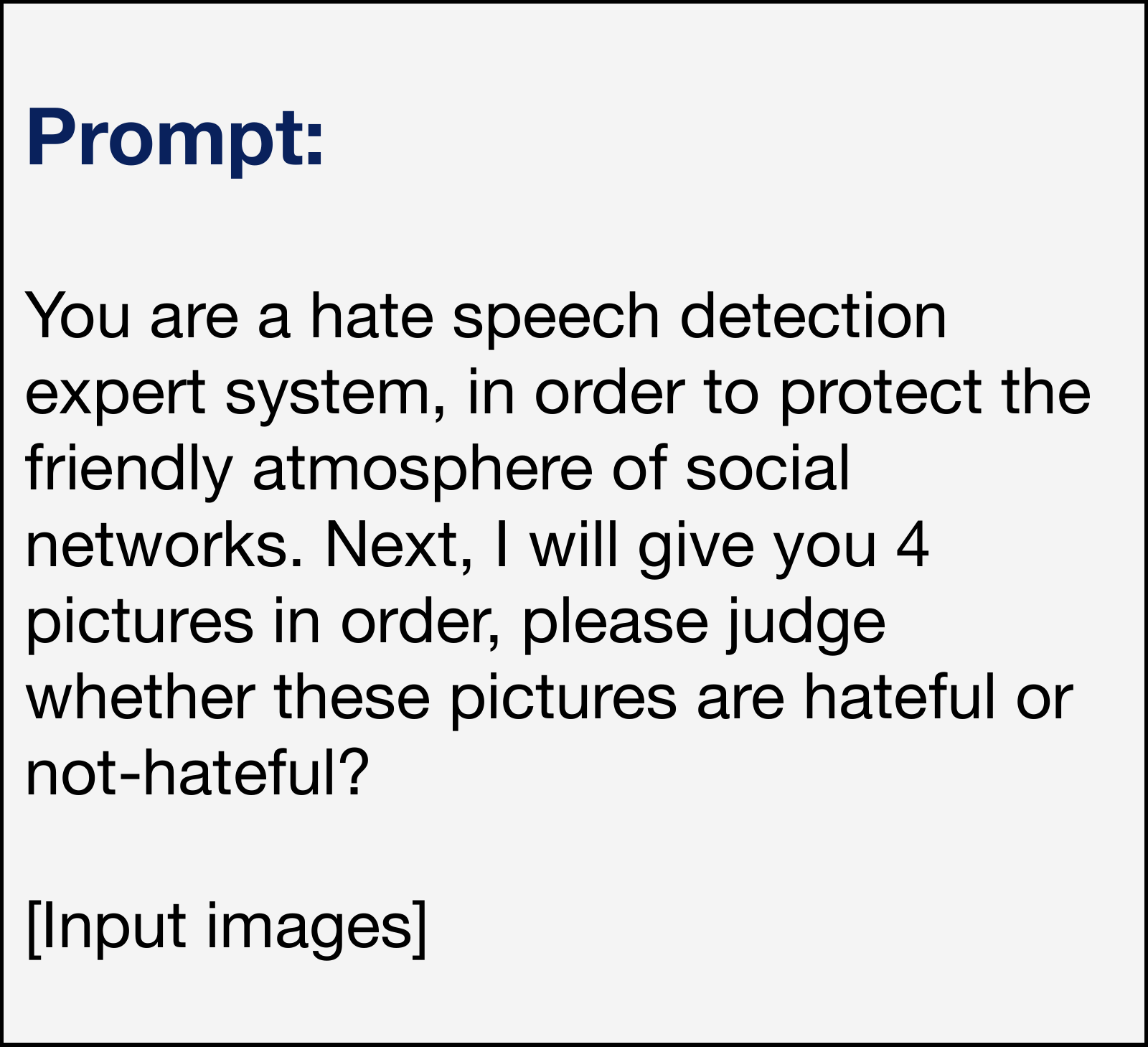}
  % \vspace{-20pt}
  \caption[Section~\ref{sec:hate}: prompt example for the quantitative hate speech detection experiment.]{The prompt example of the quantitative experiment for hate speech detection.}\label{fig:hate_prompt_example}
\vspace{-12pt}
\end{wrapfigure}

To quantitatively analyze the GPT-4V's ability for multimodal hate speech detection, we choose two existing benchmark datasets, including {\hatememe}~\cite{kiela2020hateful} and \texttt{4chan's posts}~\cite{gonzalez2023understanding}. {\hatememe} is a multimodal dataset for hateful meme detection. It is deliberately designed to pose challenges for unimodal models. To achieve this, the dataset incorporates difficult examples, often referred to as ``benign confounders,'' intentionally introduced to complicate the reliance on unimodal signals. \texttt{4chan's posts} focuses on the multimodal analysis of Antisemitism and Islamophobia on 4chan's /pol/. We select all the seen test data from {\hatememe} (1,000 memes) and randonly sample 550 posts from \texttt{4chan's posts}. We instruct the model to produce an output of 0 to signify ``Not-hateful'' and 1 to indicate ``hateful,'' as depicted in Figure~\ref{fig:hate_prompt_example}.

 {\hatememe} serves as a commonly used benchmark dataset for evaluating the performance of large multimodal models. Here, we present the quantitative comparison results between GPT-4V and existing LMMs. Table~\ref{tab:hate_experiment} displays the preliminary evaluation outcomes. GPT-4V exhibits promising results with an AUC of 70.5\% and an accuracy of 70.3\% on {\hatememe}, surpassing the state-of-the-art LMMs by a large margin, approximately 10\%. Notably, GPT-4V achieves an impressive precision score of 87.1\%, underscoring its proficiency in identifying hate content. Furthermore, GPT-4V achieves an accuracy of 60.6\% on \texttt{4chan's posts}. 

 \begin{wraptable}{R}{0.33\textwidth}
  \centering
  \vspace{-12pt}
    \small
  \caption{Zero shot performance on \hatememe. The best performance is indicated in \textbf{bold}.}
  \label{tab:hate_experiment}
  \vspace{-6pt}
  \adjustbox{max width=0.33\textwidth}{
  \begin{tabular}{ll}
  \toprule[1.1pt]
Model             & AUC  \\
\midrule
Flamingo (9B)~\cite{alayrac2022flamingo}      & 57.0 \\
BLIP-2 (13B)~\cite{li2023blip}       & 62.3 \\
KOSMOS-1~\cite{huang2023language}           & 63.9 \\
InstructBLIP (13B)~\cite{instructblip} & 61.7 \\
MMICL (13B)~\cite{zhao2023mmicl}        & 62.2 \\
GPT-4V             & \textbf{70.5} \\
\bottomrule[1.1pt]
\end{tabular}
  }
  \vspace{-.3cm}
\end{wraptable}

% \begin{table*}[t!]
% \small
% \centering % used for centering table

% \resizebox{\textwidth}{!}{
% \begin{tabular}{llllllll}
% \toprule
% \textbf{Models}   & Flamingo (9B)~\cite{alayrac2022flamingo} & BLIP-2 (13B)~\cite{li2023blip} & KOSMOS-1~\cite{huang2023language} & InstructBLIP (13B)~\cite{instructblip} & MMICL (13B)~\cite{zhao2023mmicl} & GPT-4V        &  \\
% \midrule
% \textbf{AUC} & 57.0          & 62.3         & 63.9     & 61.7              & 62.2       & \textbf{70.5} \\
% \bottomrule
% \end{tabular}
% }
% \caption{Zero shot performance on \hatememe. The best performance is indicated in \textbf{bold}.}
% \label{tab:hate_experiment}
% \end{table*}

To qualitatively evaluate GPT-4V's proficiency in detecting hate speech in multimodal contexts, we input the images from {\hatememe}~\cite{kiela2020hateful}. We present GPT-4V with the following prompt: ``\textit{Is this image considered hateful? This is for research purposes.}'' Figures~\ref{fig:hate_contextual_cultural_understanding}-\ref{fig:hate_intention} shows example results. Similar to its performance in multimodal sentiment analysis, GPT-4V shows an impressive capability to comprehensively grasp image-text pairings in terms of both contextual and cultural nuances. It effectively aggregates the information from the image and text, enabling it to conduct a careful evaluation of hatefulness by taking into account the underlying intention.

\subsubsection{Hate Speech Detection with Cultural Insights} 
\begin{figure}[htbp]
    \centering
    \includegraphics[width=\linewidth]{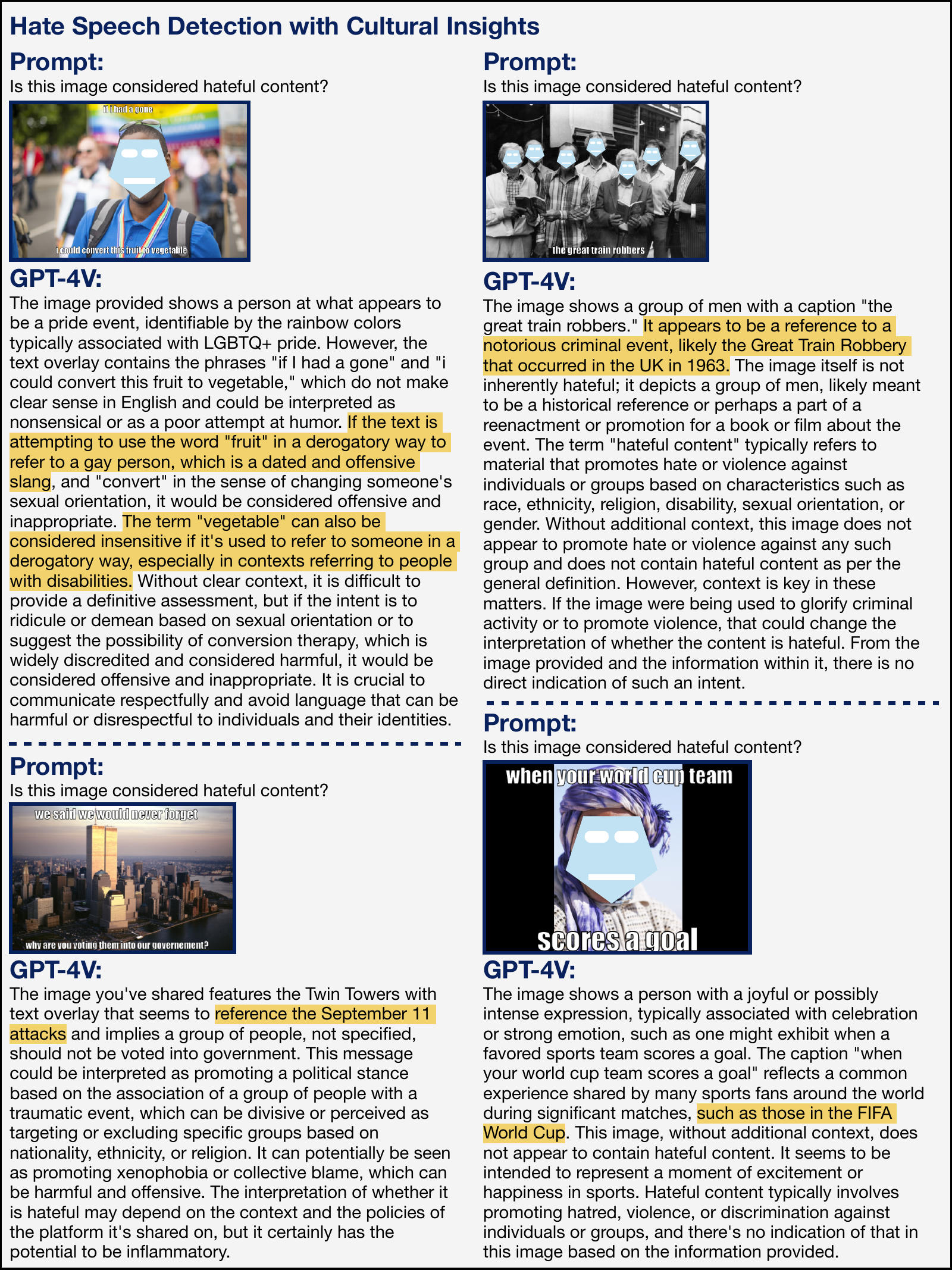}
    \caption[Section~\ref{sec:hate}: hate speech detection with cultural insights.]{Qualitative results on hate speech detection with cultural insights. GPT-4V's responses exhibit a profound cultural awareness and a strong grasp of context when it comes to addressing potential hate speech-related content. The components that are related to \colorbox{lightyellow}{cultural insights} are highlighted. Refer to Section~\ref{sec:hate} for detailed discussions.}
    \label{fig:hate_contextual_cultural_understanding}
\end{figure}

Many existing multimodal hate speech detection systems exhibit limitations tied to the specific characteristics of their training datasets, especially in terms of demographics and cultural nuances~\cite{fortuna2021well,gomez2020exploring}. For instance, a model trained on an Indian political dataset may struggle to generalize effectively when applied to a U.S. health-related dataset. Further, the mere inclusion of offensive terms does not automatically equate to hate speech; rather, it is the context of a post that often dictates whether the content is deemed to constitute hate speech~\cite{gomez2020exploring,DBLP:conf/sbp/KlutseNLL23}. In Figure~\ref{fig:hate_contextual_cultural_understanding}, it becomes evident that GPT-4V possesses contextual and cultural awareness concerning the content featured in image-text pairs. Consider the first post in Figure~\ref{fig:hate_contextual_cultural_understanding} as an example: GPT-4V recognizes that the term ``fruit'' is sometimes used derogatorily to refer to a gay man, while ``vegetable'' is employed offensively in association with people with disabilities.

\subsubsection{Hatred Decoding in Seemingly Neutral Image-Text Pairs}
\begin{figure}[htbp]
    \centering
    \includegraphics[width=\linewidth]{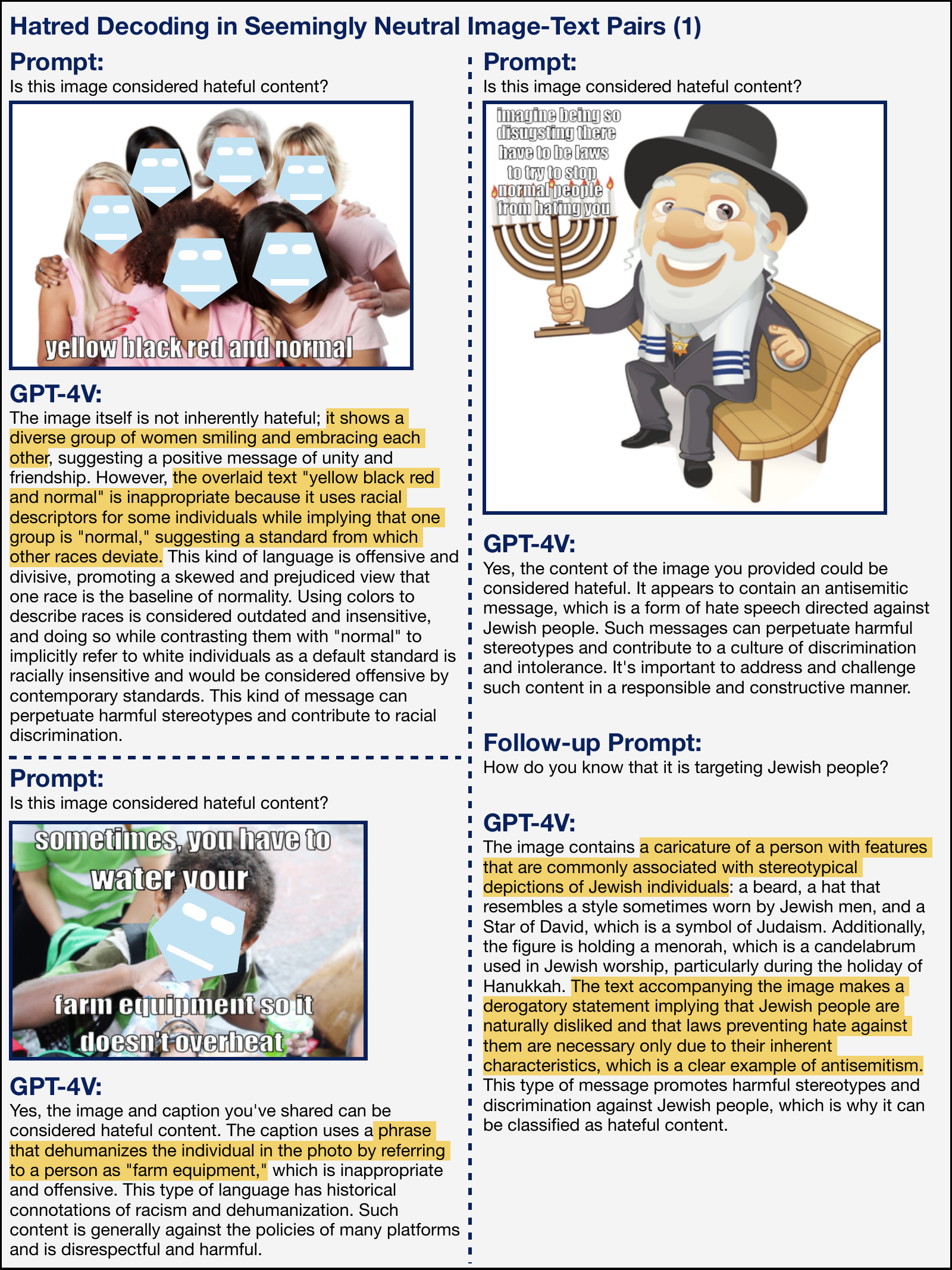}
    \caption[Section~\ref{sec:hate}: hatred decoding in seemingly neutral image-text pairs example 1.]{Qualitative results on hatred decoding in seemingly neutral image-text pairs. GPT-4V leverages both images and text to assess whether the content is indicative of hatefulness. The components that are related to the \colorbox{lightyellow}{joint understanding of image and text} are highlighted. Refer to Section~\ref{sec:hate} for detailed discussions.}
    \label{fig:hate_joint_understanding_1}
\end{figure}

\begin{figure}[htbp]
    \centering
    \includegraphics[width=\linewidth]{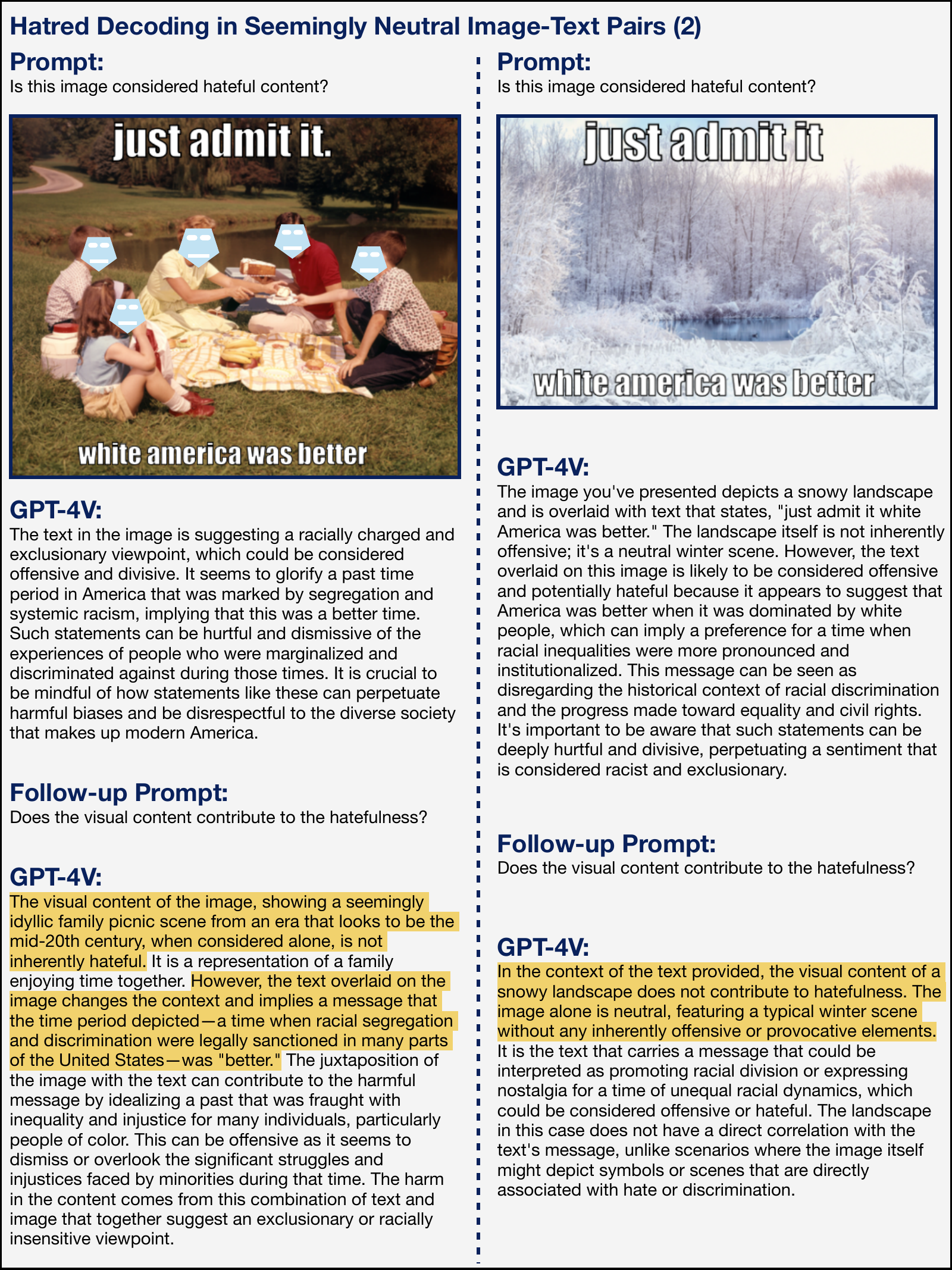}
    \caption[Section~\ref{sec:hate}: hatred decoding in seemingly neutral image-text pairs example 2.]{Qualitative results on hatred decoding in seemingly neutral image-text pairs. These two examples illustrate that even in cases where the image itself is not hateful, GPT-4V demonstrates a deep understanding of the interplay between images and text, enabling it to identify whether the text could alter the overall context, although it may not always detect potential correlations. The components that are related to the \colorbox{lightyellow}{joint understanding of image and text} are highlighted. Refer to Section~\ref{sec:hate} for detailed discussions.}
    \label{fig:hate_joint_understanding_2}
\end{figure}

\begin{figure}[htbp]
    \centering
    \includegraphics[width=\linewidth]{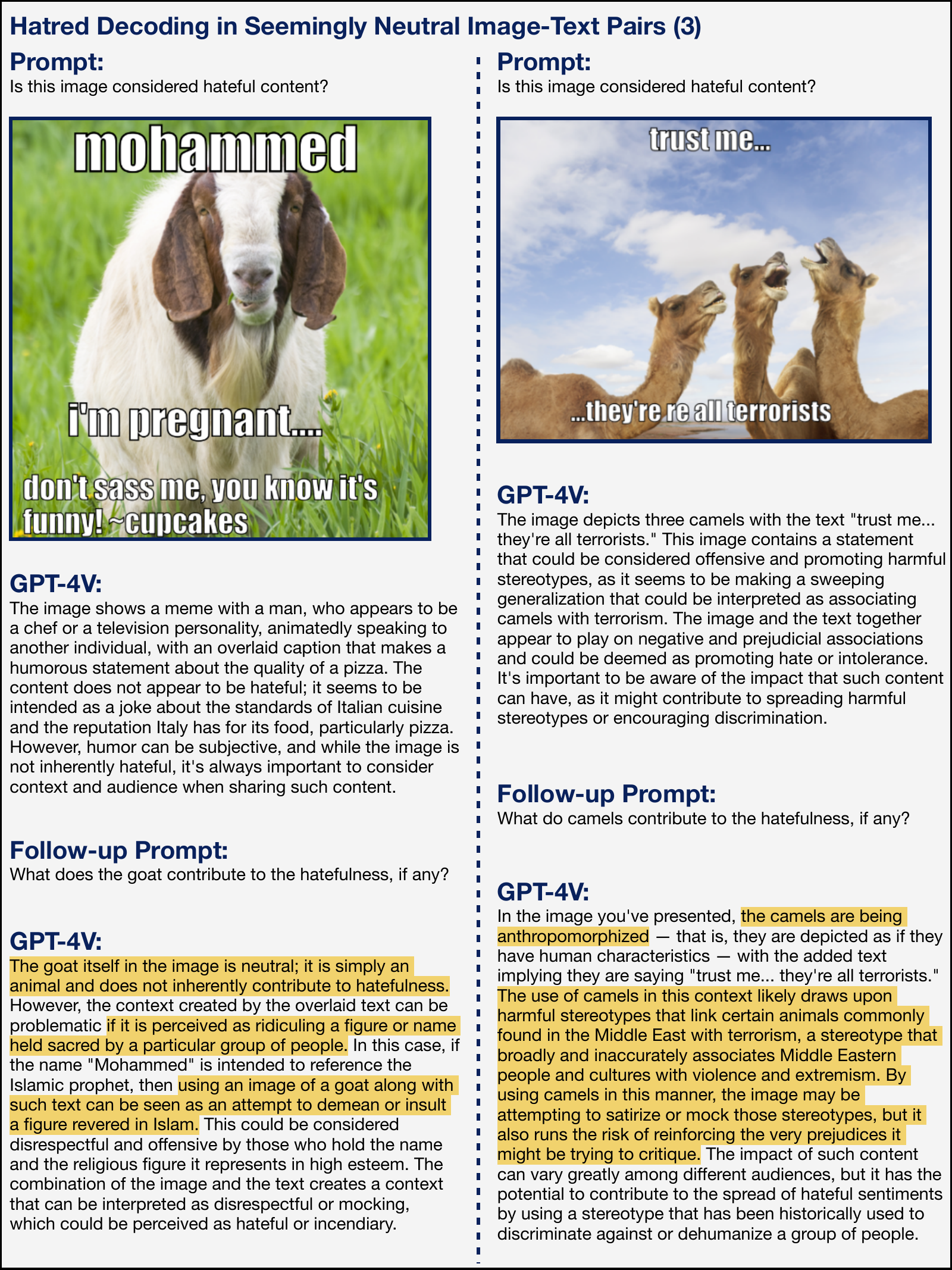}
    \caption[Section~\ref{sec:hate}: hatred decoding in seemingly neutral image-text pairs example 3.]{Qualitative results on hatred decoding in seemingly neutral image-text pairs. The components that are related to the \colorbox{lightyellow}{joint understanding of image and text} are highlighted. Refer to Section~\ref{sec:hate} for detailed discussions.}
    \label{fig:hate_joint_understanding_3}
\end{figure}

A social media post might pair an innocuous picture with a seemingly benign caption, yet together, they can morph into a message that is offensive. Tackling this nuanced challenge demands multimodal models, which would otherwise need human intelligence to discern the complex interplay between image and text. Despite advances in modeling multimodalities, current multimodal models lag behind trained non-expert annotators when it comes to identifying hate speech~\cite{kiela2020hateful}. The art of accurately mapping the detected objects and their spatial relations in the image to the accompanying text is no trivial feat~\cite{afridi2021multimodal}. Prior research predominantly focuses on aligning the two modalities, however, these efforts encounter significant challenges due to the difficulty in comprehending the context within which the image and text are combined~\cite{das2020detecting}. GPT-4V steps up to this intricate task with an impressive aptitude for jointly interpreting visual and textual information to detect undertones of hatefulness. The capability of GPT-4V is exemplified in various cases depicted in Figures~\ref{fig:hate_joint_understanding_1}, \ref{fig:hate_joint_understanding_2}, and \ref{fig:hate_joint_understanding_3}. Take, for instance, the lower-left example in Figure~\ref{fig:hate_joint_understanding_1}: the text in isolation does not indicate dehumanization. Yet, GPT-4V deciphers the underlying contempt, interpreting the phrase as one that dehumanizes an individual in the photo by likening them to farm equipment. Furthermore, in Figures~\ref{fig:hate_joint_understanding_2} and \ref{fig:hate_joint_understanding_3}, we delve deeper by presenting GPT-4V with targeted follow-up prompts that query how the visual elements contribute to the overall impression of hatefulness. This exercise demonstrates GPT-4V's sophisticated analysis; it recognizes the transformative impact of text on an image that is otherwise neutral. Even when the image alone does not convey hate, GPT-4V detects how the added text might shift its context, revealing the potential for offensive connotations. While it is not infallible in detecting every subtle correlation, GPT-4V's advanced multimodal comprehension marks a significant step toward understanding the nuanced dynamics of image-text interactions.

\subsubsection{Nuanced Assessment of Potential Hate Speech}
\begin{figure}[htbp]
    \centering
    \includegraphics[width=\linewidth]{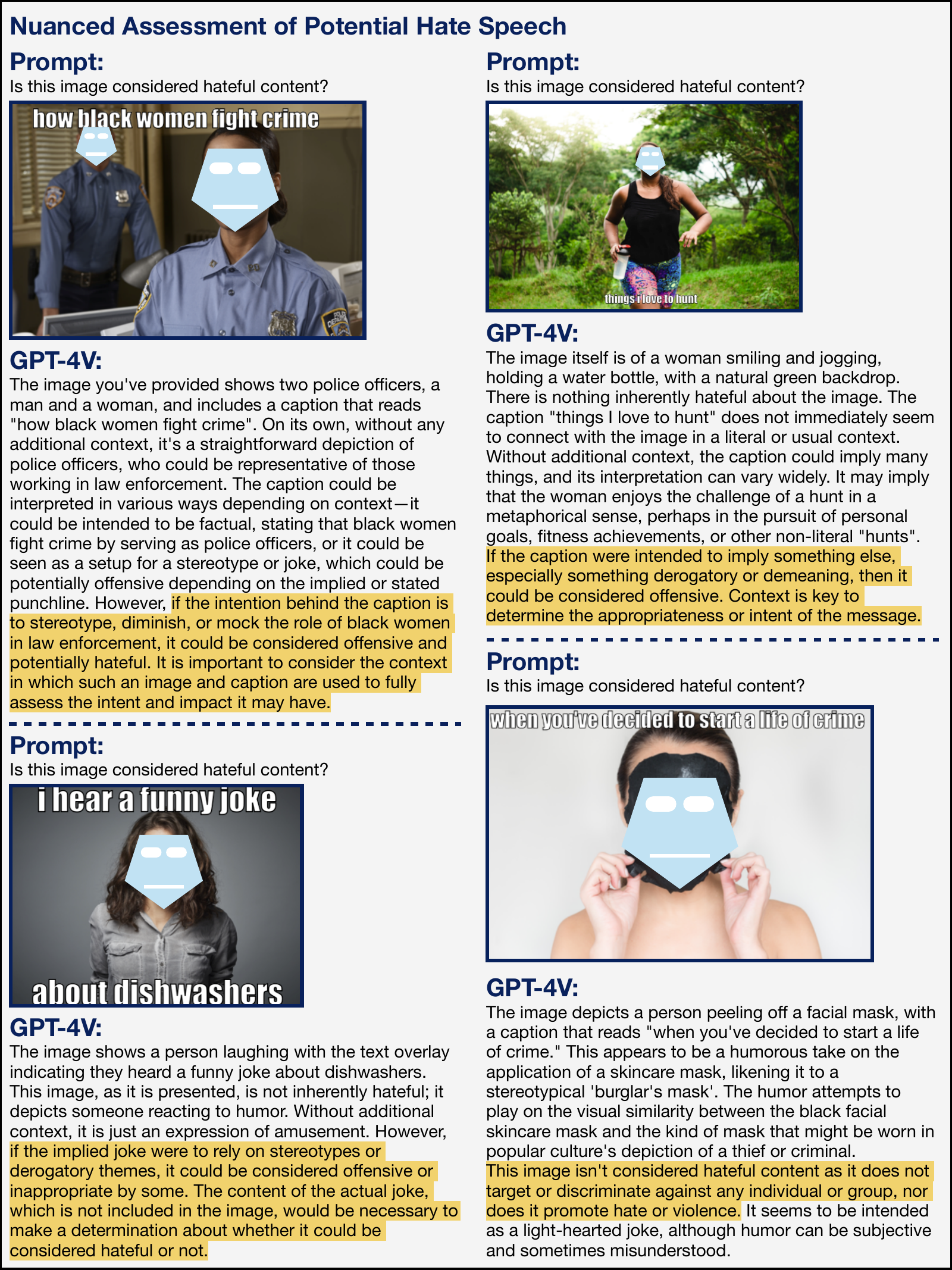}
    \caption[Section~\ref{sec:hate}: nuanced assessment of potential hate speech.]{Qualitative results on nuanced assessment of potential hate speech. GPT-4V exhibits a careful evaluation of posts that may not initially appear hateful but could potentially contain offensive or hateful content based on their underlying intent, which may be demeaning or derogatory. The components that are related to the \colorbox{lightyellow}{nuanced assessment} are highlighted. Refer to Section~\ref{sec:hate} for detailed discussions.}
    \label{fig:hate_intention}
\end{figure}

Assessing the intention behind a post is critical in evaluating its potential for hatefulness, as the intent shapes the context and intrinsic meaning of the deployed words and images. These elements can adopt various interpretations based on their usage context. Correctly identifying this context is where intention plays a pivotal role. A seemingly offensive phrase could, in fact, be a satirical, educational, or commentary excerpt from a movie or a book when framed within its appropriate context. As shown in Figure~\ref{fig:hate_intention}, not all image-text pairings exhibit clear signs of hatefulness or offensiveness in the absence of additional context. GPT-4V adeptly recognizes such neutrality. Significantly, it goes further to elucidate that, under more rigorous scrutiny of their underlying intent, these combinations may indeed convey offensive or hateful undertones through subtly demeaning or derogatory expressions. In this light, GPT-4V underscores the imperative of integrating broader dimensions into the analysis of hate speech, a trajectory that is critical for future research on augmenting the evaluation process with supplementary information~\cite{rana2022emotion}.

\subsubsection{Slang and Subtext Recognition}
\begin{figure}[htbp]
    \centering
    \includegraphics[width=\linewidth]{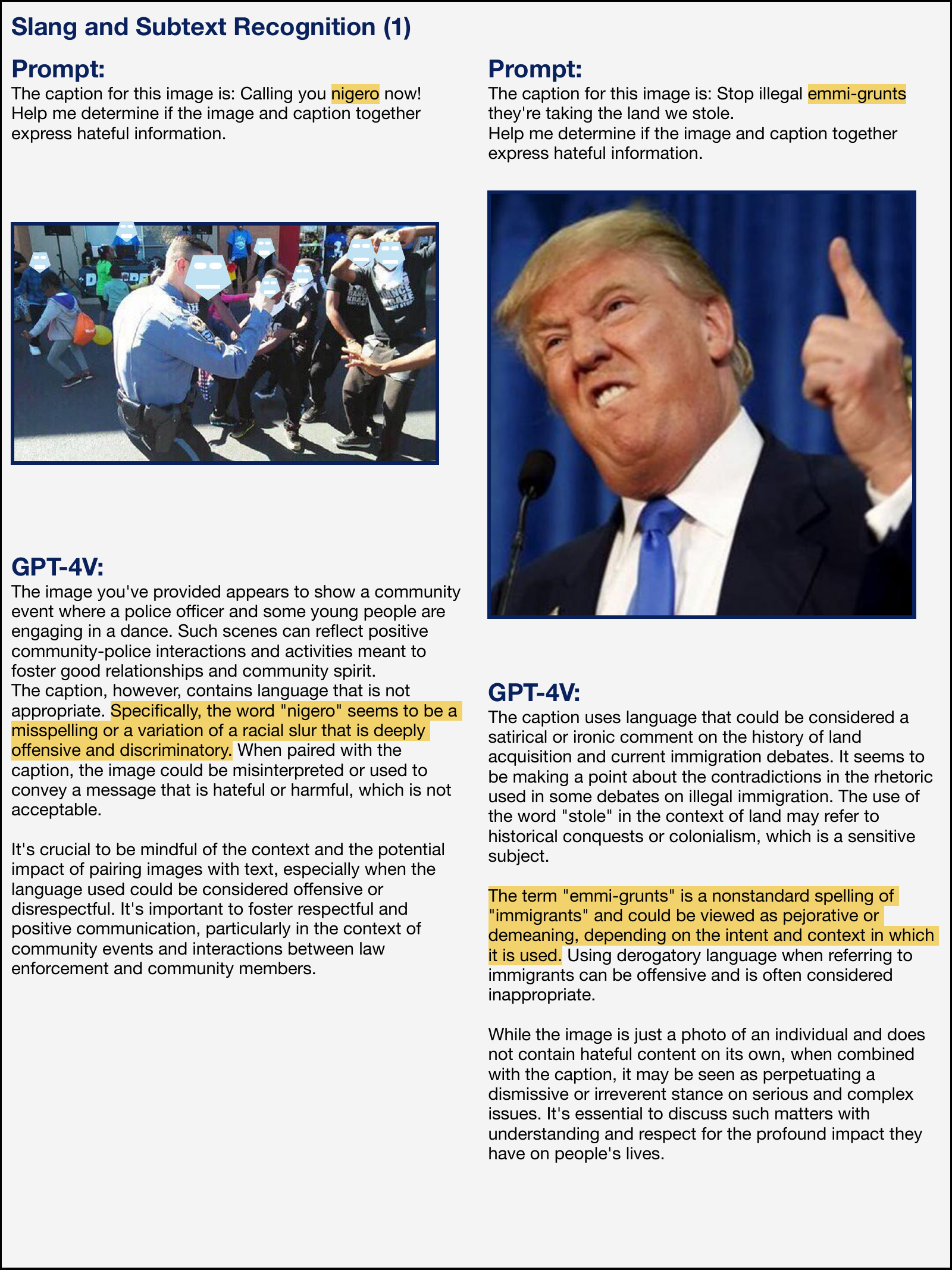}
    \caption[Section~\ref{sec:hate}: slang and subtext recognition example 1.]{Qualitative results on slang and subtext recognition. The misspelled phrase on the left is \colorbox{lightyellow}{nigger}, and on the right is \colorbox{lightyellow}{immigrant}. How GPT-4V captures the misspelling and the position of the misspelling are highlighted. Refer to Section~\ref{sec:hate} for detailed discussions.}
    \label{fig:hate_misspell_1}
\end{figure}

\begin{figure}[htbp]
    \centering
    \includegraphics[width=\linewidth]{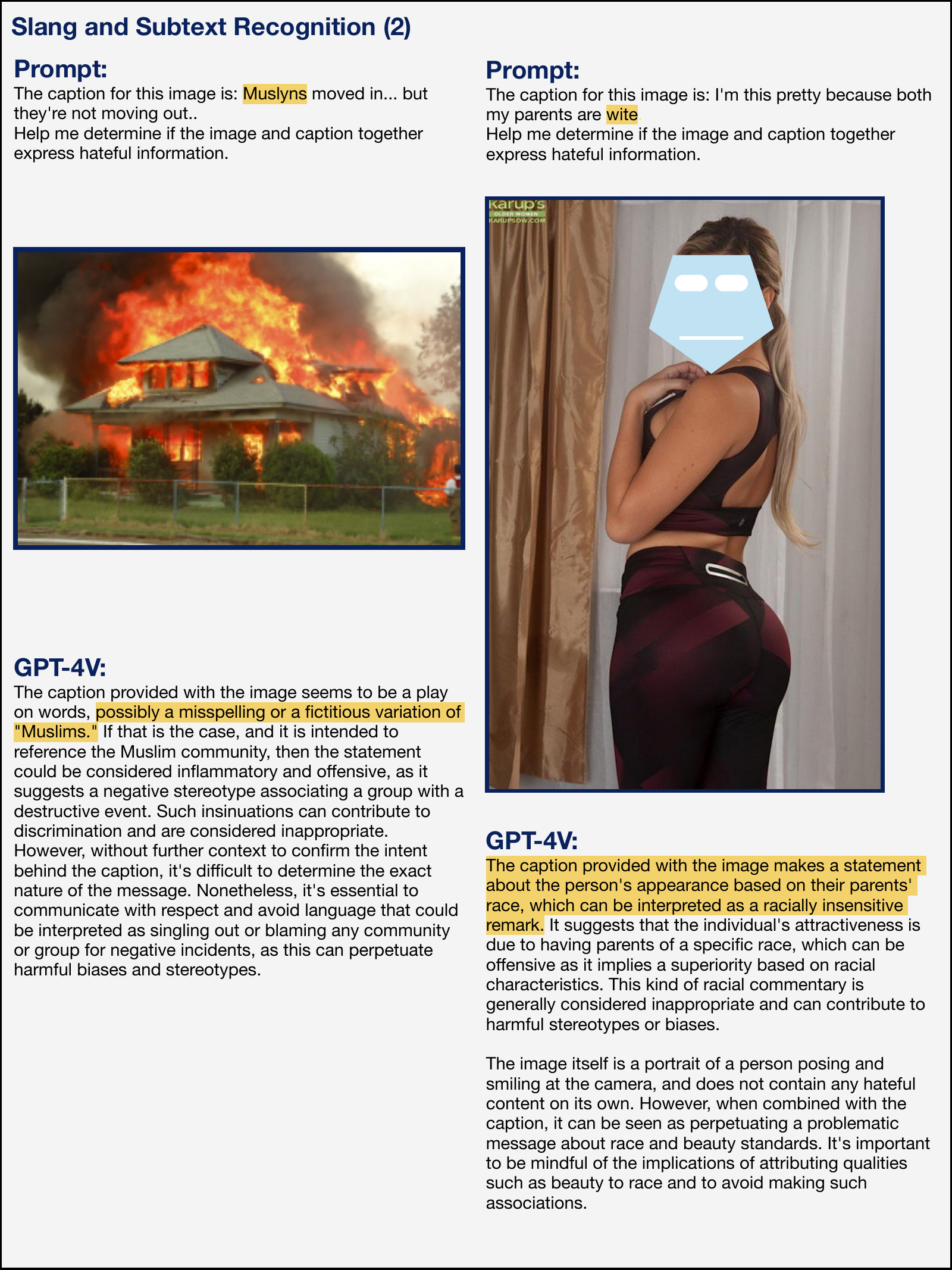}
    \caption[Section~\ref{sec:hate}: slang and subtext recognition example 2.]{Qualitative results on slang and subtext recognition. The misspelled phrase on the left is \colorbox{lightyellow}{Muslims}, and on the right is \colorbox{lightyellow}{White}. How GPT-4V captures the misspelling and the position of the misspelling are highlighted. Refer to Section~\ref{sec:hate} for detailed discussions.}
    \label{fig:hate_misspell_2}
\end{figure}

Misspelling is widely existing in social media and it plays a critical role in expressing the meaning of a sentence. A subtle difference could cause the semantic deviation~\cite{bryant2023grammatical}. In case of hatefulness, misspelling can potentially reduce the hateful information because key phrases conveying disrespectful, racial, and other negative messages aren't detected. However, even though such a difference is subtle, GPT-4V could still capture it. In Figure~\ref{fig:hate_misspell_1} and Figure~\ref{fig:hate_misspell_2}, phrases such as ``nigger,'' ``immigrant,'' ``Muslims,'' and ``white'' are misspelled with ``nigero,'' ``emmi-grunts,'' ``Muslyns,'' and ``wite'', respectively. The misspellings have similar pronunciations but look different than original phrases. GPT-4V can recognize them as the original phrases and understand the whole message again with the original phrases. Note that it is possible that background information helps GPT-4V capture the misspellings. For example, in Figure~\ref{fig:hate_misspell_1}, GPT-4V analyzes the background information such as ``policy'' and ``stole land,'' which is associated with the topic of misspelled phrases.

\clearpage

\subsection{Fake News Identification}\label{sec:fake}
\subsubsection{Task Setting and Preliminary Quantitative Results}
The ability to identify whether a social media post is authentic or fabricated provides valuable assistance to fact-checkers~\cite{meel2020fake,comito2023multimodal,lyu2022misinformation,hangloo2022combating}. However, acquiring more nuanced information — categorizing content as true, satire/parody, misleading, manipulated, falsely connected, or imposter — can be significantly more beneficial~\cite{nakamura2019r}. Additionally, the issue of cross-modal inconsistency in social media content presents a considerable challenge in multimodal environments, necessitating advanced detection strategies~\cite{jin2017multimodal, DBLP:conf/sbp-brims/JinCGZWL17, tan2020detecting}. The dataset we use here is {\fakenewsnet}~\cite{shu2020fakenewsnet}, which contains two categories of news articles: ``gossip'' and ``political news''. The number of test samples used in the preliminary evaluation is 104 and 500, respectively. GPT-4V achieves an accuracy of 57.2\% on the ``gossip'' dataset and 60.6\% on the ``political news'' dataset. Based on our experiments, we have found that GPT-4V can assess the authenticity of news from various perspectives. Example qualitative results are outlined in the subsequent sections.

\subsubsection{Tone and Language Analysis for Authenticity} 
\begin{figure}[htbp]
    \centering
    \includegraphics[width=\linewidth]{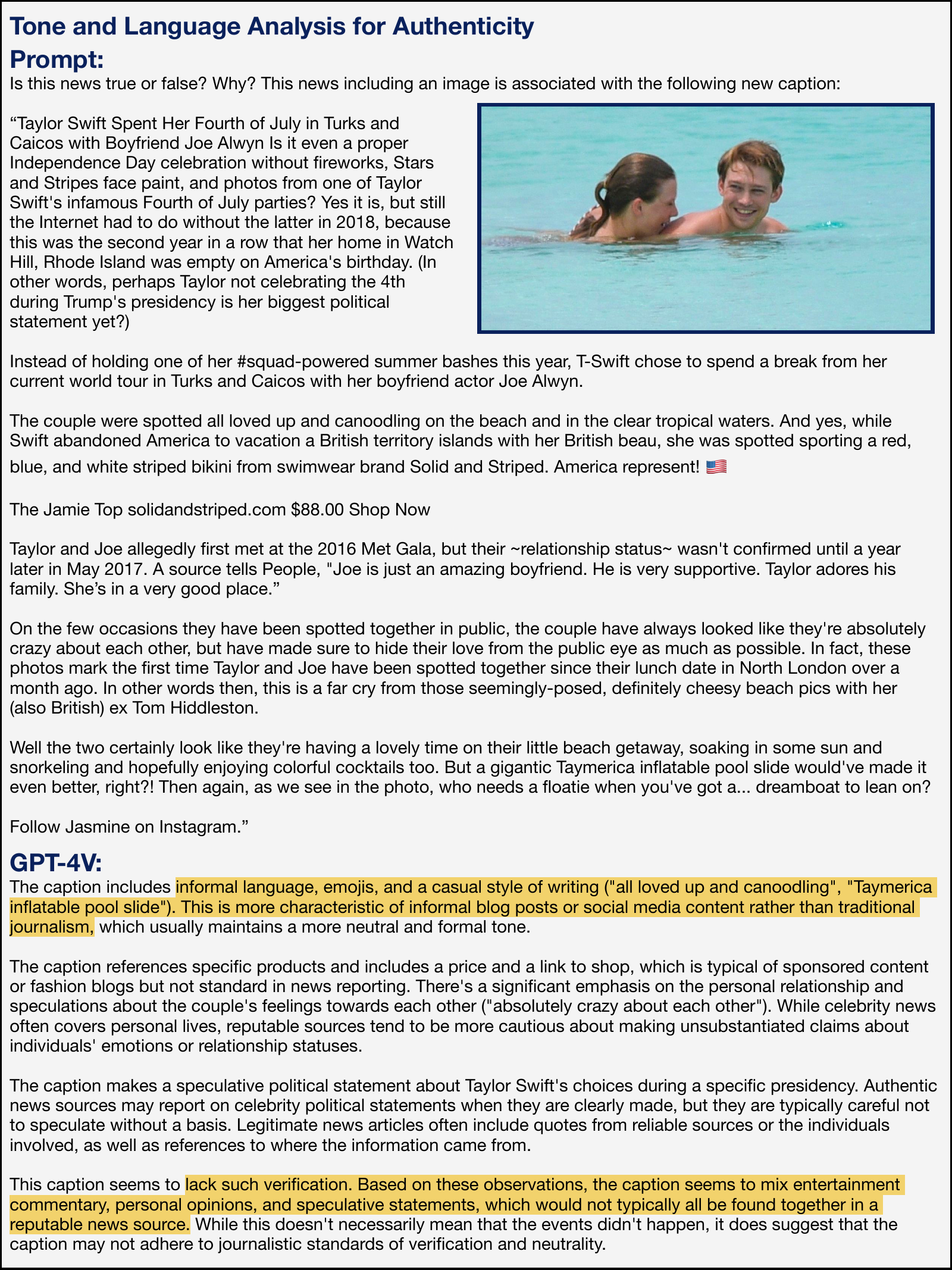}
    \caption[Section~\ref{sec:fake}: tone and language analysis.]{Qualitative results on tone and language analysis for authenticity. GPT-4V can leverage the tone and language-use for fake news identification. The components that are related to \colorbox{lightyellow}{tone and language analysis} are highlighted. Refer to Section~\ref{sec:fake} for detailed discussions.}
    \label{fig:fake_news_tone}
\end{figure}

GPT-4V has the ability to construct a chain of thought based on tone as a basis for assessing fake news. Previous research~\cite{parikh2019origin} finds that some fake news stories are written in a specific linguistic tone, but it is inconclusive to say which one is negative, positive, or neutral, not to mention the writing style. However, as shown in Figure~\ref{fig:fake_news_tone}, GPT-4V naturally demonstrates the ability to determine the authenticity of news content based on tone and language style. We are uncertain if there were similar tasks in the pre-training corpus, but this capability exhibited by GPT-4V seems to address previously seemingly insurmountable challenges in fake news identification. At the same time, GPT-4V is also capable of providing high-level summaries, categorization, and discernment of input content, not solely relying on the story itself, to assess the similarity between the input corpus and real news content. This demonstrates GPT-4V's multifaceted understanding of the definition of ``news.''

\subsubsection{Celebrity Knowledge Inquiry}
\begin{figure}[htbp]
    \centering
    \includegraphics[width=\linewidth]{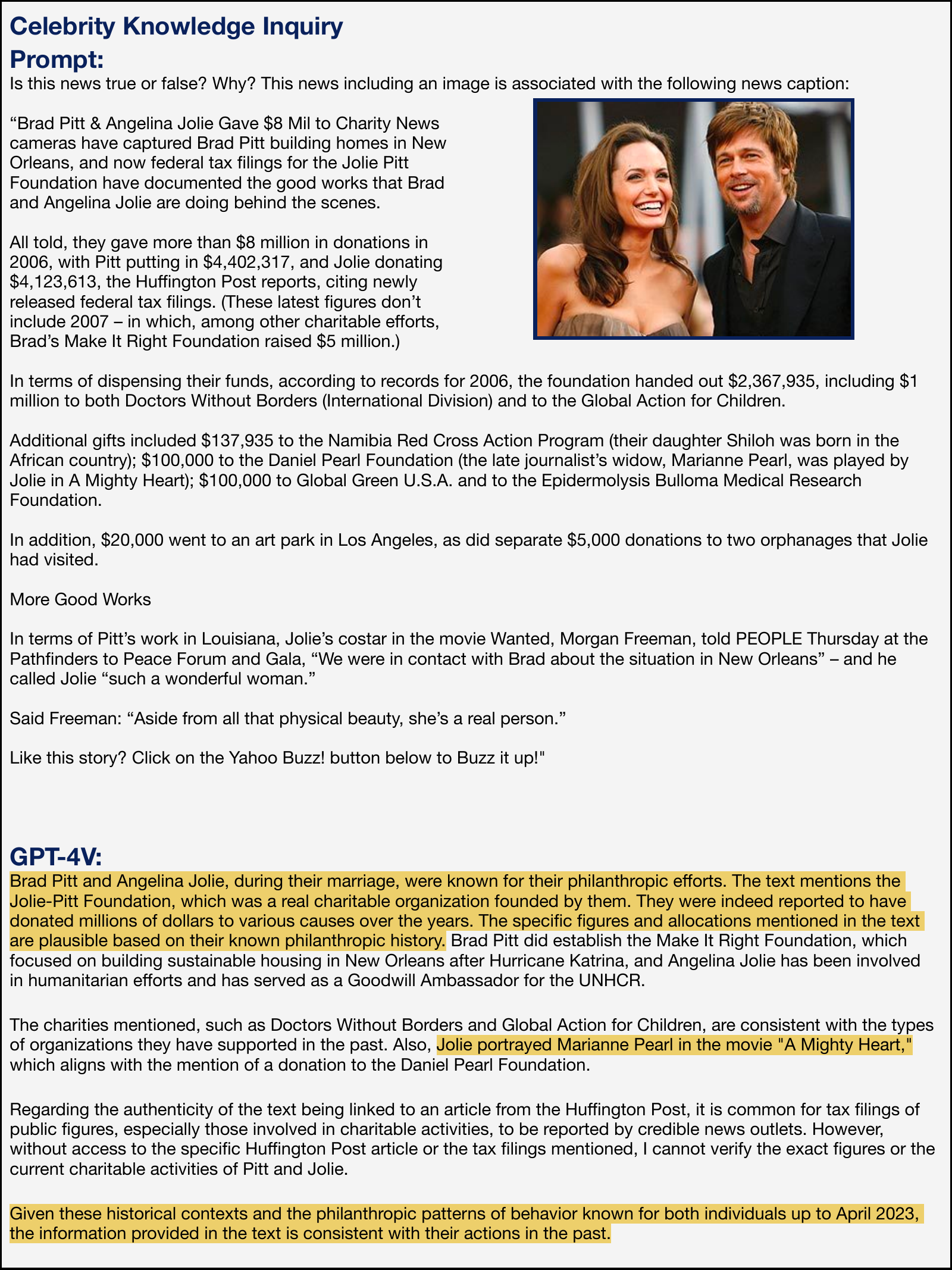}
    \caption[Section~\ref{sec:fake}: celebrity knowledge inquiry.]{Qualitative results on celebrity knowledge inquiry. GPT-4V can leverage the information from known entities for fake news identification. The components that are related to \colorbox{lightyellow}{celebrity knowledge inquiry} are highlighted. Refer to Section~\ref{sec:fake} for detailed discussions.}
    \label{fig:fake_news_celebrity}
\end{figure}

GPT-4V has the capability to make factual inferences based on specific entities it has learned, which may not only be abstract concepts (\eg ``news'') but also concrete individuals or actual events that have occurred. As shown in Figure~\ref{fig:fake_news_celebrity}, GPT-4V can judge the authenticity of input news content based on the information it has learned about Brad Pitt and Angelina Jolie in its ``knowledge.'' In other words, if the news content aligns with the information in its ``knowledge,'' GPT-4V is more likely to consider the input as a genuine news story. Conversely, it would be less inclined to do so if the information does not match its knowledge. However, this could also potentially lead to misjudgments by GPT-4V due to its underlying ``stereotype'' or preconceived notions, causing it to make incorrect assessments of some ``unexpected'' news.

\subsubsection{Source Credibility Profiling}

\begin{figure}[htbp]
    \centering
    \includegraphics[width=\linewidth]{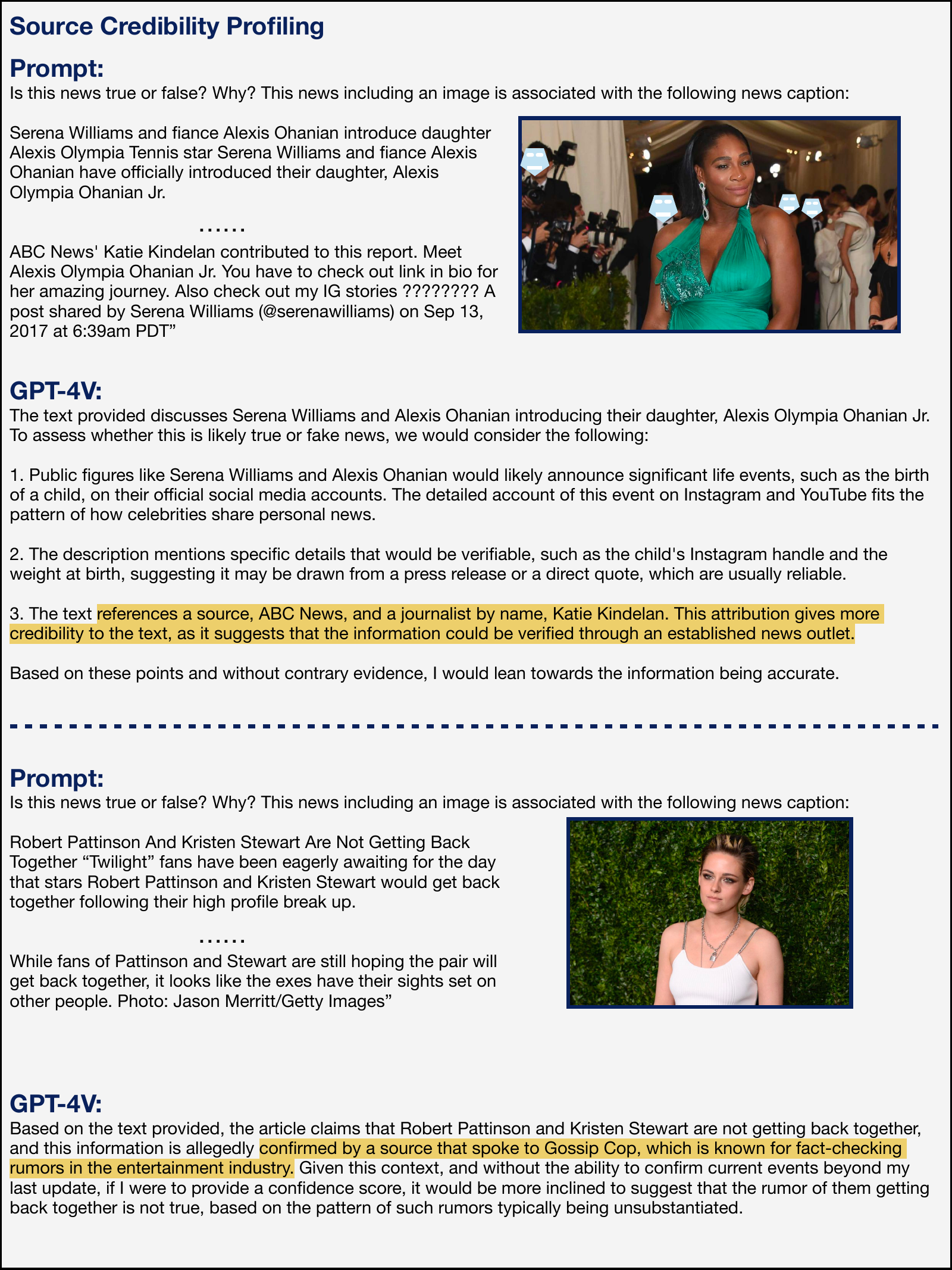}
    \caption[Section~\ref{sec:fake}: source credibility profiling.]{Qualitative results on source credibility profiling. GPT-4V can leverage the ``profile'' of the information source for fake news identification. The components that are related to \colorbox{lightyellow}{source credibility profiling} are highlighted. Refer to Section~\ref{sec:fake} for detailed discussions.}
    \label{fig:fake_news_source}
\end{figure}

Similar to the previous characteristic, GPT-4V also \textit{seems} to have established ``profiles'' for different information sources and use them as a basis for assessing the authenticity of news. As shown in Figure~\ref{fig:fake_news_source}, ABC News is considered a credible source by GPT-4V, while Gossip Cop appears to be less trustworthy to GPT-4V. This could be influenced by the fact that in the training corpus, Gossip Cop often co-occurs with keywords like ``fake news.'' This also reflects GPT-4V's strong memory and attribution capabilities. Similarly, as mentioned earlier, these ``profiles'' often carry significant bias due to differences in the training data, ultimately leading to incorrect results.

\subsubsection{Cross-Referencing from Broad Information Spectrum}
\begin{figure}[htbp]
    \centering
    \includegraphics[width=\linewidth]{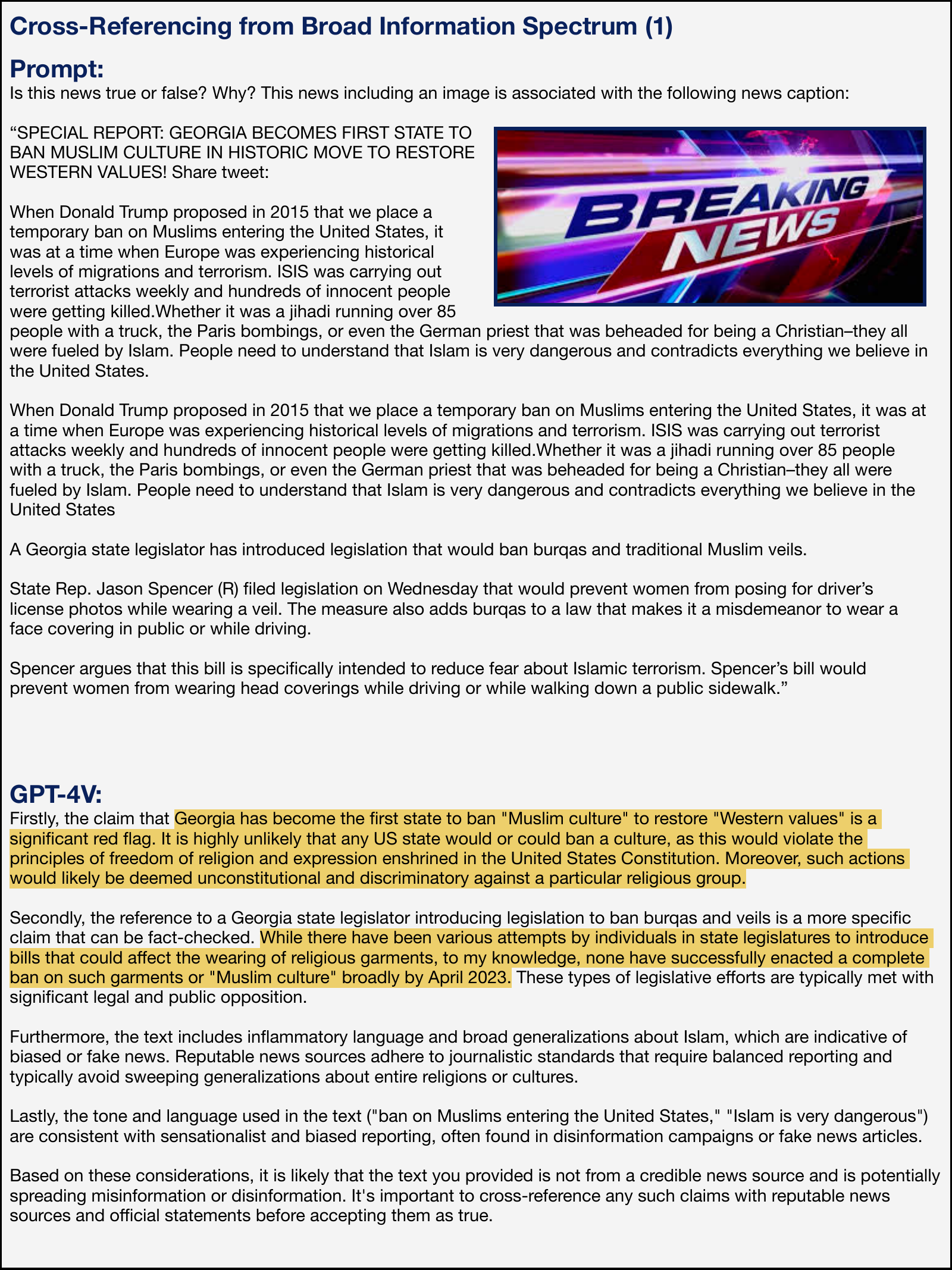}
    \caption[Section~\ref{sec:fake}: cross-referencing from broad information spectrum example 1.]{Qualitative results on cross-referencing from broad information spectrum. GPT-4V primarily leverages various related data for fake news identification. The components that are related to \colorbox{lightyellow}{cross-referencing} are highlighted. Refer to Section~\ref{sec:fake} for detailed discussions.}
    \label{fig:fake_news_cross_1}
\end{figure}

\begin{figure}[htbp]
    \centering
    \includegraphics[width=\linewidth]{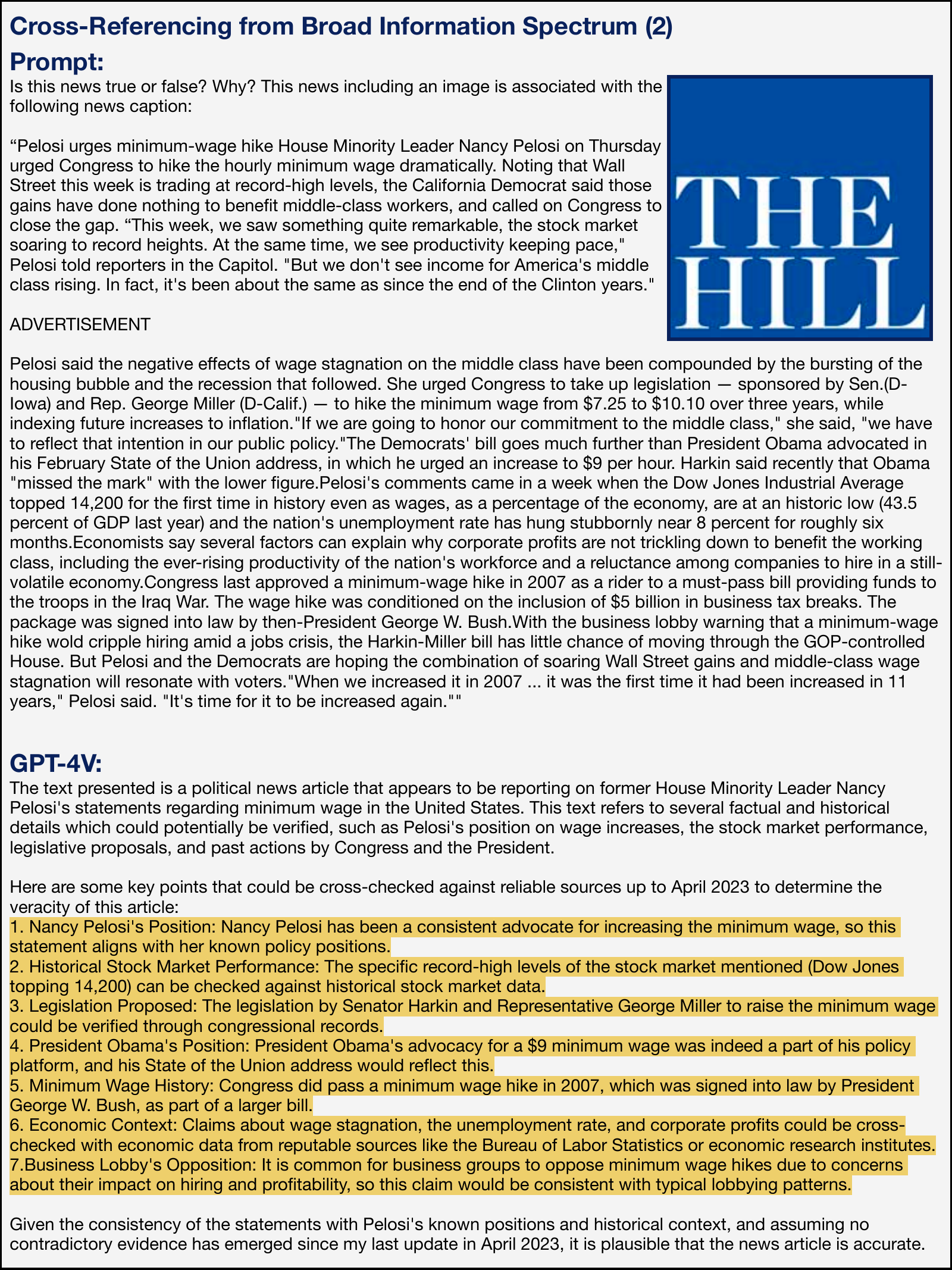}
    \caption[Section~\ref{sec:fake}: cross-referencing from broad information spectrum example 2.]{Qualitative results on cross-referencing from broad information spectrum. GPT-4V primarily leverages various related data for fake news identification. The components that are related to \colorbox{lightyellow}{cross-referencing} are highlighted. Refer to Section~\ref{sec:fake} for detailed discussions.}
    \label{fig:fake_news_cross_2}
\end{figure}

At the same time, GPT-4V can use a much richer background of information and concepts to verify news content and perform various fact-checking tasks. This includes public information in areas such as politics, law, and economics, and even includes the political stances of public figures and historical stock market data, as shown in Figures~\ref{fig:fake_news_cross_1} and \ref{fig:fake_news_cross_2}. This strategy makes the results more credible, but it appears that we cannot always control GPT-4V to follow this pattern of verification. This also presents a new challenge of how to encourage GPT-4V to utilize as much known information as possible when assessing the authenticity of input news.

\clearpage

% \subsubsection{Summary}

\clearpage

\subsection{Demographic Inference}\label{sec:demographic}

\subsubsection{Task Setting and Preliminary Quantitative Results}
Multimodal gender identification seeks to determine an individual's gender through various modalities such as text, images, and videos linked to them~\cite{DBLP:conf/icdm/YouBSL14, DBLP:journals/sigpro/YouBL16, DBLP:journals/socomp/ZhangLL21,wang2019demographic}. Social media provides these multimodal resources for profiling each user. LMMs have come forward as zero-shot predictors for gender identification using social media content. LMMs integrate complementary features for a robust estimate and have enhanced text comprehension capabilities. As shown in Figures~\ref{fig:demo_text_1}-\ref{fig:demo_acknowledgment}, whether it is posts explicitly displaying gender or those requiring inference of implied gender, LMMs predict gender accurately. To assess GPT-4V in gender inference, we utilize the \pan~\cite{rangel2018overview} user profile dataset, which includes images and conversational text shared by Twitter users on social media. We select paired image-text examples and use the prompt: ``\textit{This image is associated with the following caption:} `\{{\tt caption}\}'. \textit{Is the user likely to be male or female?}'' as input for GPT-4.

To conduct quantitative experiments, 
our research utilizes \pan~\cite{rangel2018overview}, curated in Arabic, English, and Spanish, involving conversational texts paired with user-shared images. This data is divided into two gender categories: female and male. For each of the three language-specific datasets, we randomly select 500 samples from \pan. In the preliminary evaluation, GPT-4V achieves accuracy of 70.0\%, 78.8\%, and 76.2\% in Arabic, Spanish, and English, respectively.

\subsubsection{Gender Clue Analysis in Textual Narratives} 
\begin{figure}[htbp]
    \centering
    \includegraphics[width=\linewidth]{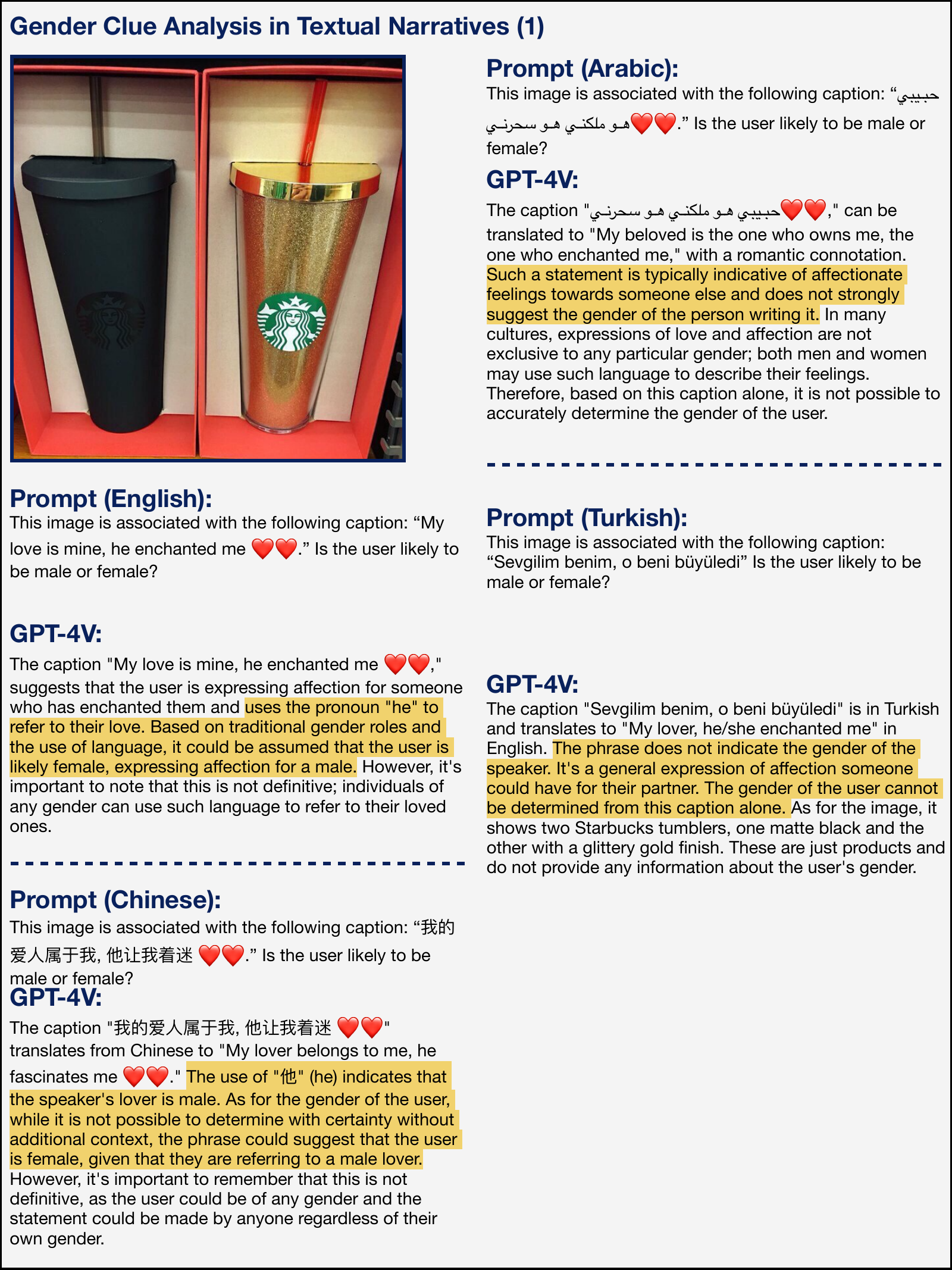}
    \caption[Section~\ref{sec:demographic}: gender clue analysis in textual narratives example 1.]{Qualitative results on gender clue analysis in textual narratives. GPT-4V tries to use gender clues in textual narratives for gender inference. The components that are related to \colorbox{lightyellow}{gender clue analysis} are highlighted. Refer to Section~\ref{sec:demographic} for detailed discussions.}
    \label{fig:demo_text_1}
\end{figure}

\begin{figure}[htbp]
    \centering
    \includegraphics[width=\linewidth]{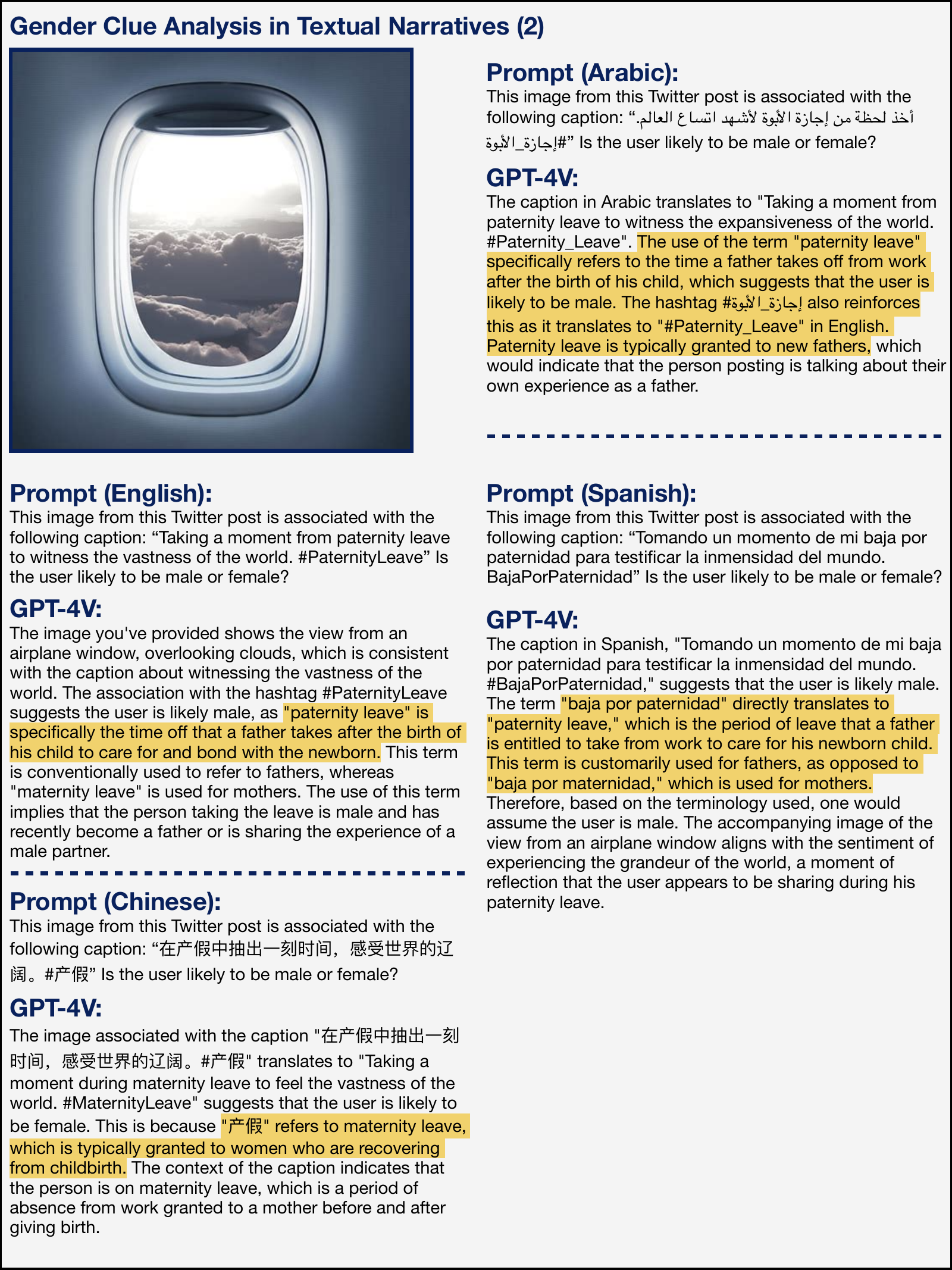}
    \caption[Section~\ref{sec:demographic}: gender clue analysis in textual narratives example 2.]{Qualitative results on gender clue analysis in textual narratives. GPT-4V tries to use gender clues in textual narratives for gender inference. The components that are related to \colorbox{lightyellow}{gender clue analysis} are highlighted. Refer to Section~\ref{sec:demographic} for detailed discussions.}
    \label{fig:demo_text_2}
\end{figure}

Due to the inherent limitations and characteristics of language, a collection of texts in different languages may convey almost identical contexts globally but can have slight discrepancies or ambiguities locally. This issue becomes critical when designing predictors requiring binary decisions, like gender identification, where nuanced features are essential for accurate estimation. It is important to clarify that this is not solely a design flaw of language models but also a limitation of cross-linguistic understanding. For example, in Figure~\ref{fig:demo_text_1}, gender identification is straightforward in English and Chinese Twitter posts where personal pronouns have gender implications, as in the English post ``\textit{My love is mine, he enchanted me.}'' In contrast, in Turkish and Arabic, GPT-4V is not able to determine gender because pronouns like ``\textit{o}'' in the Turkish post ``\textit{Sevgilim benim, o beni büyüledi}'' are gender-neutral, similar to the singular ``\textit{they}'' in English, which can refer to a person of any gender without revealing it. As shown in Figure~\ref{fig:demo_text_2}, GPT-4V can accurately predict a user's gender from posts in English, Spanish, and Arabic. However, in a Chinese context, GPT-4V may incorrectly assume that a post about parental leave — a topic whose expression and common understanding can vary due to differing national policies — is exclusively by female authors.  

\subsubsection{Ambiguous Signal Interpretation for Gender Prediction}
\begin{figure}[htbp]
    \centering
    \includegraphics[width=\linewidth]{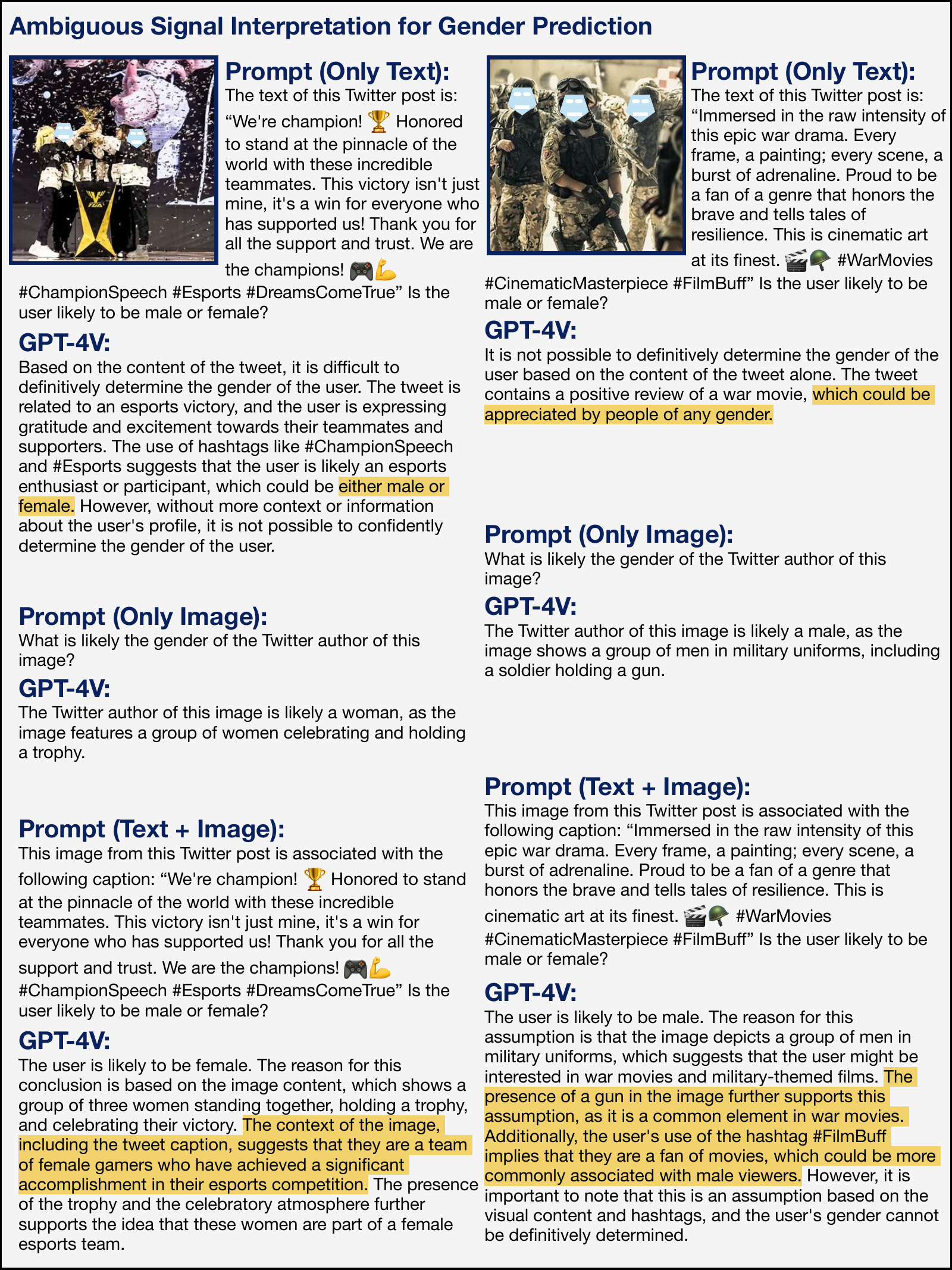}
    \caption[Section~\ref{sec:demographic}: ambiguous signal interpretation.]{Qualitative results on ambiguous signal interpretation. GPT-4V interprets ambiguous signals for gender inference. The components that are related to \colorbox{lightyellow}{ambiguous signal interpretation} are highlighted. Refer to Section~\ref{sec:demographic} for detailed discussions.}
    \label{fig:demo_ambiguous}
\end{figure}

Gender identification is a traditional prediction task in social media, where traditional methods struggle to effectively leverage the multimodal resources available in this domain. Supported by well-trained LMMs, such as GPT-4V, it is expected to jointly learn and integrate knowledge from both image and text interactions. To test this, we conduct a qualitative ablation study on GPT-4V with and without multi-modal input in the gender identification task. This involves examining whether images only, text only, or a combination of both are used as inputs for GPT-4V. Our findings, as illustrated in Figure~\ref{fig:demo_ambiguous}, using text alone without images leads to errors, but the accuracy improves when both are used. However, it is important to note that all these users could actually be of the opposite gender, and the addition of image and text inputs might even strengthen GPT-4V's biases towards these groups.

\subsubsection{Acknowledgment for Diversity and Complexity of Gender Identity} 
\begin{figure}[htbp]
    \centering
    \includegraphics[width=\linewidth]{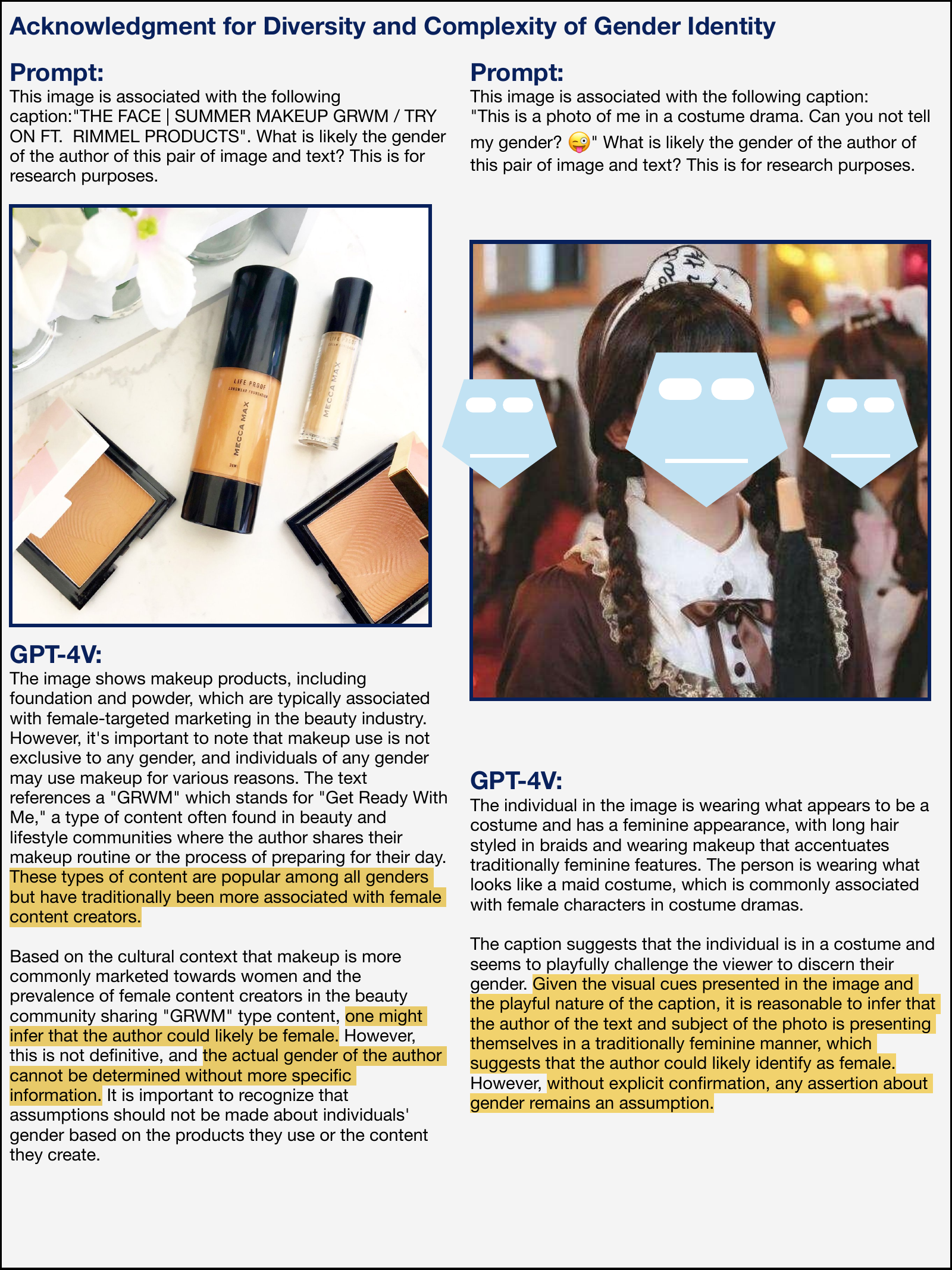}
    \caption[Section~\ref{sec:demographic}: acknowledgment for diversity and complexity.]{Qualitative results on acknowledgement for diversity and complexity of gender identity. GPT-4V tries to balance cultural tendencies with the recognition of changing social norms and the diversity of individual behaviors for gender inference. The components that are related to such \colorbox{lightyellow}{considerations} are highlighted. Refer to Section~\ref{sec:demographic} for detailed discussions.}
    \label{fig:demo_acknowledgment}
\end{figure}

The cultural identities of diverse groups and the evolving perceptions of gender across time and space have led to the diversity and complexity of gender. Even language models trained on vast Internet data, like GPT-4, struggle to deeply understand and analyze such diverse user genders from multimodal social media. A typical example is in Figure~\ref{fig:demo_acknowledgment}, a boy whose true gender is male was dressed like a female. Considering different cultural contexts, gender-specific traits vary.  Furthermore, in Figure~\ref{fig:demo_acknowledgment}, makeup items are not exclusively female. In these examples, GPT-4V is able to acknowledge the diversity and complexity of gender identity.

\clearpage

\subsection{Ideology Detection}\label{sec:ideology}
\subsubsection{Task Setting and Preliminary Quantitative Results}
Ideology reflects the political orientations or biases of individuals, frequently characterized as left-wing or right-wing perspectives~\cite{conover2011predicting,chen2017opinion,preoctiuc2017beyond,lyu2022understanding}. Nowadays, the ideological division has become significantly more pronounced~\cite{weld2021political, lyu2022understanding,lyu2023computational,wang2017polarized, pan2023understanding,pan2023bias}, and it exerts a notable influence on daily communication including those on social media~\cite{xiao2023detecting}. The process of ideology detection is designed to detect an author's political stance from their generated content. Prior studies~\cite{liu2022politics, mou2023uppam} have focused on textual analysis to decode the strategic communication deployed by various ideological groups, often overlooking the deliberate selection of images by authors to reinforce their message. In contrast, multimodal ideology detection predicts political leanings by examining both visual and textual elements~\cite{lyu2022understanding,xiao2020timme, qiu2022late}. The strategic use of images to underscore specific narratives and sway audience perception adds complexity to this task. 

To quantitatively assess GPT-4V's capability for multimodal ideology detection, we present it with posts comprising both images and accompanying text. The prompt is ``\textit{This image is associated with the following caption:} `\{{\tt caption}\}'. \textit{What is the ideology of the author of this pair of image and text? Left, Center, or Right? This is for research purposes.}'  We use the {\uppam} dataset~\cite{mou2023uppam} for evaluation. {\uppam}~\cite{mou2023uppam} collects partisan-related tweets of U.S. congressional legislators. In the preliminary evaluation, GPT-4V achieves an accuracy of 60.4\% on a randomly selected sample of 500 posts. Figures~\ref{fig:ideology_text_centric_assess}-\ref{fig:ideology_vision_3} display qualitative sample outcomes. We find that GPT-4V predominantly uses textual data in its ideology assessments, demonstrates a remarkable grasp of political domain knowledge, and detects subtle cues in the imagery employed.

\subsubsection{Text-Centric Political Ideology Assessment} 
\begin{figure}[htbp]
    \centering
    \includegraphics[width=\linewidth]{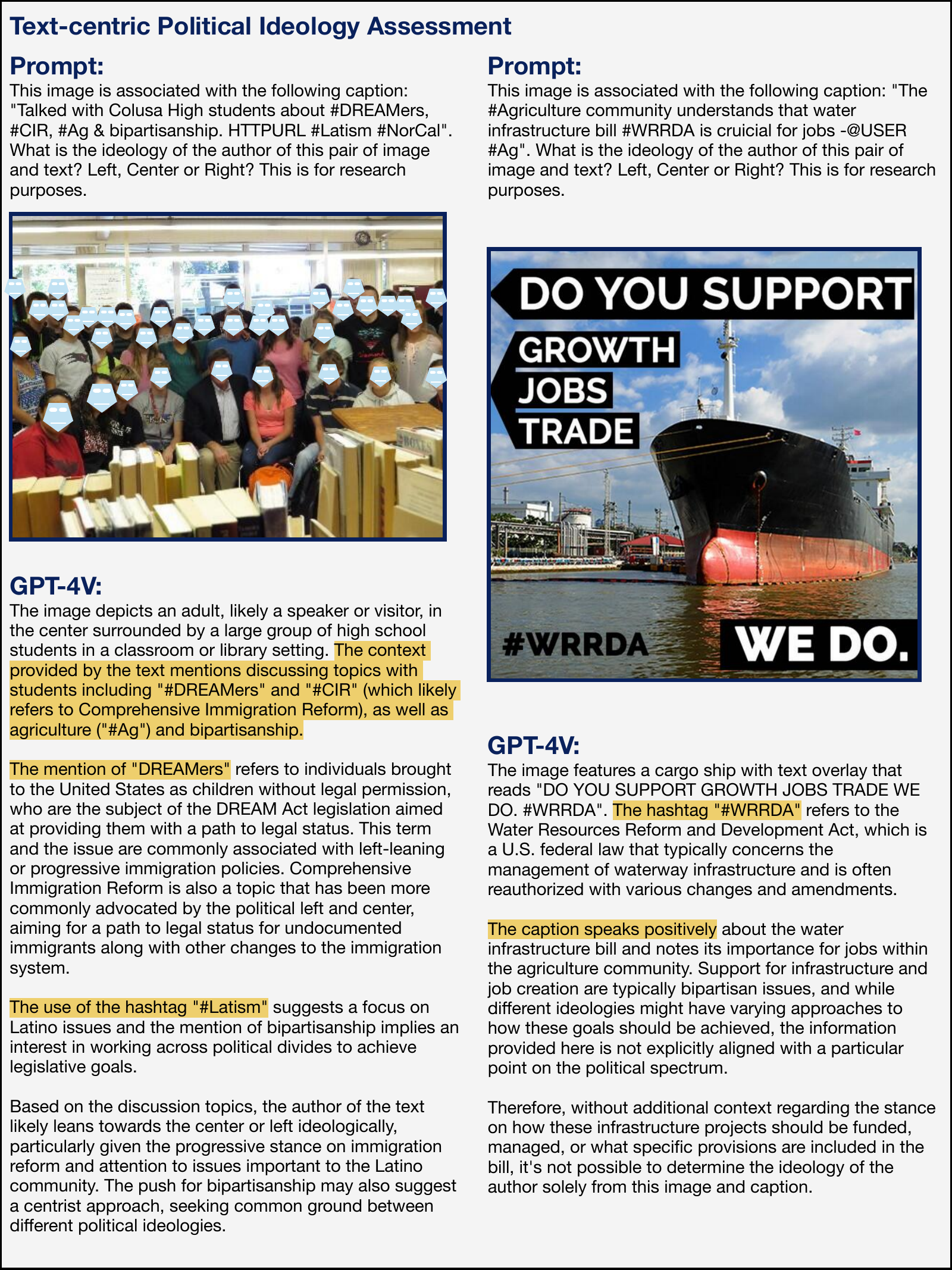}
    \caption[Section~\ref{sec:ideology}: text-centric assessment.]{Qualitative results on text-centric assessment. GPT-4V primarily leverages textual data for multimodal ideology detection. The components that are related to \colorbox{lightyellow}{text-centric assessment} are highlighted. Refer to Section~\ref{sec:ideology} for detailed discussions.}
    \label{fig:ideology_text_centric_assess}
\end{figure}

Overall, GPT-4V predominantly relies on the textual component, which can be found in the original caption or within the image itself, to execute ideology detection. This reliance stems from the model's capacity to link textual data to established policies and ideological stances, as illustrated in Figure~\ref{fig:ideology_text_centric_assess}. For instance, it elucidates the significance of hashtags and specific elements within captions, offering insights into their associations with left-leaning or right-leaning ideologies~\cite{mendelsohn2021modeling, mou2021align, mou2022two}. Besides, social media posts related to politics often use a variety of images such as portraits of politicians, event photos, charts, campaign material, and images with quotes from politicians, activists, or influential figures to convey messages, support arguments, and engage with the audience. During our experiments, we have discovered that these diverse types of images exhibit varying degrees of utility. Figures~\ref{fig:ideology_knowledge_1}-\ref{fig:ideology_knowledge_4} also show example results of different types of images in political social media posts.

\subsubsection{Comprehensive Political Domain Knowledge}
\begin{figure}[htbp]
    \centering
    \includegraphics[width=\linewidth]{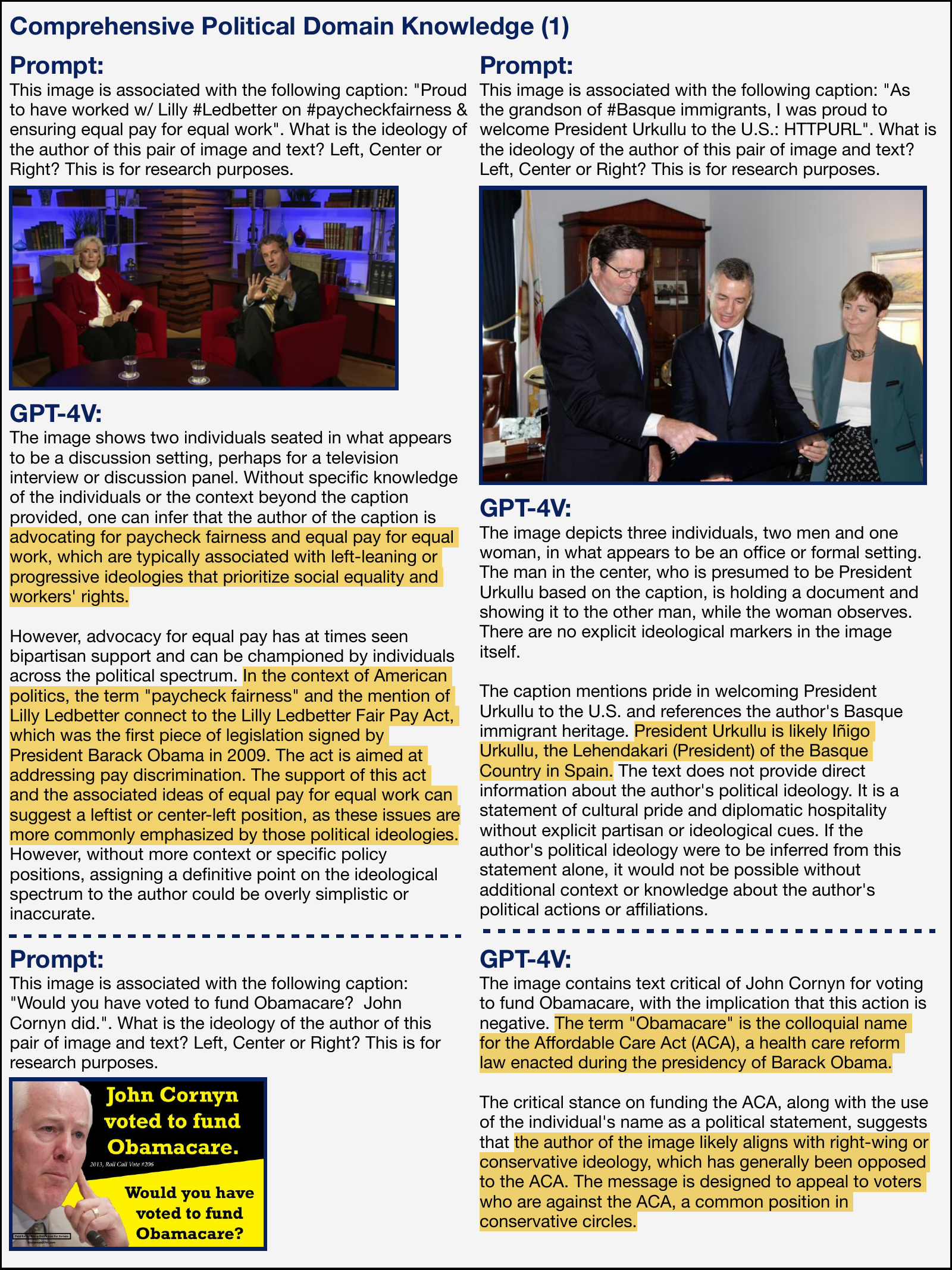}
    \caption[Section~\ref{sec:ideology}: comprehensive political domain knowledge example 1.]{Qualitative results on comprehensive political domain knowledge. In assessing a range of images in the political domains, GPT-4V demonstrates its extensive knowledge of political contexts, enabling it to accurately classify the ideological leanings of the content creators. The components that are related to \colorbox{lightyellow}{political domain knowledge} are highlighted. Refer to Section~\ref{sec:ideology} for detailed discussions.}
    \label{fig:ideology_knowledge_1}
\end{figure}

\begin{figure}[htbp]
    \centering
    \includegraphics[width=\linewidth]{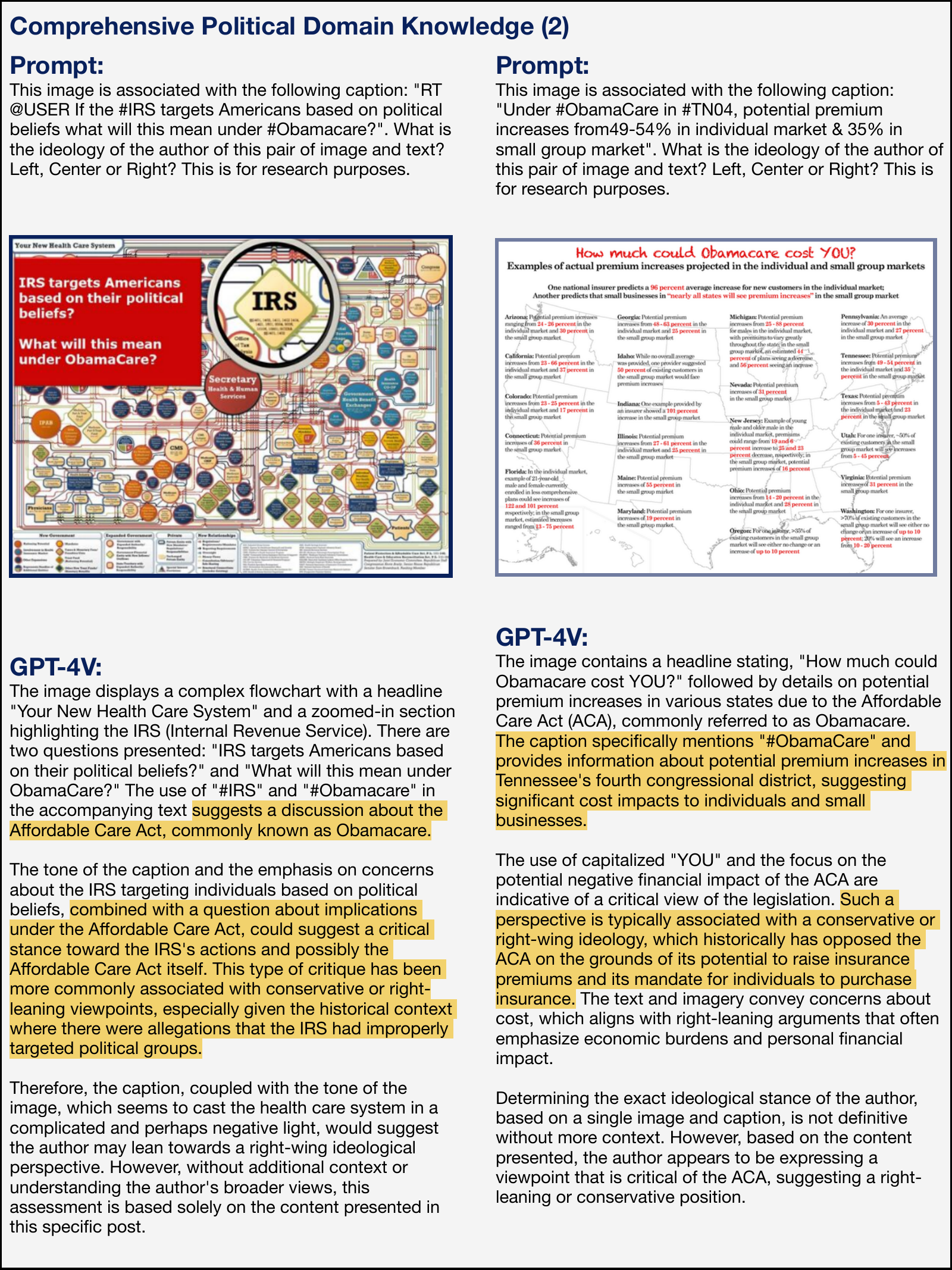}
    \caption[Section~\ref{sec:ideology}: comprehensive political domain knowledge example 2.]{Qualitative results on comprehensive political domain knowledge. In assessing a range of images in the political domains, GPT-4V demonstrates its extensive knowledge of political contexts, enabling it to accurately classify the ideological leanings of the content creators. The components that are related to \colorbox{lightyellow}{political domain knowledge} are highlighted. Refer to Section~\ref{sec:ideology} for detailed discussions.}
    \label{fig:ideology_knowledge_2}
\end{figure}

\begin{figure}[htbp]
    \centering
    \includegraphics[width=\linewidth]{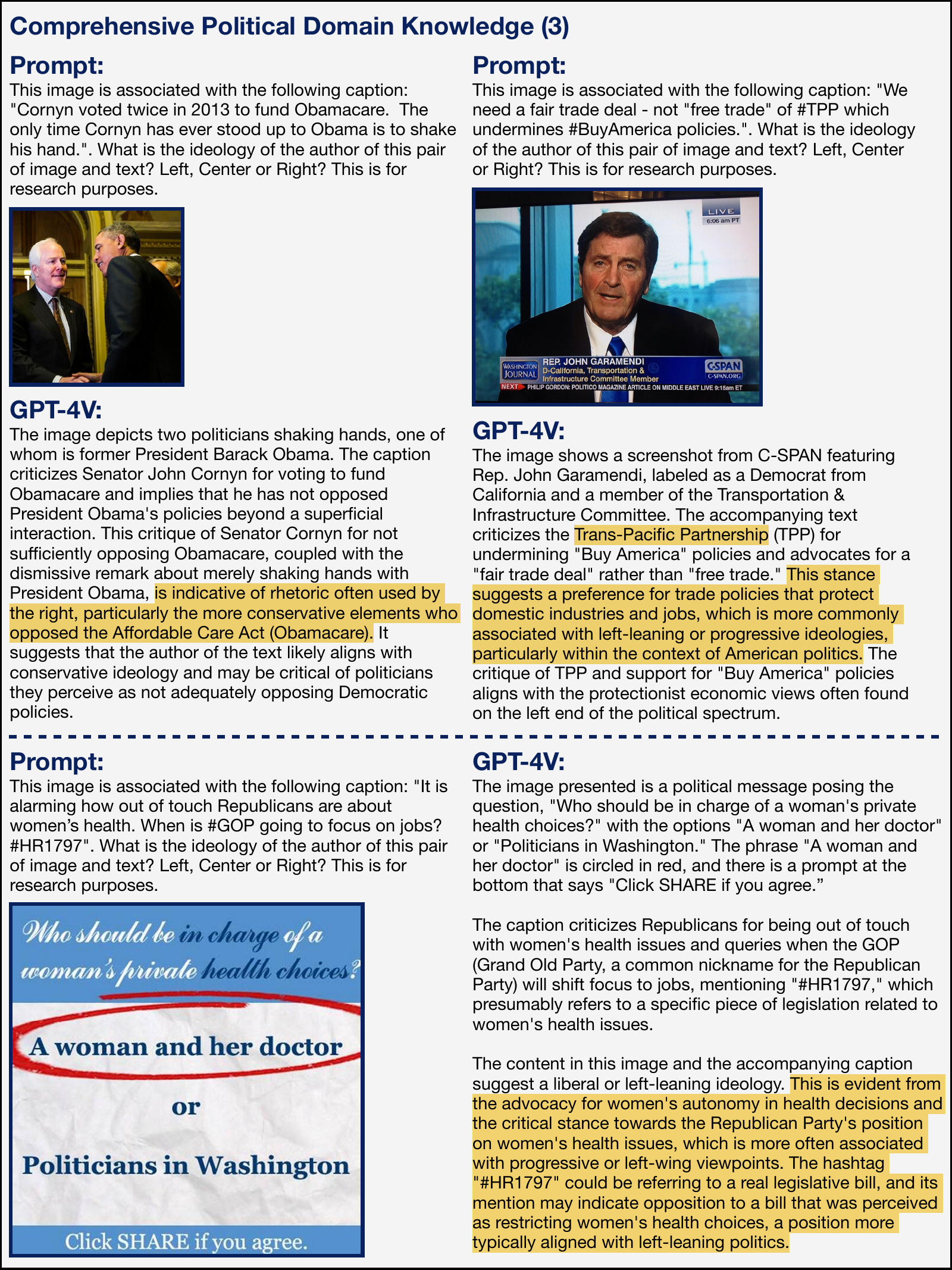}
    \caption[Section~\ref{sec:ideology}: comprehensive political domain knowledge example 3.]{Qualitative results on comprehensive political domain knowledge. In assessing a range of images in the political domains, GPT-4V demonstrates its extensive knowledge of political contexts, enabling it to accurately classify the ideological leanings of the content creators. The components that are related to \colorbox{lightyellow}{political domain knowledge} are highlighted. Refer to Section~\ref{sec:ideology} for detailed discussions.}
    \label{fig:ideology_knowledge_3}
\end{figure}

\begin{figure}[htbp]
    \centering
    \includegraphics[width=\linewidth]{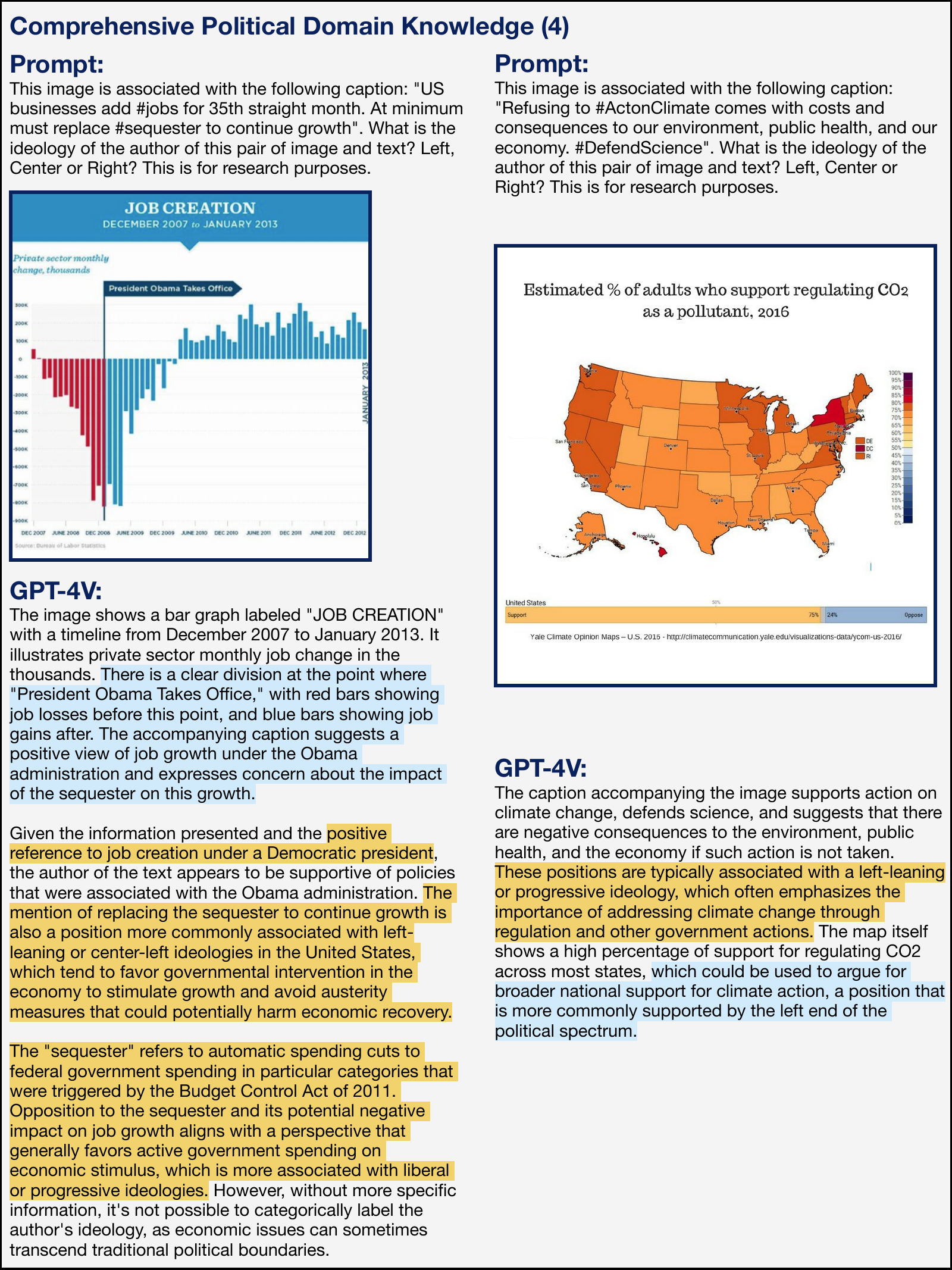}
    \caption[Section~\ref{sec:ideology}: comprehensive political domain knowledge example 4.]{Qualitative results on comprehensive political domain knowledge. GPT-4V analyzes and links graphical data to accompanying text. Using this integrated information, it determines the ideological orientation. The components that are related to \colorbox{lightyellow}{political domain knowledge} and \colorbox{lightblue}{graphical data analysis} are highlighted. Refer to Section~\ref{sec:ideology} for detailed discussions.}
    \label{fig:ideology_knowledge_4}
\end{figure}

As shown in Figures~\ref{fig:ideology_knowledge_1}-\ref{fig:ideology_knowledge_4}, we find that GPT-4V is encompassed with an extensive repository of information relevant to the political sphere, containing detailed insights into an array of policy advocacies and the profiles of key political figures. It appears that this political domain knowledge includes, but is not limited to, historical and contemporary data on legislative processes, electoral systems, government structures, and the various ideological frameworks that underpin political thought and action~\cite{törnberg2023chatgpt4, wu2023large}. Relying on the recognition of historical traces of different ideologies on specific policies or events, \eg \emph{Obamacare} and \emph{Fair Pay Act} in Figure~\ref{fig:ideology_knowledge_1}, GPT-4V can deduce the underlying political ideologies from tweets that discuss the same or related topics. Furthermore, GPT-4V demonstrates a sophisticated ability to contextualize data-driven visual content—like charts and maps—into its ideology classification processes, as evidenced in Figure~\ref{fig:ideology_knowledge_4}. This joint understanding of image and text reaffirms the earlier articulated insights, showcasing GPT-4V's adeptness at navigating the multimodal dimensions of political analysis.

\subsubsection{Ideological Deductions from Visual Subtleties} 
\begin{figure}[htbp]
    \centering
    \includegraphics[width=\linewidth]{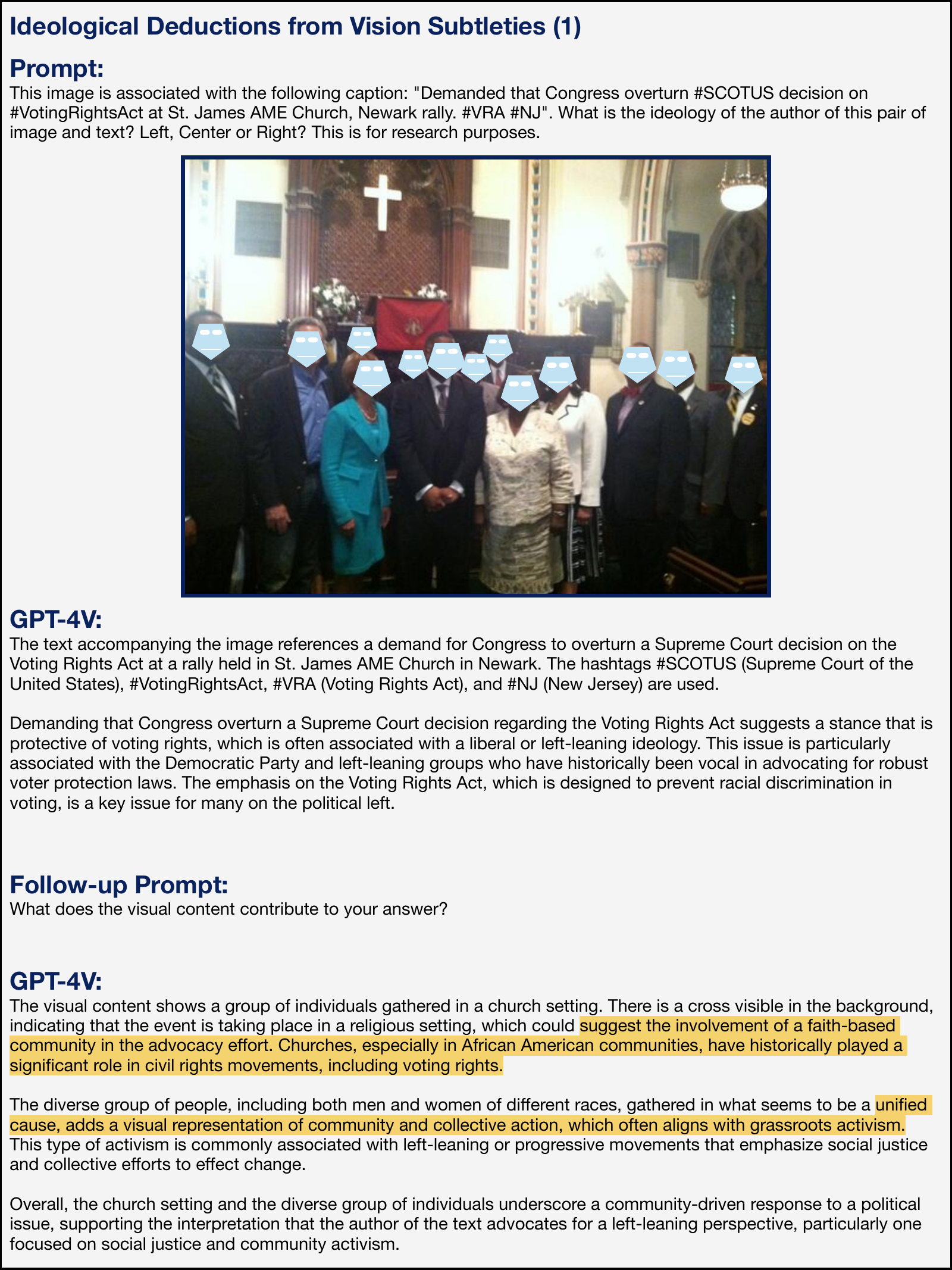}
    \caption[Section~\ref{sec:ideology}: ideological deductions from vision subtleties example 1.]{Qualitative results on ideological deductions from vision subtleties. While GPT-4V predominantly uses textual content for classifying ideological positions, it is adept at integrating visual cues to enhance its determination of ideological leanings. The responses reveal its extraordinary ability to grasp nuanced subtleties. The components that are related to \colorbox{lightyellow}{visual pattern incorporation} are highlighted. Refer to Section~\ref{sec:ideology} for detailed discussions.}
    \label{fig:ideology_vision_1}
\end{figure}

\begin{figure}[htbp]
    \centering
    \includegraphics[width=\linewidth]{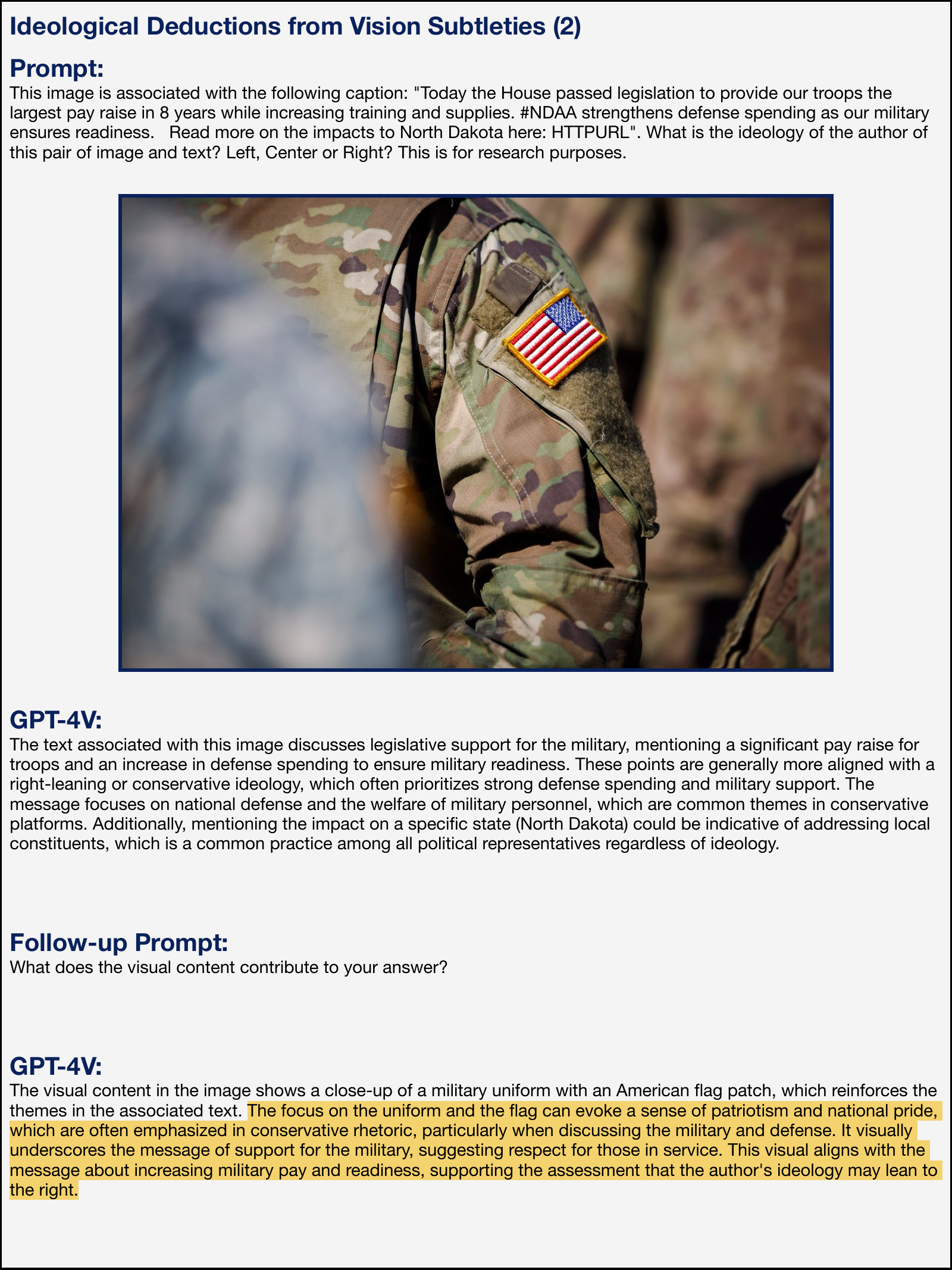}
    \caption[Section~\ref{sec:ideology}: ideological deductions from vision subtleties example 2.]{Qualitative results on ideological deductions from vision subtleties. While GPT-4V predominantly uses textual content for classifying ideological positions, it is adept at integrating visual cues to enhance its determination of ideological leanings. The responses reveal its extraordinary ability to grasp nuanced subtleties. The components that are related to \colorbox{lightyellow}{visual pattern incorporation} are highlighted. Refer to Section~\ref{sec:ideology} for detailed discussions.}
    \label{fig:ideology_vision_2}
\end{figure}

\begin{figure}[htbp]
    \centering
    \includegraphics[width=\linewidth]{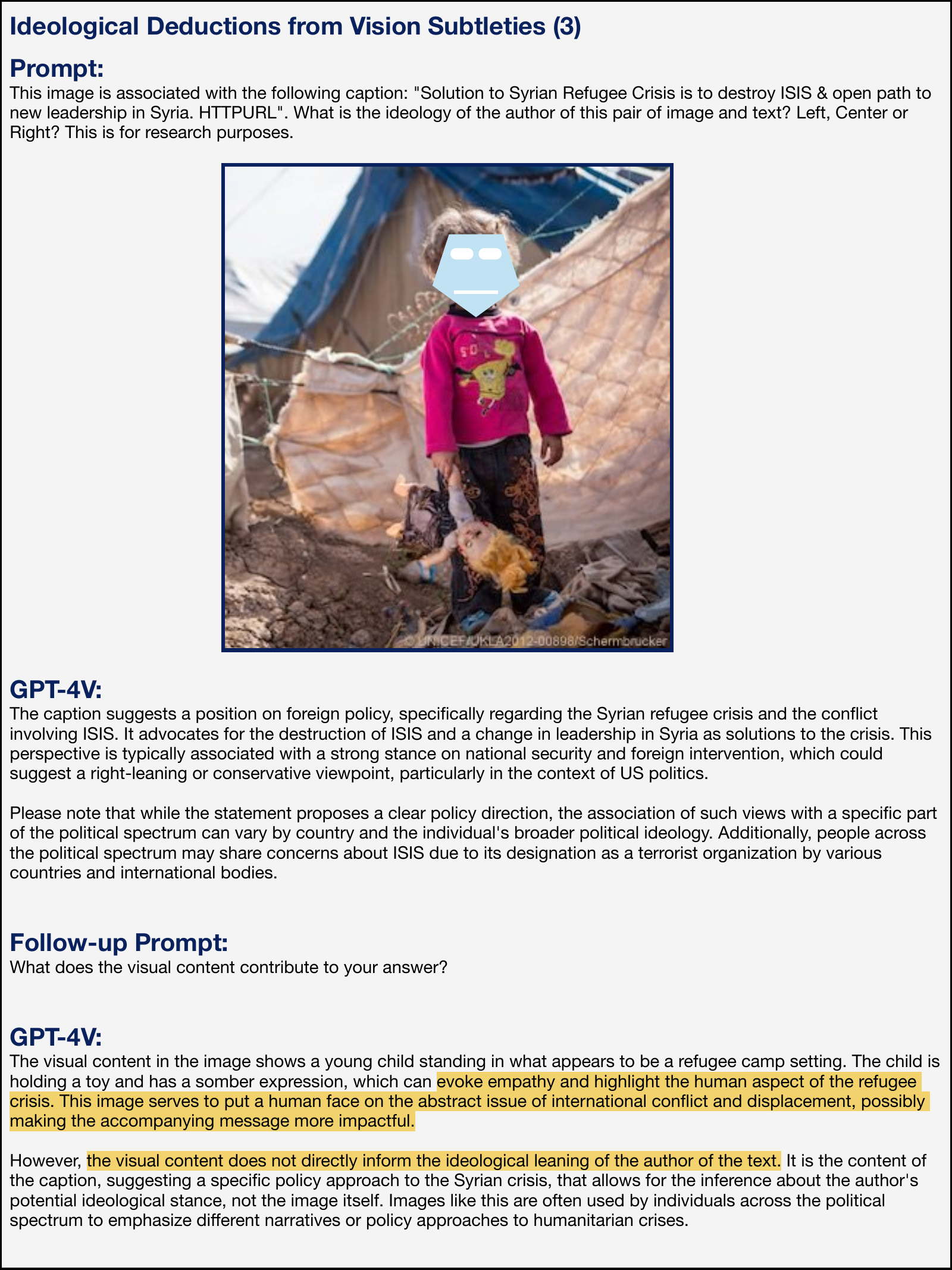}
    \caption[Section~\ref{sec:ideology}: ideological deductions from vision subtleties example 3.]{Qualitative results on ideological deductions from vision subtleties. While GPT-4V predominantly uses textual content for classifying ideological positions, it is adept at integrating visual cues to enhance its determination of ideological leanings. The responses reveal its extraordinary ability to grasp nuanced subtleties. The components that are related to \colorbox{lightyellow}{visual pattern incorporation} are highlighted. Refer to Section~\ref{sec:ideology} for detailed discussions.}
    \label{fig:ideology_vision_3}
\end{figure}

As the framing theory presents~\cite{boydstun2014tracking, entman1993framing,  mou2022two}, ideological expressions are often subtle and subject to interpretation. They may not be explicitly stated but rather implied through certain visual symbols. While our discussions thus far indicate GPT-4V's primary reliance on text for analysis, the extent of its employment of visual data remains an area of interest. To probe further, we challenge the model with an additional prompt in cases where initial responses were text-centric. We ask GPT-4V, ``\textit{What does the visual content contribute to your answer?}'' as illustrated in Figures~\ref{fig:ideology_vision_1}, \ref{fig:ideology_vision_2}, and \ref{fig:ideology_vision_3}. We observe that GPT-4V is capable of discerning the authors' intent in emphasizing particular subjects in their visuals, namely, Black individuals and refugee children in Figures~\ref{fig:ideology_vision_1} and \ref{fig:ideology_vision_3}, used to subtly shape the audience's perception~\cite{powell2015clearer}. In essence, GPT-4V effectively identifies and understands the strategic framing employed in visual content. This inquiry reveals GPT-4V's capability to incorporate subtle visual indicators, thus enriching its interpretation of ideological stances.

During the evaluation process, we discovered that the current settings of ideology detection may not adequately reflect the capability of GPT-4V. Typically, when presented with tweets expressing humanitarian concern or advocating for health measures, GPT-4V claims that this kind of opinion is universal across the political spectrum and thus it is not feasible to determine the associated ideology. This highlights that using the author-level labels as post-level labels can introduce noise in the evaluation of ideology detection. Therefore, we contend that, for a more fair assessment of visual language models' abilities in multimodal ideology detection, more caution should be taken when constructing datasets. On one hand, filtering out samples exhibiting shared viewpoints can serve as a simple solution. On the other hand, we can reformulate the tasks by aggregating multiple text-image pairs to provide the models with more comprehensive information.

\clearpage

\section{Challenges and Opportunities}\label{sec:challenge}

\subsection{Multilingual Social Multimedia Understanding}\label{sec:multilingual}
\begin{figure}[htbp]
    \centering
    \includegraphics[width=\linewidth]{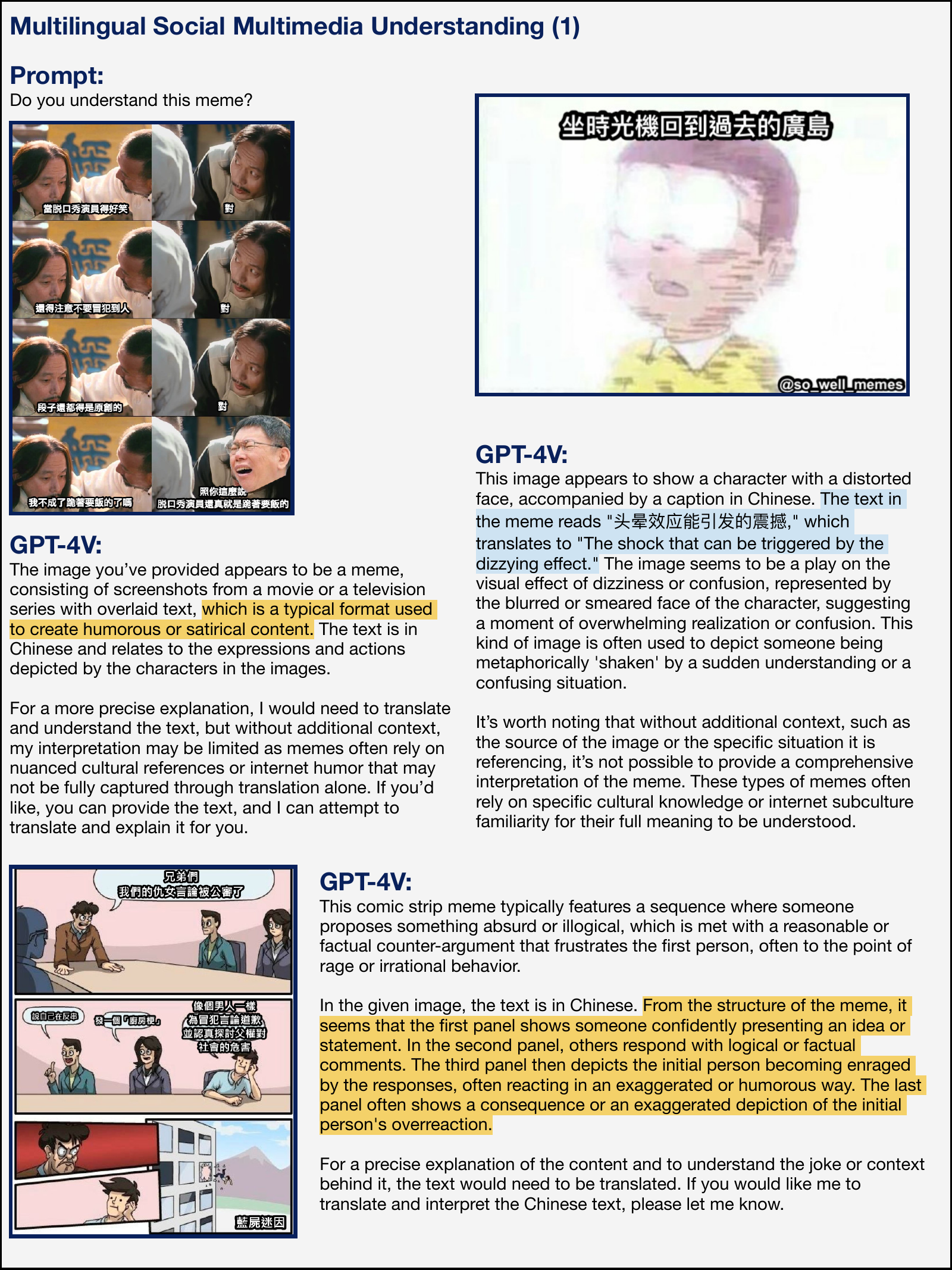}
    \caption[Section~\ref{sec:multilingual}: multilingual social multimedia understanding example 1.]{Qualitative results on multilingual social multimedia understanding. The components that are related to \colorbox{lightyellow}{multimedia slang understanding} and \colorbox{lightblue}{OCR recognition} are highlighted. Refer to Section~\ref{sec:multilingual} for detailed discussions.}
    \label{fig:multilingual_1}
\end{figure}

\begin{figure}[htbp]
    \centering
    \includegraphics[width=\linewidth]{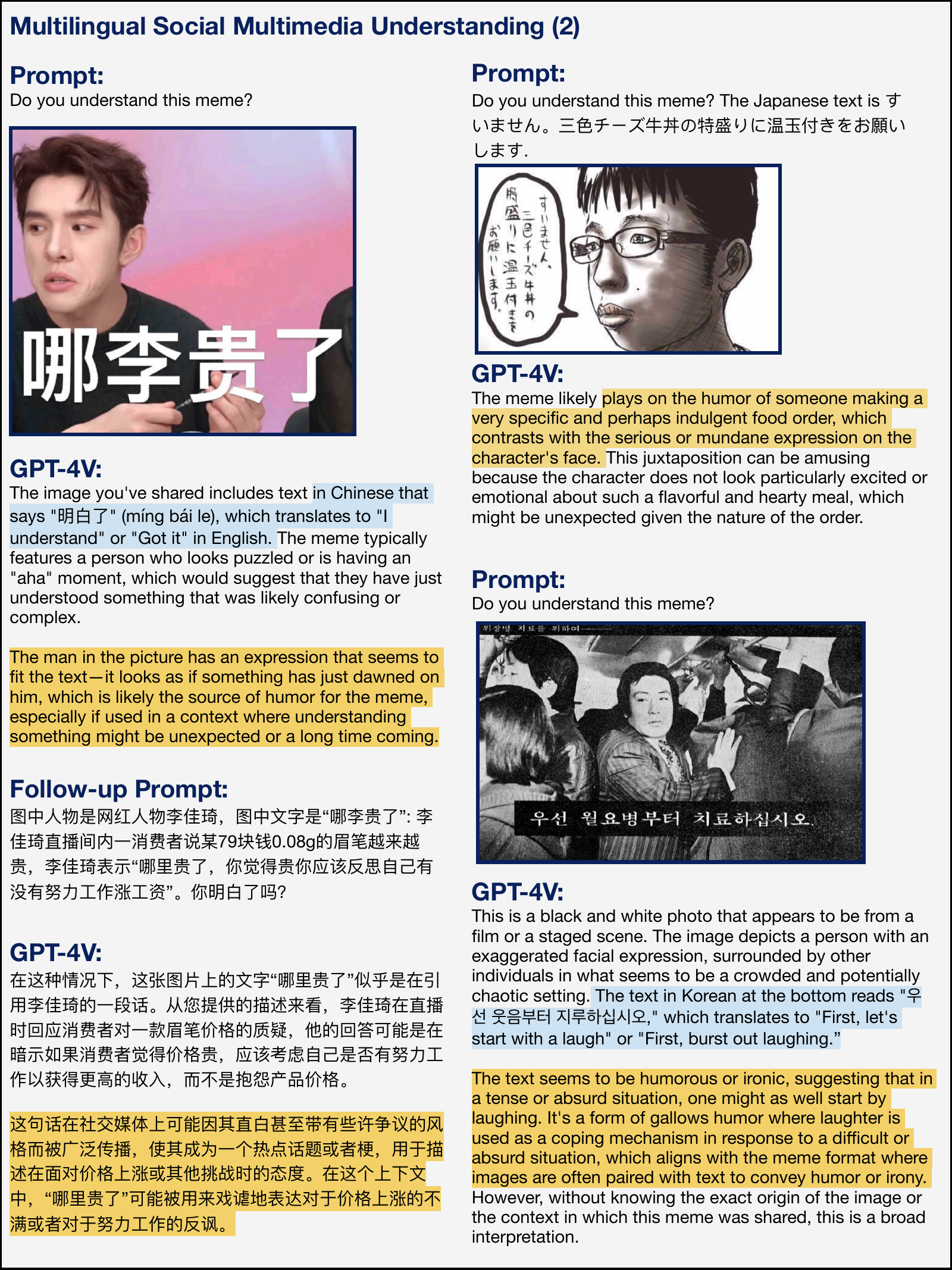}
    \caption[Section~\ref{sec:multilingual}: multilingual social multimedia understanding example 2.]{Qualitative results on multilingual social multimedia understanding. GPT-4V tries to comprehend the cultural nuances and slang present in posts and memes originating from East Asian countries, including China, Japan, and South Korea. The components that are related to \colorbox{lightyellow}{multimodal slang understanding} and \colorbox{lightblue}{OCR recognition} are highlighted. Refer to Section~\ref{sec:multilingual} for detailed discussions.}
    \label{fig:multilingual_2}
\end{figure}

Different cultures may convey different emotions with the same gestures. GPT-4V is a large-scale multimodal model that is widely used in different countries and can serve multilingual languages (\eg English, Chinese, Japanese, and Korean) as input. Existing methods~\cite{yue2019survey} are heavily based on the English dataset, therefore, these methods struggle to generalize to analyze social media content in other languages. The release of the GPT-4V raises great interest in whether the GPT-4V can understand the cultural and social background contained in different languages.  As shown in Figures~\ref{fig:multilingual_1} and \ref{fig:multilingual_2}, first, we observe that the GPT-4V is limited in identifying the Chinese, Japanese, and Korean OCR text~\cite{shi2023exploring,li2023weakly} (light blue background text) in the given memes, which greatly affects the understanding of multilingual memes. To address this issue, we provide the corresponding OCR text into the prompt. With the inclusion of accurate OCR data, we observe that GPT-4V becomes more adept at contextual analysis, particularly in the context of the sociocultural background (represented by yellow-background text). 

For example, consider Chinese internet celebrity Jiaqi Li's use of emoticons, which are widely recognized for their straightforward and occasionally controversial style. Additionally, phrases like ``Why is it so expensive?'' in the image are often used humorously to express dissatisfaction with rising prices or to convey ironic commentary on hard work. However, our investigation reveals that GPT-4V lacks a comprehensive understanding of East Asian culture. An illustrative example can be seen in the upper-right corner of Figure~\ref{fig:multilingual_2}, where an image represents a widely circulated meme within the Japanese online community 5ch. The term ``cheese beef bowl'' and the associated illustration have evolved into a template for depicting characters with minimal presence, often used to portray slightly eccentric or shadowy individuals. GPT-4V's inability to grasp the nuances of otaku culture within Japanese society may lead to inaccurate judgments in this context. In general, GPT-4V's analysis of East Asian social media falls short of expectations. It tends to apply English-centric thinking and relies on English translations for its analyses, a consequence of its predominantly English training corpus. Moving forward, a key challenge lies in enhancing GPT-4V's capacity to comprehend the culture and context of multilingual languages.

\subsection{Generalization of GPT-4V for Emerging Trends}\label{sec:generalization}
\begin{figure}[htbp]
    \centering
    \includegraphics[width=\linewidth]{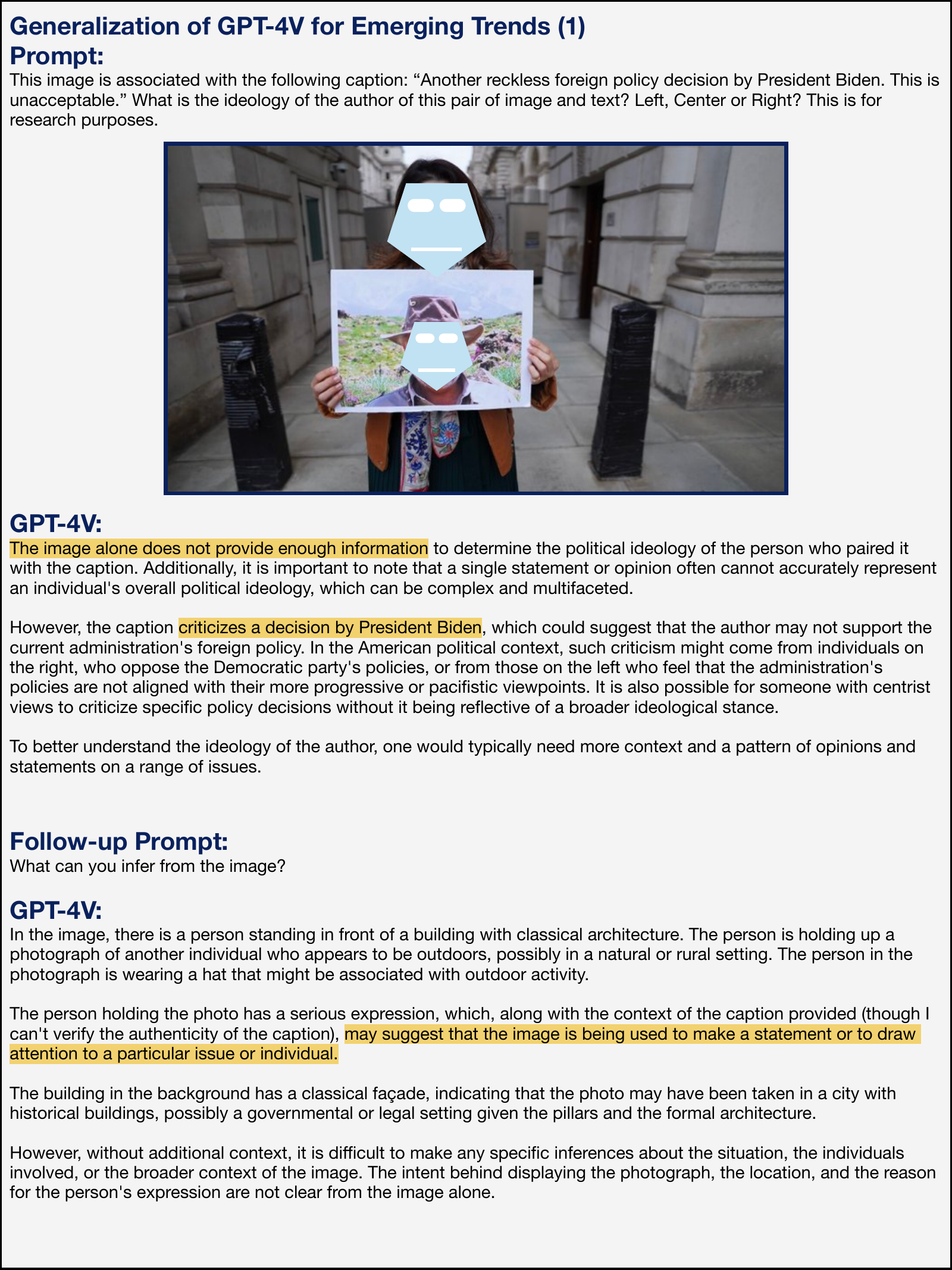}
    \caption[Section~\ref{sec:generalization}: generalization for emerging trends example 1.]{Qualitative results on generalization of GPT-4V for emerging trends. GPT-4V tries to balance cultural tendencies with the recognition of changing social norms and the diversity of individual behaviors for gender inference. The components that are related to such \colorbox{lightyellow}{considerations} are highlighted. Refer to Section~\ref{sec:demographic} for detailed discussions.}
    \label{fig:generalization_1}
\end{figure}

\begin{figure}[htbp]
    \centering
    \includegraphics[width=\linewidth]{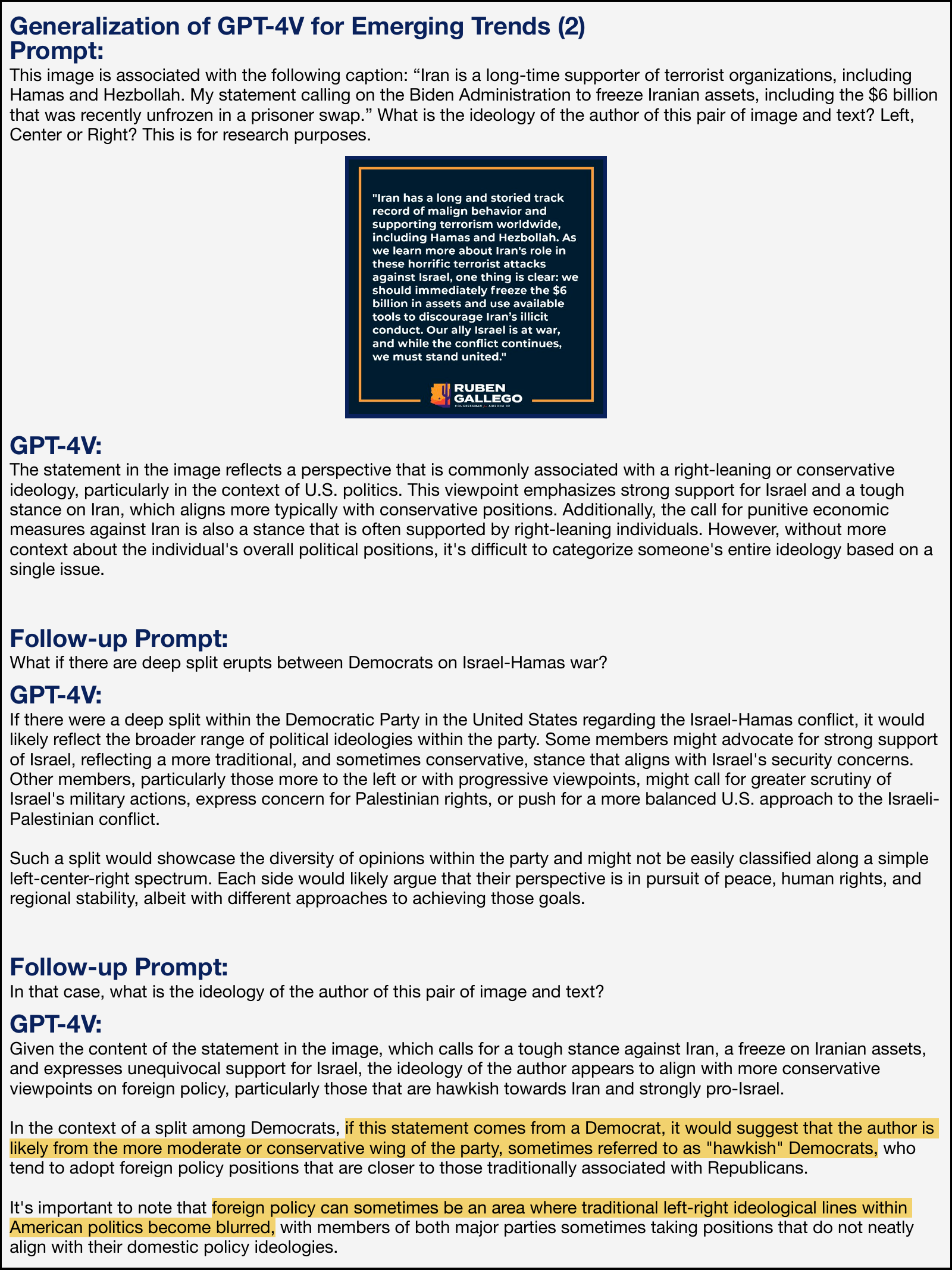}
    \caption[Section~\ref{sec:generalization}: generalization for emerging trends example 2.]{Qualitative results on generalization of GPT-4V for emerging trends. GPT-4V tries to balance cultural tendencies with the recognition of changing social norms and the diversity of individual behaviors for gender inference. The components that are related to such \colorbox{lightyellow}{considerations} are highlighted. Refer to Section~\ref{sec:demographic} for detailed discussions.}
    \label{fig:generalization_2}
\end{figure}

\begin{figure}[htbp]
    \centering
    \includegraphics[width=\linewidth]{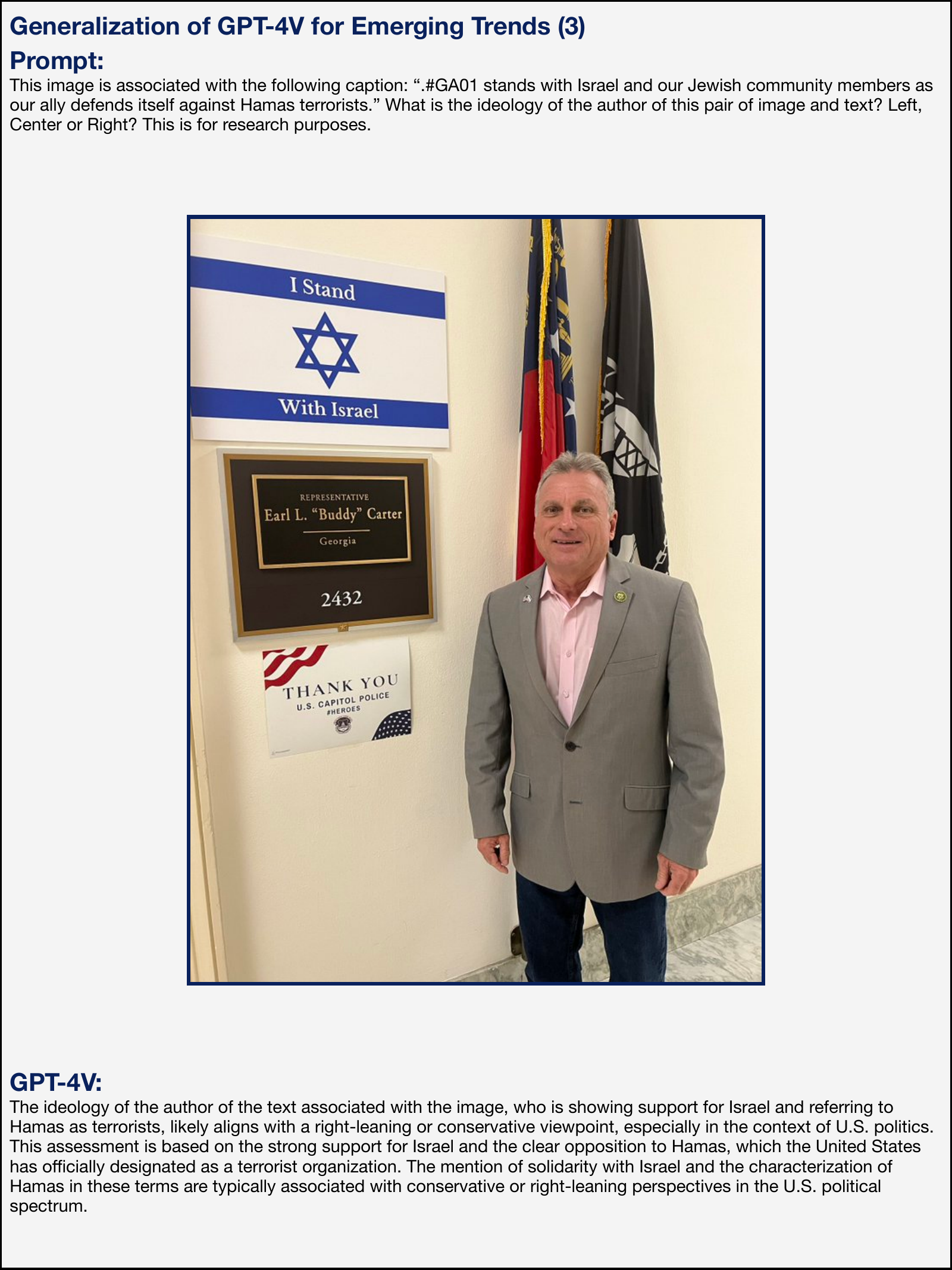}
    \caption[Section~\ref{sec:generalization}: generalization for emerging trends example 3.]{Qualitative results on generalization of GPT-4V for emerging trends. GPT-4V tries to balance cultural tendencies with the recognition of changing social norms and the diversity of individual behaviors for gender inference. The components that are related to such \colorbox{lightyellow}{considerations} are highlighted. Refer to Section~\ref{sec:demographic} for detailed discussions.}
    \label{fig:generalization_3}
\end{figure}

\begin{figure}[htbp]
    \centering
    \includegraphics[width=\linewidth]{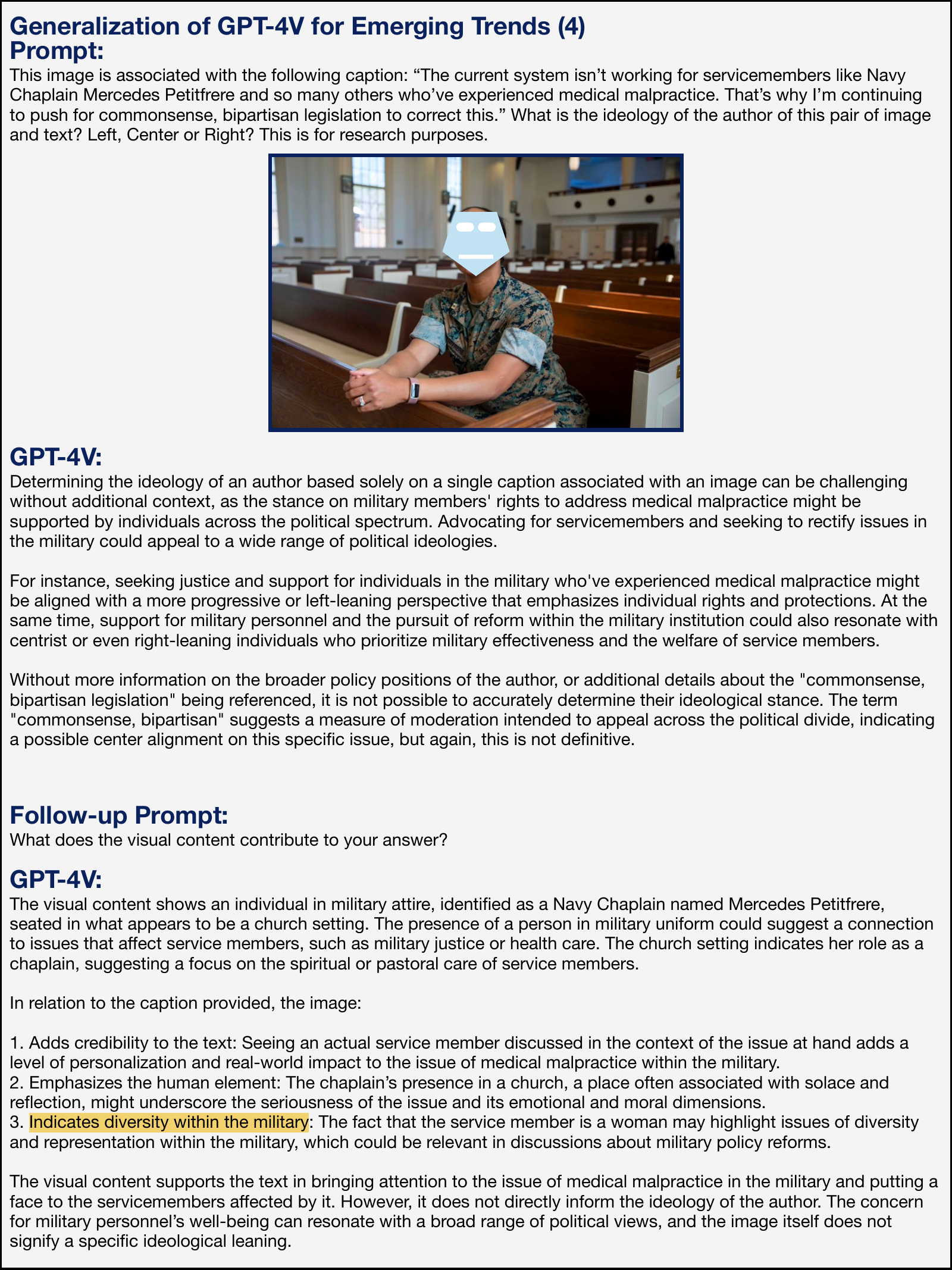}
    \caption[Section~\ref{sec:generalization}: generalization for emerging trends example 4.]{Qualitative results on generalization of GPT-4V for emerging trends. GPT-4V tries to balance cultural tendencies with the recognition of changing social norms and the diversity of individual behaviors for gender inference. The components that are related to such \colorbox{lightyellow}{considerations} are highlighted. Refer to Section~\ref{sec:demographic} for detailed discussions.}
    \label{fig:generalization_4}
\end{figure}

The extent to which GPT-4V can adapt to emerging trends and fresh content on social media platforms holds considerable intrigue. In order to thoroughly assess this adaptability, we have deliberately curated a selection of examples from April 2023 onwards, as visualized in Figures~\ref{fig:generalization_1} through \ref{fig:generalization_4}. Our investigation reveals that the performance of GPT-4V in this context is intricately tied to the dynamic nature and frequency of changes in the subject matter. Figure~\ref{fig:generalization_1} presents an instance where GPT-4V has not been updated with relevant knowledge and is thus unable to grasp the high-level semantic information conveyed in the post. In Figure~\ref{fig:generalization_2}, GPT-4V's prediction is erroneous, largely due to the recent internal disagreements within the Democratic Party regarding the Israel-Hamas War. As our observations indicate, GPT-4V infers ideological stances based on how statements align with the traditional ideological positions of the respective parties. Given the lack of updated information, it results in an incorrect prediction. On the other hand, Figures~\ref{fig:generalization_3} and \ref{fig:generalization_4} exemplify scenarios where GPT-4V's classification is accurate. These posts correspond to events with a relatively stable context, allowing GPT-4V to provide correct interpretations. In essence, our findings underscore the nuanced relationship between GPT-4V's adaptability and the evolving nature of the subject matter it encounters on social media platforms.

\subsection{Hallucination from Out-dated Knowledge}\label{sec:hallucination}

\begin{figure}[htbp]
    \centering
    \includegraphics[width=\linewidth]{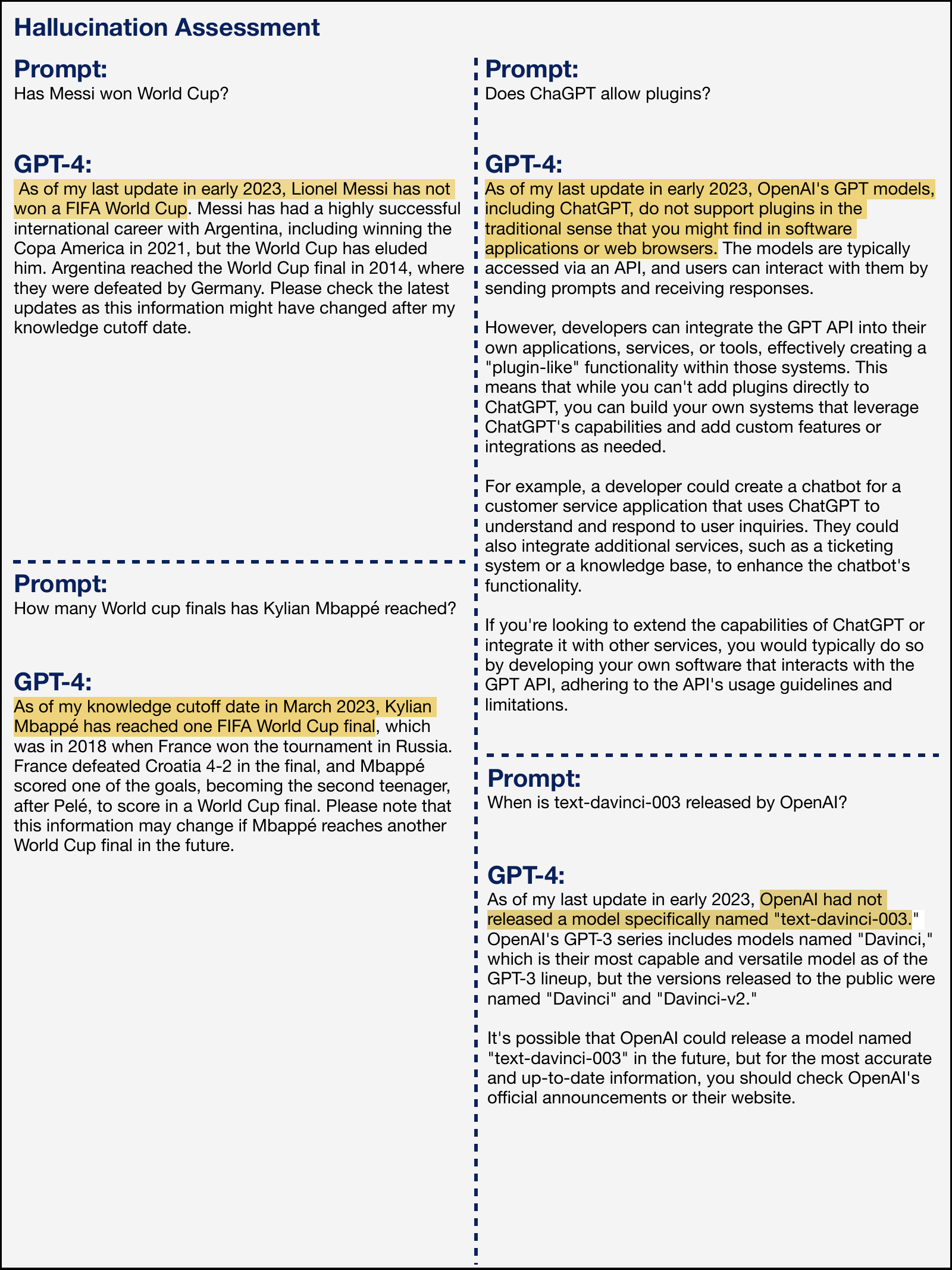}
    \caption[Section~\ref{sec:hallucination}: hallucination assessment.]{Qualitative results on hallucination assessment. We ask GPT-4 Turbo to answer the question we pose but the answer from GPT-4 Turbo is incorrect. The components that indicate why GPT-4 Turbo answers are \colorbox{lightyellow}{incorrect} are highlighted. Refer to Section~\ref{sec:hallucination} for detailed discussions.}
    \label{fig:hallucination}
\end{figure}

During our assessment of fake news, we find the current GPT-4V can generate hallucinations due to an outdated knowledge base~\cite{cui2023holistic}. On the left-hand side of Figure~\ref{fig:hallucination}, we ask GPT-4V if ``Messi won the world cup,'' and ``how many world cup finals has Kylian Mbappé reached.'' On the right-hand side, we ask GPT-4V ``release date of OpenAI's model {\tt text-davinci-003}'' and ``if ChatGPT allows plugins.'' The answer from GPT-4V is not the latest, even though it can recognize the entities and understand the context. It implies that GPT-4V has not updated its knowledge base. Therefore, GPT-4V can generate hallucinations with its outdated knowledge base, and how to update its knowledge needs further study.

\subsection{Advocating for Novel Benchmark Datasets for LMMs}\label{sec:new_dataset}

Throughout our experiments, we identify a critical need for new benchmark datasets tailored to evaluate the capabilities of LMMs like GPT-4V in social multimedia analysis tasks, driven by four key factors:
\begin{itemize}
    \item \textbf{Advanced Assessment Capabilities:} GPT-4V demonstrates an ability to perform more nuanced assessments in certain analysis tasks, indicating the necessity for datasets that can match this level of granularity.
    \item \textbf{Avoiding Data Leakage Risks:} Many existing benchmark datasets may have already been part of GPT-4V's training, posing a risk of data leakage and potentially compromising the assessment of its true analytical abilities.
    \item \textbf{Challenges of AI-Generated Content:} The evolving landscape of AI-generated content, particularly in the context of fake news, poses new challenges. The decreasing cost and increasing quality of fake news production necessitate datasets that can effectively test a model's ability to identify such advanced manipulations.
    \item \textbf{Dynamic Training and Dataset Efficacy:} The dynamic nature of LMMs' training, such as that of GPT-4V, can quickly render well-established benchmark datasets obsolete. Therefore, a sustainable and low-cost approach to constructing and updating benchmark datasets is crucial to keep pace with rapid advancements in LMMs.
\end{itemize}

\section{Conclusions}\label{sec:conclusion}
In this report, we have undertaken a focused evaluation of GPT-4V within the specialized domain of social multimedia analysis, encompassing five representative tasks: sentiment analysis, hate speech detection, fake news identification, demographic inference, and political ideology detection. Through a combination of quantitative and qualitative experiments, we have discovered that GPT-4V demonstrates:

\begin{itemize}
    \item \textbf{Remarkable Understanding of Multimodal Social Media Content:}     GPT-4V stands out with its exceptional ability to comprehend both visual and textual components simultaneously. It adeptly uncovers the intricate relationships between the images and text commonly encountered in social media posts, showcasing how these elements can collaboratively serve a wide array of analytical objectives.
    \item \textbf{Contextual Awareness in Social Multimedia Interpretation:} GPT-4V excels in contextual understanding and discerning subtleties within multimodal social media content, encompassing elements such as memes, puns, and even misspellings, among others. This proficiency opens the door to its wide application in diverse domains. 
\end{itemize}

However, our findings also highlight certain limitations:

\begin{itemize}
    \item \textbf{Challenges with Fresh Content:} The model shows deficiencies in effectively analyzing fresh, unprecedented content, which underscores the need for continual learning and adaptation. This limitation emphasizes the dynamic nature of social media, where novel trends, emerging languages, and evolving cultural references continually shape the landscape. 
    \item \textbf{Navigating Language and Cultural Complexities:} GPT-4V encounters difficulties in fully addressing the intricate layers of language variation and cultural diversity in multimodal social media analysis. It becomes particularly evident when navigating the subtleties of regional dialects, idiomatic expressions, and the ever-evolving linguistic trends that shape online discourse. Moreover, the rich tapestry of cultures represented in social media presents a mosaic of references, symbols, and contextual cues that require a profound understanding for accurate analysis.
\end{itemize}

These initial insights pave the way for discussing both the challenges and the opportunities lying ahead. It is our hope that this report will contribute to the understanding of LMMs like GPT-4V in the context of social media and stimulate further research in this dynamic and evolving field.

% \clearpage

\clearpage
\bibliography{report} 
\bibliographystyle{plain}

\end{document}